\documentclass[conference]{IEEEtran}
\usepackage{times}

\usepackage[numbers]{natbib}
\usepackage[english]{babel}
\usepackage{multicol}
\usepackage[bookmarks=true,pagebackref=true]{hyperref}
\usepackage[nolist]{acronym}
\usepackage{amsmath}
\usepackage{amsthm}
\usepackage{amsfonts}
\usepackage{graphicx}
\usepackage{sidecap}
\usepackage{wrapfig}
\usepackage{array}
\usepackage{caption}
\usepackage{ragged2e}
\usepackage{subcaption}
\usepackage{makecell}
\usepackage[nice]{nicefrac}
\usepackage[font=small,labelfont=small]{caption}
\usepackage{booktabs}
\usepackage{float}
\usepackage{siunitx}
\usepackage[table]{xcolor}
\usepackage[ruled,vlined,linesnumbered]{algorithm2e}
\usepackage{algpseudocode}
\usepackage{stfloats}
\usepackage[framemethod=tikz]{mdframed}

\makeatletter
\let\labelindent\relax
\makeatother
\usepackage{enumitem}

\graphicspath{ {Graphics/} }
\theoremstyle{definition}
\newtheorem{definition}{Definition}
\newtheorem{theorem}{Theorem}
\newcolumntype{P}[1]{>{\RaggedRight\arraybackslash}p{#1}}

\newcommand{\dif}[0]{\mathrm{d}} % differential, e.g. dx/dy

\begin{document}

\title{Diffeomorphic Obstacle Avoidance for Contractive Dynamical Systems via Implicit Representations}

\author{
\authorblockN{Ken-Joel Simmoteit\\}
\authorblockA{Karlsruhe Institute of Technology\\
Karlsruhe, Germany\\
Ken-Joel.Simmoteit@student.kit.edu}
\and
\authorblockN{Philipp Schillinger}
\authorblockA{Bosch Center for Artificial Intelligence\\
Renningen, Germany\\
Philipp.Schillinger@de.bosch.com}
\and
\authorblockN{Leonel Rozo}
\authorblockA{Bosch Center for Artificial Intelligence\\
Renningen, Germany\\
Leonel.Rozo@de.bosch.com}
}

\begin{acronym}[paper]
    \acro{SDF}[SDF]{Signed Distance Fields}
    \acro{RDF}[RDF]{Robot Signed Distance Fields}
    \acro{CDF}[CDF]{Configuration Space Distance Fields}
    \acro{BP}[BP]{Bernstein polynomial basis functions}
    \acro{NN}[NN]{Neural Network}
    \acro{MLP}[MLP]{Multilayer Perceptron}
    \acro{NCDS}[NCDS]{Neural Contractive Dynamical System}
    \acro{NF}[NF]{Normalizing Flow}
    \acro{CNF}[CNF]{Conditional Normalizing Flow}
    \acro{M-Flow}[M-Flow]{Manifold Flow}
    \acro{IRKF}[IRKF]{Injective Robot Kinematics Flow}
    \acro{DT}[DT]{Diffeomorphic Transforms}
    \acro{SDT}[SDT]{Signed Distance Field Diffeomorphic Transform}
    \acro{SDDC}[SDDC]{Signed Distance Field Differential Coordinate Change}
    \acro{SDC}[SDC]{Signed Distance Field Coordinate Change}
    \acro{MM}[MM]{Modulation Matrix}
    \acro{IK}[IK]{Inverse Kinematics}
    \acro{ODE}[ODE]{Ordinary Differential Equation}
    \acro{LNS}[LNS]{Latent Nullspace States}
    \acro{LfD}[LfD]{Learning from Demonstration}
    \acro{DTWD}[DTWD]{Dynamic Time Warping Distance}
    \acro{MR}[MR]{Modulation Range}
    \acro{PL}[PL]{Penetration Length}
    \acro{RFC}[RFC]{Relative Flow Curvature}
    \acro{MJ}[MJ]{Modulation Jerk}
    \acro{QP}[QP]{Quadratic Programming}
    \acro{ROS2}[ROS2]{Robot Operating System 2}
    \acro{RTB}[RTB]{Robotics Toolbox}
    \acro{KNN}[KNN]{k-Nearest Neighbors}
    \acro{ARPF}[ARPF]{Artificial Repulsive Potential Field}
    \acro{VM}[VM]{Vector Field Misalignment}
    \acro{HM}[HM]{Hilbert Map}
    \acro{CBF}[CBF]{Control Barrier Function}
\end{acronym}

\maketitle

\begin{abstract}
Ensuring safety and robustness of robot skills is becoming crucial as robots are required to perform increasingly complex and dynamic tasks.
The former is essential when performing tasks in cluttered environments, while the latter is relevant to overcome unseen task situations.
This paper addresses the challenge of ensuring both safety and robustness in dynamic robot skills learned from demonstrations. Specifically, we build on neural contractive dynamical systems to provide robust extrapolation of the learned skills, while designing a full-body obstacle avoidance strategy that preserves contraction stability via diffeomorphic transforms. 
This is particularly crucial in complex environments where implicit scene representations, such as Signed Distance Fields (SDFs), are necessary. To this end, our framework called Signed Distance Field Diffeomorphic Transform, leverages SDFs and flow-based diffeomorphisms to achieve contraction-preserving obstacle avoidance.
We thoroughly evaluate our framework on synthetic datasets and several real-world robotic tasks in a kitchen environment. 
Our results show that our approach locally adapts the learned contractive vector field while staying close to the learned dynamics and without introducing highly-curved motion paths, thus outperforming several state-of-the-art methods.
\end{abstract}

\IEEEpeerreviewmaketitle

\section{Introduction}
\label{sec:introduction}
Teaching robots skills has been a key challenge in robotics research for over four decades \cite{ALozano-Perez1983}. The dominant paradigm involves learning robotic skills through expert examples, commonly referred to as \ac{LfD}~\cite{Argall2009,Billard2016,Ravichandar2020,Schaal1996}. By enabling robots to execute complex motion skills, \ac{LfD} has led to numerous promising methods~\cite{Billard2008, Huang19:KMP, Ijspeert13:dmp, Rana2020} with applications in flexible manufacturing~\cite{HernandezMoreno2024,Rozo2024}, household environments~\cite{Rana2020,Welschehold16:LfDhousehold}, human-robot collaboration~\cite{Koskinopoulou16:LfD4HRC,Rozo16:pHRC_LfD,Rozo2018:EditorialHRC}, and robot-assisted minimally invasive surgery~\cite{Su2021}, among many others. 
\looseness=-1

For safe operation in human-centric and dynamic settings, learned skills must be stable and reliable, avoiding unexpected movements under unseen situations such as different task conditions or external perturbations. This highlights the importance of \textit{stable \ac{LfD} skills}~\cite{Ravichandar2020}. Early efforts to ensure stability in \ac{LfD} focused on asymptotic stability criteria, often leveraging Lyapunov stability~\cite{Dawson2022SafeCW,Khansari-Zadeh2011,Rana2020,Urain2020}. 
While Lyapunov stability primarily ensures asymptotic point-wise stability, many robot skills require following complex task-relevant trajectories. To address this limitation, researchers have conceptualized a more general stability criterion, namely the \textit{contraction property} of dynamical systems, which guarantees exponential convergence to a trajectory~\cite{LOHMILLER1998,Tsukamoto2021}. 
This concept has been recently leveraged in \ac{LfD} frameworks, where a robot skill, represented by a first-order dynamical system, is endowed with a contractive behavior~\cite{BeikMohammadi2024,Blocher2017,Ravichandar2018,Singh2021}.

\begin{figure}[t!]
    \centering
    \includegraphics[width=0.5\textwidth]{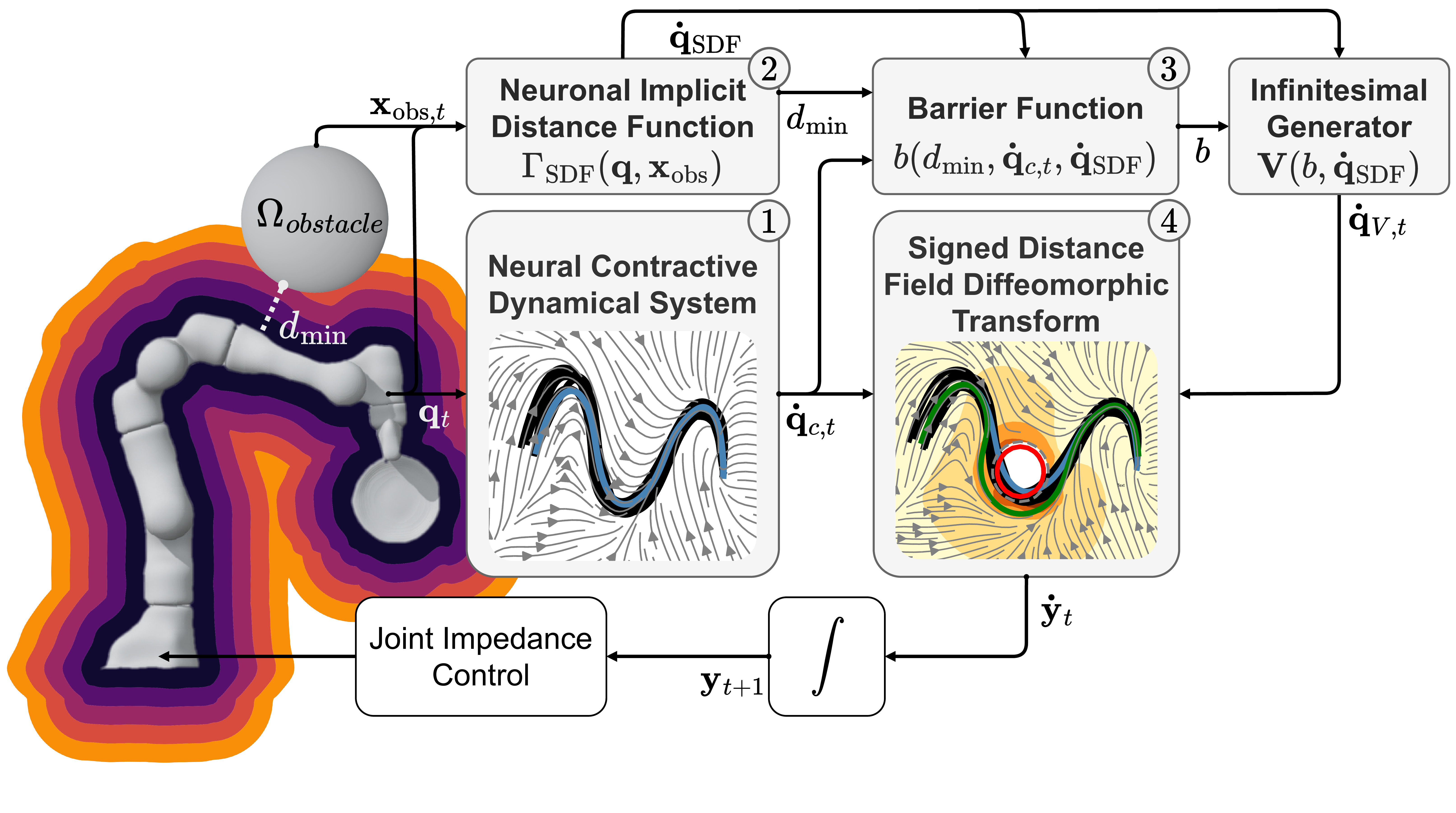}
    \caption{Overview of the proposed \textbf{Signed Distance Field Diffeomorphic Transform}: \emph{(1)} A neural contractive dynamical system; \emph{(2)} A learned implicit distance function; \emph{(3)} A barrier function, and \emph{(4)} a contraction-preserving diffeomorphic transform.}
    \label{fig:sdt_overview}
\end{figure}

In addition to stability, ensuring safety is paramount. This often demands integrating \textit{obstacle avoidance} into robot skills while still providing stability guarantees.
In the context of contractive systems, \citet{Huber2019} introduced the \ac{MM} method to preserve contraction stability during obstacle avoidance. Subsequently, \citet{BeikMohammadi2024} applied this approach to neural contractive dynamical systems (NCDS), enabling stable collision-free motion in the robot's task space. 
Despite these advances, articulated robotic arms require whole-body obstacle avoidance. 
While traditional methods like artificial potential fields~\citep{Khatib1985} and recent approaches like \ac{DT}~\citep{Zhi2022}, tackle this problem via robust joint-space obstacle avoidance solutions that account for asymptotic stability, they do not consider contraction stability guarantees. 
Therefore, preserving contraction during whole-body obstacle avoidance remains largely unexplored, posing a key challenge for robot motion control, which we address in this paper.

When considering cluttered and dynamic environments, such as household settings, conventional obstacle representations using simple geometric primitives are insufficient~\cite{BeikMohammadi2024,Huber2022}, as they rely on coarse approximations and may not be easily integrated with neural architectures. 
Instead, a differentiable \textit{implicit distance representation} is more advantageous as it facilitates a fine-grained, real-time, and resource-efficient solution, while being seamlessly integrated with other neural network architectures. 
This has led to the rise of \ac{SDF}s~\cite{Gropp2020,Irshad2024,Ortiz2022,Park2019}, which have recently been leveraged to learn articulated robot surfaces~\cite{Yiting2023,Koptev2023,YimlinLi2023}. 
In these approaches, the \ac{SDF} is a function of the robot's joint state, capturing the continuous robot geometry over its entire configuration space.
Still, integrating implicit representations with contraction-preserving obstacle avoidance remains an open problem. 
To bride this gap, we propose to combine the robustness of contraction stability with the precision of implicit representations for efficient and safe whole-body obstacle avoidance.

\begin{mdframed}[hidealllines=true,backgroundcolor=blue!5,innerleftmargin=.2cm,
  innerrightmargin=.2cm]
Specifically, \textbf{this paper} introduces a novel framework, the \textbf{\ac{SDT}}, for contraction-preserving obstacle avoidance based on an implicit scene representation. 
The key contributions of this work are:

\begin{enumerate}
    \item \textbf{Contraction-Preserving Implicit Obstacle Avoidance}: We introduce a new method for achieving contraction-preserving obstacle avoidance in contractive dynamical systems by leveraging \ac{DT}, \ac{SDF} techniques, and barrier functions. 
    First, we define an infinitesimal generator derived from the robot's implicit representation and regulate it using barrier functions. Second, using the resulting vector field, we construct a flow that reshapes the contractive dynamics, either through a differential coordinate change, or by applying a standard coordinate change via a pullback.
    \looseness=-1
    \item \textbf{Obstacle Avoidance Metrics}: We introduce a set of quantitative metrics to assess obstacle avoidance performance, focusing on flow curvature and vector field misalignment of the modulated contractive motion.
    \item \textbf{Experiments}: We extensively evaluated our framework on the 2D LASA dataset~\citep{Khansari-Zadeh2011} and on two real-world tasks in a kitchen environment: \emph{(1)} \textsc{Emptying a dishwasher}; and \emph{(2)} \textsc{Opening a dishwasher}. We also provide a comprehensive comparison among various \ac{SDF} methods and obstacle avoidance approaches including \acf{MM}, \ac{DT} and artificial potential functions. 
\end{enumerate}
\end{mdframed}

%%%%%%%%%%%%%%%%%%%%%%%%%%%%%%%%%%%%%%%%%%%%%%%%%%%%%%%%%%%%%%%%%%%%%%%%%%%%%%%%%%%%%%%%%%%%%%%%%%%%%%%%%%%
\section{Related Work}
\label{sec:related_work}

\subsection{Learning Contractive Dynamical Systems}
\label{sec:NCDS}
The use of contractive guarantees in learned dynamical systems is a growing trend in robot motion skill learning. In general, contraction guarantees can be introduced into the skill model via three methods: \emph{(i)} As a  regularizer or constraint on the main skill learning objective~\citep{Ravichandar2018,Rezazadeh2022}; \emph{(ii)} As a contraction metric designed to stabilize a non-contractive skill~\citep{sun2020,Tsukamoto2021b}; and \emph{(iii)} By learning a skill model that is contractive by design~\citep{Abyaneh2025,BeikMohammadi2024,Jaffe2024,Beik-Mohammadi2024b}.  
The latter strategy is the one we follow in our framework as it provides stronger theoretical and practical guarantees. We review this set of works in more detail next.

\citet{BeikMohammadi2024} introduced \ac{NCDS}, which inherently embeds the contraction property within a \ac{NN}, thus avoiding regularizers or a separate model for learning a contraction metric. \ac{NCDS} was later extended with regularizers controlling the system's contraction rate, in addition to conditioning on task variables and providing contraction-preserving obstacle avoidance via pullback Riemannian metrics~\cite{Beik-Mohammadi2024b}. Similarly, \citet{Jaffe2024} presented an inherently contractive model, which leverages diffeomorphisms between the data space and a latent space to provide global contraction guarantees. Another approach is followed by~\citet{Abyaneh2025}, who propose learning an inherent contractive dynamical system via recurrent equilibrium networks and coupling layers.
We build upon previous work to learn a contractive dynamical system via \ac{NCDS}. However, our approach can be seamlessly integrated with any contractive dynamical system. Thereby, we address the contraction-preserving obstacle avoidance problem, which has been largely overlooked in previous works except in~\cite{BeikMohammadi2024,Huber2019}.

\subsection{Stability-Preserving Obstacle Avoidance}
\label{sec:obs_avoidance} 
Applying reactive obstacle avoidance in \ac{LfD} settings requires careful consideration to ensure safe and robust robot behavior, particularly when motion skills are encoded as stable dynamical systems. 
To address this, \citet{KhansariZadeh2012} enhanced the Harmonic Potential Function~\cite{Kim1992}, which uses potential flows from fluid mechanics for obstacle avoidance, to ensure that asymptotically-stable dynamical systems maintain stability via an \ac{MM}. Later,~\citet{Huber2019} demonstrated that this approach can also preserve contraction guarantees, even for concave obstacles~\cite{Huber2022, Fourie2024}. While~\citet{Huber2022} primarily relied on simple geometric primitives for obstacle avoidance,~\citet{Fourie2024} employed learned obstacle representations of the entire robot configuration space combined with \ac{MM}. Their work, conducted in parallel to our research, preserves asymptotic stability but did not address contraction stability guarantees.

Building on prior work in learning stable dynamical systems via diffeomorphisms~\citep{NEUMANN2015,Rana2020,Urain2020,Zhang2022},~\citet{Zhi2022} introduced the \ac{DT} method, which defines an obstacle-avoidance diffeomorphism via a flow field. This diffeomorphism transforms the learned dynamics while providing asymptotically-stable obstacle avoidance. Importantly, the \ac{DT} method also accounts for joint-space stable skills using the Moore-Penrose inverse and the robot’s forward kinematics. This work, however, shared a limitation with some of the foregoing approaches: A focus on asymptotic stability rather than the more general contraction property.
In summary, whole-body obstacle avoidance methods that preserve contraction, especially in complex scenes, presents a significant challenge that, to our knowledge, remains unresolved.

\subsection{Implicit Robot Representations}
\label{sec:implicit_robot_representation}
Capturing the robot's spatial structure is essential for enabling effective obstacle avoidance, preventing self-collisions, and facilitating physical interaction.
In this context, implicit representations provide a powerful way to model a robot's shape and volume. For example, \ac{SDF}s provide an implicit, memory-efficient, and computationally fast surface representation. Studies by~\citet{YimlinLi2023} and~\citet{Yiting2023} demonstrate that \ac{SDF}s offer significant advantages over geometric primitives by providing a higher level of detail, facilitating precise interaction.
\looseness=-1

The use of implicit representations for articulated kinematic chains is particularly challenging as the configuration of the robot's structure in the workspace varies as a function of the robot joint position. In this case, each robot link can be modeled by separate \acp{SDF} using architectures such as~\acp{NN}~\cite{Koptev2023} and \ac{BP}~\cite{YimlinLi2023}, or the entire robot body can be represented using a single model~\cite{Liu2022}, all of which yielding smooth surfaces.
Notably, \citet{Yiming2024} observe that learning the Euclidean distance between the robot surface and a workspace query point is highly non-linear in configuration space. To mitigate this, they propose computing the distance in configuration space, leading to the \ac{CDF} method.
While~\citet{Yiming2024} and \citet{Marticorena2024} made obstacle avoidance for point-to-point motion possible using \acp{SDF} and a QP solver, no existing approaches, to the best of our knowledge, perform contraction-preserving obstacle avoidance with \acp{SDF}.

%%%%%%%%%%%%%%%%%%%%%%%%%%%%%%%%%%%%%%%%%%%%%%%%%%%%%%%%%%%%%%%%%%%%%%%%%%%%%%%%%%%%%%%%%%%%%%%%%%%%%%%%%%%
\section{Background}
\label{sec:background}

\subsection{Contraction Stability}
\label{sec:Contraction}
We are interested in learning contractive robot motion skills, as contraction provides a rigorous and general stability certificate for robust generalization. 
Intuitively, contraction stability studies the evolution of two trajectories of a dynamical system given different initial conditions~\citep{LOHMILLER1998}. 
In other words, contraction implies that a dynamical system ``forgets'' its initial conditions rapidly, with the distance between any two trajectories decreasing exponentially over time.
Formally, consider two trajectories $\tau_i$ and $\tau_j$ generated by a dynamical system of the form $\dot{\mathbf{x}}_t = f(\mathbf{x}_t)$, where $\mathbf{x}_t \in \mathbb{R}^D$ is the system state, $f\colon \mathbb{R}^D \rightarrow \mathbb{R}^D$, and $\dot{\mathbf{x}}_t = \nicefrac{\mathrm{d}\mathbf{x}}{\mathrm{d}t}$. Using their virtual displacement $\delta \mathbf{x} = \tau_j - \tau_i$, the squared distance between the trajectories corresponds to $\delta \mathbf{x}^\top \delta \mathbf{x}$~\cite{LOHMILLER1998}. Contraction theory cares about the evolution of this distance, i.e., its rate of change,\looseness=-1
\begin{equation}
    \frac{\dif}{\dif t}(\delta \mathbf{x}^\top \delta \mathbf{x}) = 2\delta \mathbf{x}^\top \delta \dot{\mathbf{x}} = 2 \delta \mathbf{x}^\top \mathbf{J}_f(\mathbf{x}) \delta \mathbf{x} ,
    \label{eq:ContractionRateOfChange}
\end{equation}
where $\delta \dot{\mathbf{x}} = \mathbf{J}_f(\mathbf{x}) \delta \mathbf{x}$, with $\mathbf{J}_f(\mathbf{x}) = \nicefrac{\partial f}{\partial \mathbf{x}}$.
\citet{LOHMILLER1998} show that a dynamical system exhibits contraction if the largest eigenvalue $\lambda_{\text{max}}$ of the symmetric part of its Jacobian is uniformly and strictly negative. Therefore,
\begin{equation}
    \frac{\dif}{\dif t}(\delta \mathbf{x}^\top \delta \mathbf{x}) \leq 2 \lambda_{\text{max}} \delta \mathbf{x}^\top \delta \mathbf{x} \implies \lVert \delta \mathbf{x} \rVert \leq \lVert \delta \mathbf{x}_0 \rVert e^{-\alpha t} ,
\end{equation}
which means that the virtual displacement decreases exponentially to zero with contraction rate $\alpha$ \cite{LOHMILLER1998}.

\begin{definition}[Contraction stability~\cite{LOHMILLER1998}]
    \label{def:contraction_stability}
    A dynamical system $\dot{\mathbf{x}}_t = f(\mathbf{x}_t)$ is contractive if its Jacobian $\mathbf{J}_f = \frac{\partial f}{\partial \mathbf{x}}$ is uniformly negative definite for all $\mathbf{x} \in \mathbb{R}^n$ and at all times $t \geq 0$. This means that,

    \begin{equation}
        \exists \alpha > 0, \ \forall \mathbf{x}, \ \forall t \geq 0, \ \frac{1}{2} \left( \frac{\partial f}{\partial \mathbf{x}} + \left( \frac{\partial f}{\partial \mathbf{x}} \right)^\top \right) \preceq -\alpha \mathbf{I} \prec \mathbf{0}.
    \end{equation}
\end{definition}
Definition~\ref{def:contraction_stability} can be generalized to account for a more general distance definition, namely $\delta \mathbf{x}^\top \mathbf{G}(\mathbf{x}) \delta \mathbf{x}$, where $\mathbf{G}(\mathbf{x})$ is a Riemannian metric~\citep{Lee1997,SimpsonPorco2014}. 
This also allows us to analyze the contractive behavior of a dynamical system under a coordinate change via a smooth diffeomorphism $\mathbf{y} = \psi(\mathbf{x})$. The corresponding differential change in coordinates is given by $\delta_\mathbf{y} = \mathbf{J}_\psi \delta_\mathbf{x}$, where $\mathbf{J}_\psi$ is the Jacobian of the diffeomorphism $\psi$. \citet{Manchester2017} demonstrated that contraction is invariant under such coordinate changes.

\begin{mdframed}[hidealllines=true,backgroundcolor=gray!8,innerleftmargin=.2cm,
  innerrightmargin=.2cm]
\begin{theorem}[Invariance under coordinate change~\cite{Manchester2017}]
\label{theorem:Invariance_Under_Coordinate_Change}
If Definition~\ref{def:contraction_stability} is satisfied for a dynamical system, then it is preserved under the following transformations:
\begin{enumerate}
    \item Affine feedback transformations $u(\mathbf{x}, \mathbf{v}) = \mathbf{a}(\mathbf{x}) + \mathbf{B}(\mathbf{x})\mathbf{v}$, with $\mathbf{B}(\mathbf{x})$ being a smooth nonsingular $n \times n$ matrix function;
    \item Differential coordinate changes $\delta_\mathbf{y} = \mathbf{J}_\psi(\mathbf{x})\delta_\mathbf{x}$, where $\mathbf{J}_\psi(\mathbf{x})$ is a nonsingular matrix for all $\mathbf{x}$, and induces a new contraction metric,
    \begin{equation}
        \mathbf{G}_\mathbf{y}(\mathbf{x}) := \mathbf{J}_\psi(\mathbf{x})^{\top} \mathbf{G}(\mathbf{x}) \mathbf{J}_\psi(\mathbf{x});
    \end{equation}
    \item Coordinate changes $\mathbf{y} = \psi(\mathbf{x})$, where $\psi$ is a  diffeomorphism, with  corresponding contraction metric $\mathbf{G}_\mathbf{y}$.
\end{enumerate}
\end{theorem}
\end{mdframed}

Given a contraction metric as stated in Theorem~\ref{theorem:Invariance_Under_Coordinate_Change}, we can also determine whether a given system is contractive by following the results in~\cite{Tsukamoto2021}.

\begin{mdframed}[hidealllines=true,backgroundcolor=gray!8,innerleftmargin=.2cm,
  innerrightmargin=.2cm]
\begin{theorem}[Contraction conditions~\cite{Tsukamoto2021}]
\label{theorem:Contraction_Conditions}
Assume that a uniformly positive definite matrix $\mathbf{G}(\mathbf{x}, t) = \mathbf{J}_\psi(\mathbf{x})^{\top} \mathbf{J}_\psi(\mathbf{x}) \succ \mathbf{0}$ exists, $\forall \mathbf{x}, t$, where $\mathbf{J}_\psi(\mathbf{x})$ defines a smooth coordinate transformation of the differential $\delta \mathbf{x}$, i.e., $\delta \mathbf{y} = \mathbf{J}_\psi(\mathbf{x}) \delta \mathbf{x}$. The following equivalent condition holds for $\alpha \in \mathbb{R}_{>0}$ and $\forall \mathbf{x}, t$,
\begin{equation}
    \dot{\mathbf{G}} + \mathbf{G} \frac{\partial f}{\partial \mathbf{x}} + \frac{\partial f}{\partial \mathbf{x}}^{\top} \mathbf{G} \leq -2 \alpha \mathbf{G},
\end{equation}
then all solution trajectories of the system converge exponentially fast to a single trajectory, irrespective of their initial conditions, with an exponential convergence (contraction) rate $\alpha$.
\end{theorem}
\end{mdframed}

\subsection{Flow-based Diffeomorphisms}
\label{sec:diffeomorphic_mapping} 
As contraction is preserved under a change of coordinates, we leverage this to design obstacle-avoidance behaviors based on diffeomorphic mappings, which we revise next. A diffeomorphism $\psi\colon \mathcal{Y} \rightarrow \mathcal{X}$ is a smooth, bijective map with a smooth inverse, thereby providing a coordinate transformation between two differentiable manifolds $\mathcal{Y}$ and $\mathcal{X}$. According to~\citet[Theorem 9.12]{Lee2012}, any such diffeomorphism can be realized as a flow generated by an infinitesimal generator $\mathbf{V}$, often represented as a vector field on a smooth manifold.
Specifically, let $\mathbf{V}\colon \mathbb{R}^d \to \mathbb{R}^d$ be a time-independent vector field and the flow $\gamma\colon \mathbb{R} \times \mathbb{R}^d \to \mathbb{R}^d$ be defined by,
\begin{equation} 
    \label{eq:DT_flow_forward} 
    \gamma(t, \mathbf{y}) = \mathbf{y} + \int_0^t \mathbf{V}(\gamma(u, \mathbf{y})) \, du =  \mathbf{x}.
\end{equation}

This flow $\gamma(t,\mathbf{y})$ provides the position $\mathbf{x}$ at time $t$ of a trajectory starting at $\mathbf{y}$ when $t=0$. For each fixed time $t$, this flow defines a diffeomorphism $\psi\colon \mathcal{Y} \to \mathcal{X}$ by $\psi(\mathbf{y}) := \gamma(t, \mathbf{y})$. Therefore, the flow defines an invertible mapping, whose inverse can be computed by reversing the direction of time,
\begin{equation} 
    \label{eq:DT_flow_inverse} 
    \gamma(-t, \mathbf{x}) = \mathbf{x} + \int_{-t}^0 \mathbf{V}(\gamma(u, \mathbf{x})) \, du = \mathbf{y}.
\end{equation}
Note that this flow-based diffeomorphism $\psi(\mathbf{y}) := \gamma(t, \mathbf{y})$ maps the initial point $\mathbf{y} \in \mathcal{Y}$ to the point $\mathbf{x} = \gamma(t, \mathbf{y})\in \mathcal{X}$. Furthermore, given a vector field $f\colon \mathcal{X} \to \mathcal{T}\mathcal{X}$, where $f(\mathbf{x})$ assigns a tangent vector in $\mathcal{T}_\mathbf{x}\mathcal{X}$ to each point $\mathbf{x} \in \mathcal{X}$, we can use $\psi$ to pullback $f$ to a vector field on $\mathcal{Y}$. Specifically, let $\mathbf{J}_\psi$ be the Jacobian of $\psi$, then the pullback of $f$ via $\psi$ is,
\begin{equation} 
    \label{eq:pullback} 
    \dot{\mathbf{y}} = \mathbf{J}_\psi^{-1}f(\psi(\mathbf{y})),
\end{equation}
thereby transforming tangent vectors on $\mathcal{X}$ to corresponding tangent vectors on $\mathcal{Y}$~\cite{Lee2012}.

\subsection{Signed Distance Fields (SDFs)}
\label{sec:sdf}
\begin{SCfigure}[0.6][h]
    \caption{Illustration of a 2D \ac{SDF}: Contour lines of an star-shaped \ac{SDF} with the distance $\Gamma_\text{SDF}$ to the obstacle surface, depicted as a solid red line.}
    \includegraphics[width=0.3\textwidth]{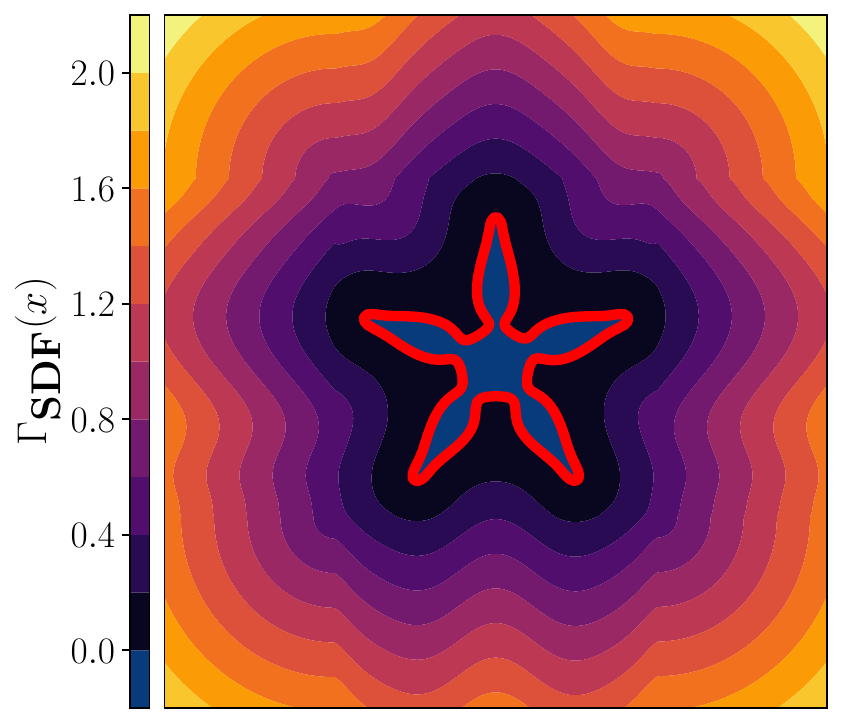}
    \label{fig:sdf_illustration}
\end{SCfigure}

Representing objects as meshes or point clouds is often inefficient due to high memory requirements and redundant information~\cite{Park2019}. Moreover, these discrete representations lack smoothness and differentiability as they approximate surfaces with sampled points or polygons.
To overcome these limitations, \ac{SDF}s offer a compact, smooth, and differentiable representation. An \ac{SDF} defined by $\Gamma_\text{SDF}\colon \mathbb{R}^d \to \mathbb{R}$, encodes the objects surface through the minimum distance $d_\text{min}$ from a point $\mathbf{x} \in \mathbb{R}^d$ to the surface using a smooth function~\cite{Park2019}. Three regions can be identified: Free space $\mathcal{H}^f = \{ \mathbf{x} \in \mathbb{R}^d \mid \Gamma_\text{SDF}(\mathbf{x}) > 0 \}$, surface boundary $\mathcal{H}^s = \{ \mathbf{x} \in \mathbb{R}^d \mid \Gamma_\text{SDF}(\mathbf{x}) = 0 \}$, and interior $\mathcal{H}^i = \{ \mathbf{x} \in \mathbb{R}^d \mid \Gamma_\text{SDF}(\mathbf{x}) < 0 \}$ (see Fig.~\ref{fig:sdf_illustration} for an illustration).

Note that this approach decouples spatial resolution from memory usage, enabling high-resolution reconstructions with fast inference~\cite{Maric2024}. \citet{Park2019} propose using a \ac{MLP} to model the \ac{SDF}, offering a lightweight and efficient model. Additionally, piecewise polynomial representations enable continuous surface modeling~\cite{Pujol2023}. \citet{Maric2024} extend this by employing \ac{BP} basis functions which enables smooth surfaces that can be incrementally refined.

\section{Contraction Preserving Obstacle Avoidance}
\label{sec:methodology}
Here we introduce our contraction-preserving obstacle avoidance method. 
First, we explain the contractive dynamical system used in this paper. Later, we describe the contraction-preserving obstacle avoidance method. 

\subsection{Learned Contractive Dynamical System}
Consider a contractive dynamical system $\dot{\mathbf{x}} = f_\text{NCDS}(\mathbf{x})$, 
where $\dot{\mathbf{x}}$ represents the first-order time derivative of the system state $\mathbf{x}$.
From Theorem \ref{theorem:Contraction_Conditions}, we know that if the symmetric part of the system dynamics Jacobian $\mathbf{J}_{f_\text{NCDS}}$ is negative definite, our dynamics is contractive.~\citet{BeikMohammadi2024} propose directly learning this Jacobian to intrinsically embed the contraction property into $f_\text{NCDS}$. Specifically, the Jacobian is modeled with parameters $\boldsymbol{\theta}$ as,
\begin{equation}
    \label{eqn:J_NCDS}
    \hat{\mathbf{J}}_{f_\text{NCDS}}(\mathbf{x}) = -(\mathbf{J}_{\boldsymbol{\theta}}(\mathbf{x})^\top \mathbf{J}_{\boldsymbol{\theta}}(\mathbf{x}) + \boldsymbol{\epsilon}),
\end{equation}
where $\mathbf{J}_{\boldsymbol{\theta}}(\mathbf{x})\colon \mathbb{R}^D \to \mathbb{R}^D$ represents the square root of the Jacobian, thus ensuring symmetry. The small positive vector $\boldsymbol{\epsilon} > \mathbf{0}$ guarantees that the eigenvalues are bounded by $-\boldsymbol{\epsilon}$~\cite{BeikMohammadi2024}.
Given the negative-definite Jacobian~\eqref{eqn:J_NCDS}, the contractive dynamics is obtained via the calculus for line integrals~\cite{Lorraine2019},
\begin{equation}
\begin{split}
    \label{eqn:f_NCDS}
    \dot{\mathbf{x}} &= f_\text{NCDS}(\mathbf{x}) \\
    &= \dot{\mathbf{x}}_0 + \int_0^{1} \hat{\mathbf{J}}_{f_\text{NCDS}}\big(c(\mathbf{x}, t, \mathbf{x}_0)\big) \dot{c}(\mathbf{x}, t, \mathbf{x}_0) \dif t,
\end{split}
\end{equation}
where,
\begin{equation}
    c(\mathbf{x}, t, \mathbf{x}_0) = (1-t)\mathbf{x}_0 + t\mathbf{x}, \quad \dot{c}(\mathbf{x}, t, \mathbf{x}_0) = \mathbf{x} - \mathbf{x}_0,
\end{equation}
and $\mathbf{x}_0$ is the initial system state~\cite{BeikMohammadi2024}.
The velocity $\dot{\mathbf{x}}_t$ is computed using a numerical integral solver given $\mathbf{x}_t$~\cite{BeikMohammadi2024,Chen2018}.

When learning robot skills via \ac{NCDS}, we have two options: Learning the \ac{NCDS} in joint space $\mathcal{C}$ or instead, encoding the skill in task space $\mathcal{X}$.
In the latter case, \ac{NCDS} uses Lie groups to represent the end-effector orientation in $\operatorname{SO}(3)$~\cite{BeikMohammadi2024}, and employs its Lie algebra $\mathfrak{so}(3)$ to obtain skew-symmetric matrices $\mathbf{\hat{w}} \in \mathfrak{so}(3)$~\cite{Humphreys1972}. The mapping between $\operatorname{SO}(3)$ and $\mathfrak{so}(3)$ is defined via the exponential and the logarithmic map.

\begin{SCfigure}[0.7][t]
\caption{Exemplary \ac{NCDS} model of a skill learned in 2D space, with demonstrated trajectories from the LASA dataset in black and the integrated motion over the \ac{NCDS} model in blue. The gray arrows represent the \ac{NCDS} vector field.}
\includegraphics[width=0.25\textwidth]{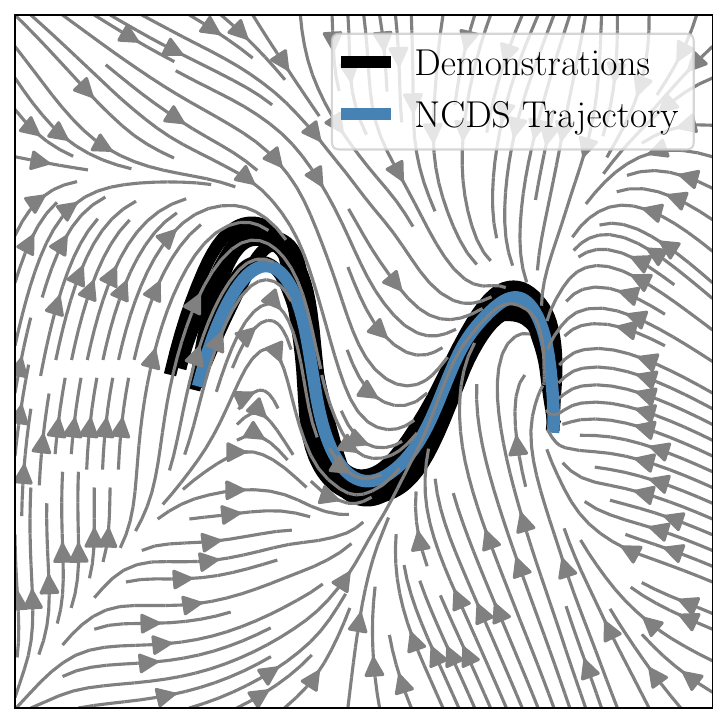}
\label{fig:contractive_sys}
\end{SCfigure}

To train a \ac{NCDS} model, the loss function minimizes the average reconstruction error between the true next state $\mathbf{x}_{t+1}$ and the predicted next state $\hat{\mathbf{x}}_{t+1} = \mathbf{x}_t + \hat{\dot{\mathbf{x}}}_{t+1}$~\cite{BeikMohammadi2024},
\begin{equation}
    \label{eq:NCDS_loss}
    \mathcal{L}_\text{Jac} = \frac{1}{T_n} \sum_{t = 0}^{T_n-1} \Vert \mathbf{x}_{t+1} - (\mathbf{x}_t + \hat{\dot{\mathbf{x}}}_{t+1}) \Vert^2.
\end{equation}
In addition, a state-independent regularizer is added to optimize the eigenvalue bound $\boldsymbol{\epsilon} = \left[ \epsilon_1, \cdots, \epsilon_D \right]$ as follows~\cite{Beik-Mohammadi2024b},
\begin{equation}
    \label{eq:NCDS_loss_state_independant}
    \mathcal{L}_{\epsilon} = - \sum^D_{n=2} \vert \epsilon_1 -\epsilon_n  \vert^2.
\end{equation}
To enhance the local contraction properties of \ac{NCDS}, we propose an additional noise-injected reconstruction loss,
\begin{equation}
    \label{eq:NCDS_loss_noise}
\mathcal{L}_{\text{noise}} = -  \frac{1}{T_n} \sum_{t = 0}^{T_n-1} \left\| \mathbf{x}_{t+1} - \left( \tilde{\mathbf{x}}_t + \hat{\dot{\mathbf{x}}}_t \right) \right\|^2, 
\end{equation}
which acts orthogonally to the direction of motion $\dot{\mathbf{x}}$, 
with $\tilde{\mathbf{x}}_t \sim \mathcal{N}(\mathbf{x}_t, \mathbf{\Sigma}_{\perp \mathbf{\dot{x}}_t})$, representing noise injected in the directions perpendicular to the movement. Our regularizer encourages \ac{NCDS} to be robust against orthogonal perturbations, thereby implicitly improving its local contractive behavior.
The total loss is $\mathcal{L}=\mathcal{L}_\text{Jac}+ \beta_\epsilon \mathcal{L}_\epsilon+\beta_\text{noise}\mathcal{L}_\text{noise}$, with $\beta_\epsilon$ and $\beta_\text{noise}$ being weights balancing the losses influence. \ac{NCDS} training is summarized in Algorithm~\ref{alg:NCDS_train}.
Additionally, Fig.~\ref{fig:contractive_sys} illustrates a vector field learned by \ac{NCDS}.

\begin{algorithm}[t]
\caption{NCDS Training in Joint Space}\label{alg:NCDS_train}
\SetKwInOut{Input}{Input}
\SetKwInOut{Output}{Output}

\Input{Demonstrations: $\tau=\{ \mathbf{q}_{n,t}\}_{n=1,t=0}^{N,T_n-1}$,\\
 where $n \in [1,N]$, $t \in [1,T_n]$, and\\
 initial parameters $\boldsymbol{\theta}$, initial regularizer $\boldsymbol{\epsilon}$}
\Output{Learned Jacobian network parameters $\boldsymbol{\theta}$,\\
learned regularizer $\boldsymbol{\epsilon}$}
 \While{not converged}{
    \For{each demonstration $n \in N$}{
        \For{each time step $t \in T_n$}{
        Calculate $\hat{\mathbf{J}}_{f_\text{NCDS}}$ of the \ac{NCDS} with Eq.~\eqref{eqn:J_NCDS} \\
        Define the contractive dynamics $f_\text{NCDS}$ with Eq.~\eqref{eqn:f_NCDS}
        }
    }
    Train the Jacobian network by solving:
    $\boldsymbol{\theta}^*,\boldsymbol{\epsilon}^* = \text{argmin}_{\boldsymbol{\theta},\boldsymbol{\epsilon}} \left(\mathcal{L}_\text{Jac} + \beta_\epsilon \mathcal{L}_\epsilon+\beta_\text{noise}\mathcal{L}_\text{noise} \right)$
}
\end{algorithm}

\subsection{SDF-based Diffeomorphic Transform}
Our obstacle avoidance method consists of two components: An intrinsic representation of the robot surface and a contraction-preserving transformation of a contractive vector field $f_c$. The latter can be: \emph{(1)} A learned joint-space \ac{NCDS} $f_\text{NCDS}(\mathbf{q})$; \emph{(2)} A learned task-space \ac{NCDS} $f_\text{NCDS}(\mathbf{x})$ mapped to joint space; and \emph{(3)} Any differentiable contractive dynamical system in joint space.
Next, we describe the implicit representation of the robot surface, which provides the basis for deriving the infinitesimal generator for obstacle avoidance.

\subsubsection{Implicit Robot Representations}
\label{implicit_representations}
\begin{SCfigure}[0.5][t]
    \caption{Illustration of a learned RDF showing the contour lines of the robot's \ac{SDF} for a Franka-Emika Panda robot grasping a plate. The contours indicate the distance $\Gamma_\text{RDF}$ to the robot's surface. The inner gray mesh shows the learned robot surface $\mathcal{R}$.}
    \includegraphics[width=0.3\textwidth]{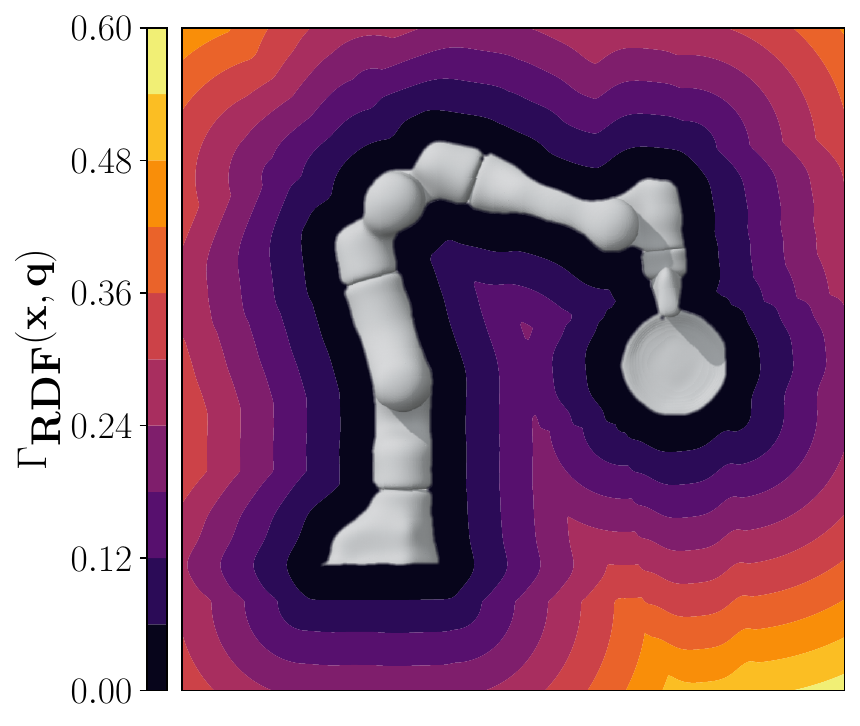}
    \label{fig:RDF}
\end{SCfigure}

To achieve obstacle avoidance, one must know the minimal distance $d_{\min}$ between the scene surface $\mathcal{S}$ and the robot surface $\mathcal{R}$. Therefore we start by learning an implicit representation of $d_{\min}$. There are two ways to achieve this: \emph{(1)} A learned \ac{SDF} of the robot and; \emph{(2)} A learned \ac{SDF} of the scene.
To learn a \ac{SDF} for the scene, we refer the reader to the established approaches in~\cite{Gropp2020,Ortiz2022,Park2019}.
On the other hand, the distance w.r.t the robot surface can be defined by $d_\text{min} = \Gamma_{\text{RDF}}(\mathbf{x}, \mathbf{q})$, given a query point $\mathbf{x}$ and the robot state $\mathbf{q}\in \mathcal{C}$ \cite{Yiting2023,Koptev2023,YimlinLi2023}. This implicit robot representation is called \ac{RDF}. 
We follow the approach from \citet{YimlinLi2023} and define the \ac{RDF} as,
\begin{equation}
\label{eq:RDF}
    \Gamma_\text{RDF}=\min(\Gamma_{\Omega^r_1}, \Gamma_{\Omega^r_2},...,\Gamma_{\Omega^r_K}),
\end{equation}
with $\Gamma_{\Omega^r_k}\ ,k=1,...,K$, and $\Gamma_\text{RDF}$ defining the \ac{SDF} of $K$ links in the robot base frame $r$. Thereby, each link is transformed via the robot's kinematic chain $^r\mathbf{T}_k(\mathbf{q})\in \mathbb{SE}(3)$ from the frame of the $k$-th link to the base frame, resulting in $\Gamma_{\Omega^r_k}(\mathbf{x}, \mathbf{q})~=~\Gamma_{\Omega^r_k}\big(^r\mathbf{T}_k(\mathbf{q})\mathbf{x} \big)$~\cite{YimlinLi2023}. Given the \ac{SDF} $\Gamma_{\Omega_o}$ (e.g., a grasped object) and its corresponding transformation $^r\mathbf{T}_o(\mathbf{q})\in \mathbb{SE}(3)$ to the robot base frame, we can easily extend the \ac{RDF} by adding the transformed $\Gamma_{\Omega^r_o}$ to the $\min$ operation in Equation~\eqref{eq:RDF}.
We choose \ac{BP} to learn the \ac{SDF} $\Gamma_{\Omega^r_k}$ of each link $k$ as it provides a particularly smooth surface. However, any other \ac{SDF} method can be used. An exemplary learned \ac{RDF} with a grasped plate is shown in Fig.~\ref{fig:RDF}.

Alternatively,~\citet{Yiming2024} propose learning joint-level distances rather than Euclidean point-to-surface distances, leading to the \ac{CDF} with  corresponding implicit distance function $\Gamma_{\text{CDF}}$. 
The \ac{CDF} measures the distance in radians, corresponding to the movement of
joint angles towards the zero-level configuration set $\mathbf{q}^{*}$, resulting in~\cite{Yiming2024},
\begin{equation}
    \Gamma_{\text{CDF}}(\mathbf{x},\mathbf{q})=\min_{\mathbf{q}^{*}}(\mathbf{q} - \mathbf{q}^{*}).
\end{equation}
\ac{CDF} solves the inverse kinematics problem in a single step by updating the current robot joint configuration along the \ac{CDF} gradient toward the zero-level configuration set $\mathbf{q}^{*}$ via,
\begin{equation}
    \mathbf{q}^{*} = \mathbf{q} - \Gamma_{\text{CDF}}(\mathbf{x},\mathbf{q})\nabla_\mathbf{q}\Gamma_{\text{CDF}}(\mathbf{x},\mathbf{q}).
\end{equation}
However, a limitation of this approach is that extending the \ac{CDF} with an object's \ac{SDF}, such as that of a grasped object, is not as straightforward as in the \ac{RDF} framework. Accomplishing this would require training a new model or incorporating additional conditioning into the existing model.
In this work, both \ac{RDF} and \ac{CDF} are employed and compared.
In the following, we use the term $\Gamma_{\text{SDF}}(\mathbf{x},\mathbf{q})$ to denote a general implicit distance function between the scene and the robot. When referring specifically to the robot's implicit distance function, we denote it by $\Gamma_{\mathcal{R}}$. Analogously, $\Gamma_{\mathcal{S}}$ is used for the implicit distance function of the scene.

\subsubsection{Implicit Infinitesimal Generator}
\label{implicit_infinitesimal_generator}
\begin{SCfigure}[0.5][t]
    \caption{Vector field of a learned star-shaped \ac{SDF} function with inverse barrier function $b_\text{inv}(\mathbf{x}, \mathbf{q})$. The contour lines represent the barrier's magnitude. The solid red line shows the surface of the object and the dashed grey line the safety threshold $t_{\text{save}}$.}
    \includegraphics[width=0.32\textwidth]{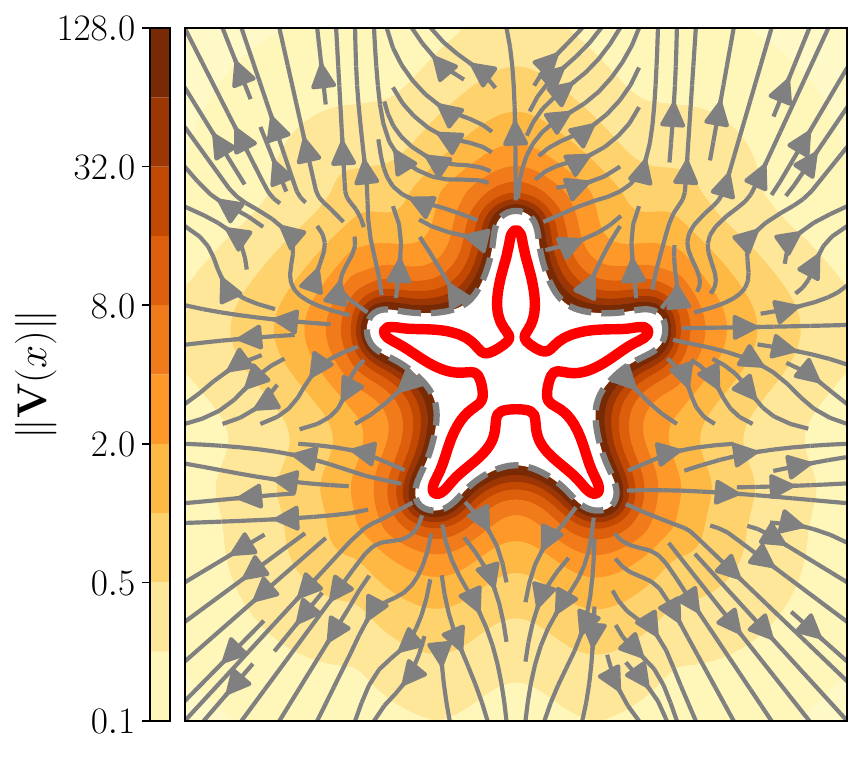}
    \label{fig:V_o_SDF}
\end{SCfigure}

For obstacle avoidance, the gradient $\nabla_\mathbf{q}\Gamma_{\text{SDF}}$ determines how the robot joints should move away from obstacles. Thus, this gradient acts as an \textit{infinitesimal generator} $\mathbf{V}$. To ensure strict collision avoidance, we use an inverse barrier function $b_\text{inv}(\mathbf{x}, \mathbf{q})$, similarly to~\cite{Klein2023}. This barrier is defined as follows,
\begin{equation}
    \label{eq:barrier_inv}
    b_\text{inv}(\mathbf{x}, \mathbf{q}) = \frac{1}{\Gamma_{\text{SDF}}(\mathbf{x}, \mathbf{q})},
\end{equation}
which is combined with the gradient, leading to the infinitesimal generator $\mathbf{V} = b_\text{inv}\nabla_\mathbf{q}\Gamma_{\text{SDF}}$.
As $\Gamma_{\text{SDF}}\to 0$, then $b_\text{inv}\to \infty$, ensuring collision avoidance. When $\Gamma_{\text{SDF}}\to\infty$, then $b_\text{inv}\to 0$, canceling the effect of the obstacle avoidance gradient $\nabla_\mathbf{q}\Gamma_{\text{SDF}}$.
The introduction of this barrier function is motivated by the fact that \ac{SDF}s are often learned via the Eikonal loss, enforcing $\|\nabla_\mathbf{x}\Gamma_{\text{SDF}}(\mathbf{x}, \mathbf{q})\|=1$~\cite{Park2019}. Thus, gradients remain non-negligible even far away from the obstacles.
Figure~\ref{fig:V_o_SDF} illustrates the resulting vector field for a geometric primitive following the aforementioned approach.

We can make the barrier function more general by adding tunable parameters like a safety threshold $t_\text{save}$ and a scaling factor $s_\text{grad}$ as follows,
\begin{equation}
\label{eq:barrier_advanced}
    b_\text{inv}(\mathbf{x}, \mathbf{q})=\frac{s_\text{grad}}{\Gamma_\text{SDF}\left(\mathbf{x}, \mathbf{q}\right)-t_\text{save}}.
\end{equation}
Moreover, we also propose to leverage \textit{swept features}, aimed at weighting movements towards and tangential to the obstacle while ignoring those away from it, using the dot product $\dot{\mathbf{q}}_{\text{NCDS}} \!\cdot\!\nabla_\mathbf{q}\Gamma_{\text{SDF}}$. To do so, we need to distinguish whether the \ac{SDF} of the scene $\Gamma_{\mathcal{S}}$ or of the robot $\Gamma_{\mathcal{R}}$ is used, since the corresponding velocity fields and gradient direction may differ. 
Specifically, if $\Gamma_{\mathcal{R}}$ is employed, then the gradient of $\Gamma_{\mathcal{R}}$ and $f_\text{c}$ point in the same direction when moving towards the obstacle and point in opposite directions when moving away from it. If $\Gamma_{\mathcal{S}}$ is used, this phenomenon is the opposite.
For the implicit distance function of the robot and the scene, swept-augmented barriers are defined as,
\begin{align}
\label{eq:barrier_swept_r}
   b_{\text{swept},\mathcal{R}}\left(\mathbf{x}, \mathbf{q}\right)&=\frac{1}{2}\left(1 + \frac{\dot{\mathbf{q}}_\text{NCDS} \cdot \nabla_\mathbf{q}\Gamma_\mathcal{R}(\mathbf{x}, \mathbf{q})}{\vert \dot{\mathbf{q}}_\text{NCDS} \vert \ \vert \nabla_\mathbf{q}\Gamma_\mathcal{R}(\mathbf{x}, \mathbf{q}) \vert} \right),\\
    \label{eq:barrier_swept_s}
   b_{\text{swept},\mathcal{S}}\left(\mathbf{x}, \mathbf{q}\right)&=\frac{1}{2}\left(1 - \frac{\dot{\mathbf{q}}_\text{NCDS} \cdot \nabla_\mathbf{q}\Gamma_\mathcal{S}(\mathbf{x}, \mathbf{q})}{\vert \dot{\mathbf{q}}_\text{NCDS} \vert \ \vert \nabla_\mathbf{q}\Gamma_\mathcal{S}(\mathbf{x}, \mathbf{q}) \vert} \right).
\end{align}
Although the foregoing barrier functions can be used independently, it is possible to leverage the properties of both barrier functions~\eqref{eq:barrier_inv} and~\eqref{eq:barrier_swept_r} or~\eqref{eq:barrier_swept_s} by simply defining $b(\mathbf{x}, \mathbf{q})=b_\text{inv}(\mathbf{x}, \mathbf{q})b_\text{swept}(\mathbf{x}, \mathbf{q})$. 
Given the barrier function $b(\mathbf{x}, \mathbf{q})$, the infinitesimal generator results in,
\begin{equation}
    \label{eq:V_sdt}
    \mathbf{V}(\mathbf{x}, \mathbf{q}) = - b(\mathbf{x}, \mathbf{q})\nabla_\mathbf{q}\Gamma_{\text{SDF}}(\mathbf{x}, \mathbf{q}).
\end{equation}

\subsubsection{Contraction-Preserving Differential Coordinate Change}
\label{sdf_diffeomorphic_transform}
Given the infinitesimal generator $\mathbf{V}$~\eqref{eq:V_sdt} obtained from the \ac{SDF} introduced in Section~\ref{implicit_infinitesimal_generator}, we know the joint directions that locally maximize obstacle avoidance. Directly forcing the robot away from the obstacle via,
\begin{equation}
    \label{eq:naive_obs_avoidance}
    \mathbf{\dot q} = \hat{f}_\text{c}(\mathbf{q}) = f_\text{c}(\mathbf{q}) - b(\mathbf{x}, \mathbf{q}) \nabla_\mathbf{q}\Gamma_\text{SDF}(\mathbf{x}, \mathbf{q}),
\end{equation}
as proposed by~\citet{Ravichandar2015}, could violate the contraction conditions of Theorem~\ref{theorem:Contraction_Conditions}. 
This is because the $b(\mathbf{x}, \mathbf{q})\nabla_\mathbf{q}\Gamma_\text{SDF}(\mathbf{x}, \mathbf{q})$ term is not guaranteed to be negative definite, potentially leading to unstable behaviors.

To achieve obstacle avoidance without breaking contraction stability, we propose reshaping the contractive flow $f_c$ around the obstacle by defining a mapping to a new obstacle-free manifold $\mathcal{Y}$ that preserves contraction. To do this, we leverage Theorem~\ref{theorem:Invariance_Under_Coordinate_Change}, which tells us that contraction is preserved under coordinate transformations or diffeomorphisms. Thus, we cast the contraction-preserving obstacle avoidance problem as finding a diffeomorphism that reshapes our contractive dynamical system $f_c$ around the obstacle as a function of the gradient of its implicit representation $\nabla_\mathbf{q}\Gamma_{\text{SDF}}$.
As explained in Section~\ref{sec:diffeomorphic_mapping}, a diffeomorphism $\psi$ can be designed from a flow. If such a flow is differentiable, it is thus a valid diffeomorphism. 

We propose to leverage the flow generated by the infinitesimal generator~\eqref{eq:V_sdt}. Specifically, using the flow $\psi\colon \mathcal{Y}\to{\mathcal{C}}$ as defined by Equation~\eqref{eq:DT_flow_forward}, we transform the obstacle-free manifold $\mathbf{y}\in \mathcal{Y}$ into the configuration space $\mathbf{q}\in \mathcal{C}$ as follows,

\begin{equation}
    \label{eq:sdt_flow_forward}
    \begin{split}
    \psi(\mathbf{x},\mathbf{y})&=\gamma(\mathbf{x},\mathbf{y},t)= \mathbf{q} ,\\
    &=\mathbf{y}-\int_0^t b\big(\mathbf{x},\gamma(\mathbf{x},\mathbf{y},u)\big)\nabla_\mathbf{y}\Gamma_\text{SDF}\big(\mathbf{x},\gamma(\mathbf{x},\mathbf{y},u)\big)du .
    \end{split}
\end{equation}
Similarly, the corresponding inverse flow is,

\begin{equation}
    \label{eq:sdt_flow_inverse}
    \begin{split}
    \psi(\mathbf{x},{\mathbf{q}})^{-1}&=\gamma(\mathbf{x},{\mathbf{q}},-t)^{-1} = \mathbf{y} ,\\
    ={\mathbf{q}}-\int_{-t}^0 b&\big(\mathbf{x},\gamma(\mathbf{x},{\mathbf{q}},u)^{-1}\big)\nabla_{\mathbf{q}}\Gamma_\text{SDF}\big(\mathbf{x},\gamma(\mathbf{x},{\mathbf{q}},u)^{-1}\big)du.
    \end{split}
\end{equation}

Given the diffeomorphism~\eqref{eq:sdt_flow_forward}, we can carry out a differential coordinate change,
\begin{equation}
    \label{eq:SDDC}
    \delta\mathbf{y}=\mathbf{J_{\psi}}^{-1}\delta\mathbf{q} ,
\end{equation}
using the Jacobian of the flow $\mathbf{J}_\psi$. We refer to this transformation as \textit{\ac{SDDC}}. Based on Theorem~\ref{theorem:Invariance_Under_Coordinate_Change}, which states that contraction is preserved under differential coordinate transformations, we conclude that the transformed system $f_\text{SDDC}$ is contractive. This therefore ensures that the robot can successfully avoid obstacles while preserving the stability of the underlying skill.

\subsubsection{Contraction-Preserving Diffeomorphic Transform}
\begin{figure*}[t]
     \centering
     \begin{subfigure}[b]{0.29\textwidth}
         \centering
         \includegraphics[width=\textwidth]{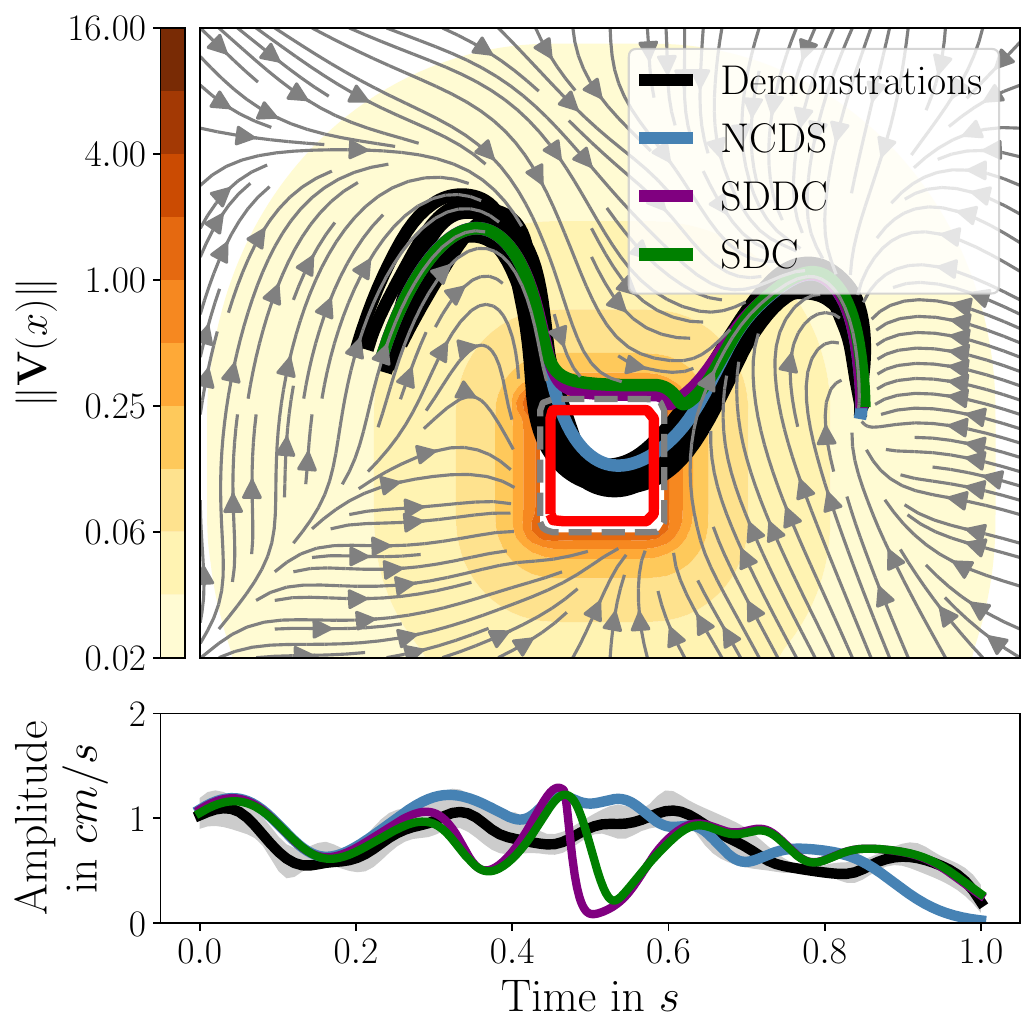}
         \caption{Without friction}
         \label{fig:SDT_no_friction}
     \end{subfigure}
     \hfill
     \begin{subfigure}[b]{0.29\textwidth}
         \centering
         \includegraphics[width=\textwidth]{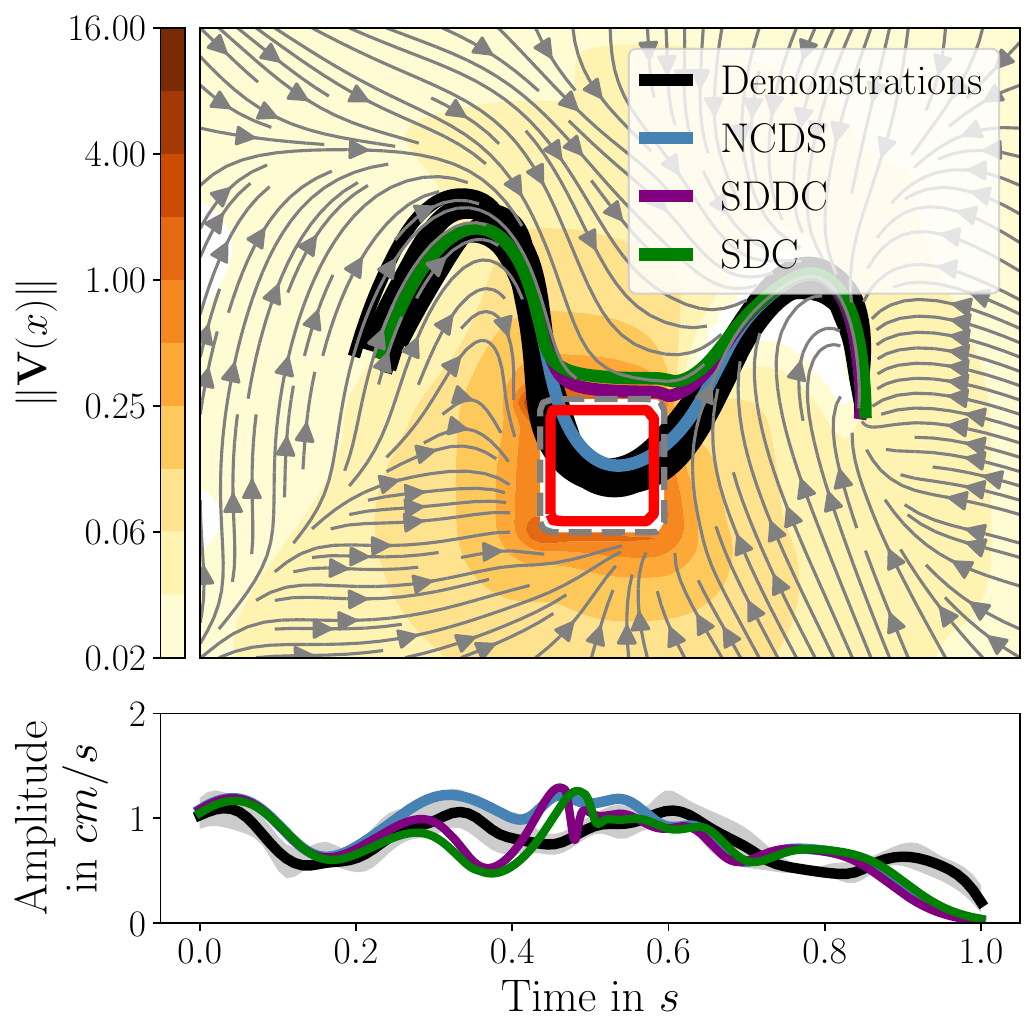}
         \caption{With swept features (Eq.~\eqref{eq:barrier_swept_s})}
         \label{fig:SDT_swept}
     \end{subfigure}
     \hfill
     \begin{subfigure}[b]{0.29\textwidth}
         \centering
         \includegraphics[width=\textwidth]{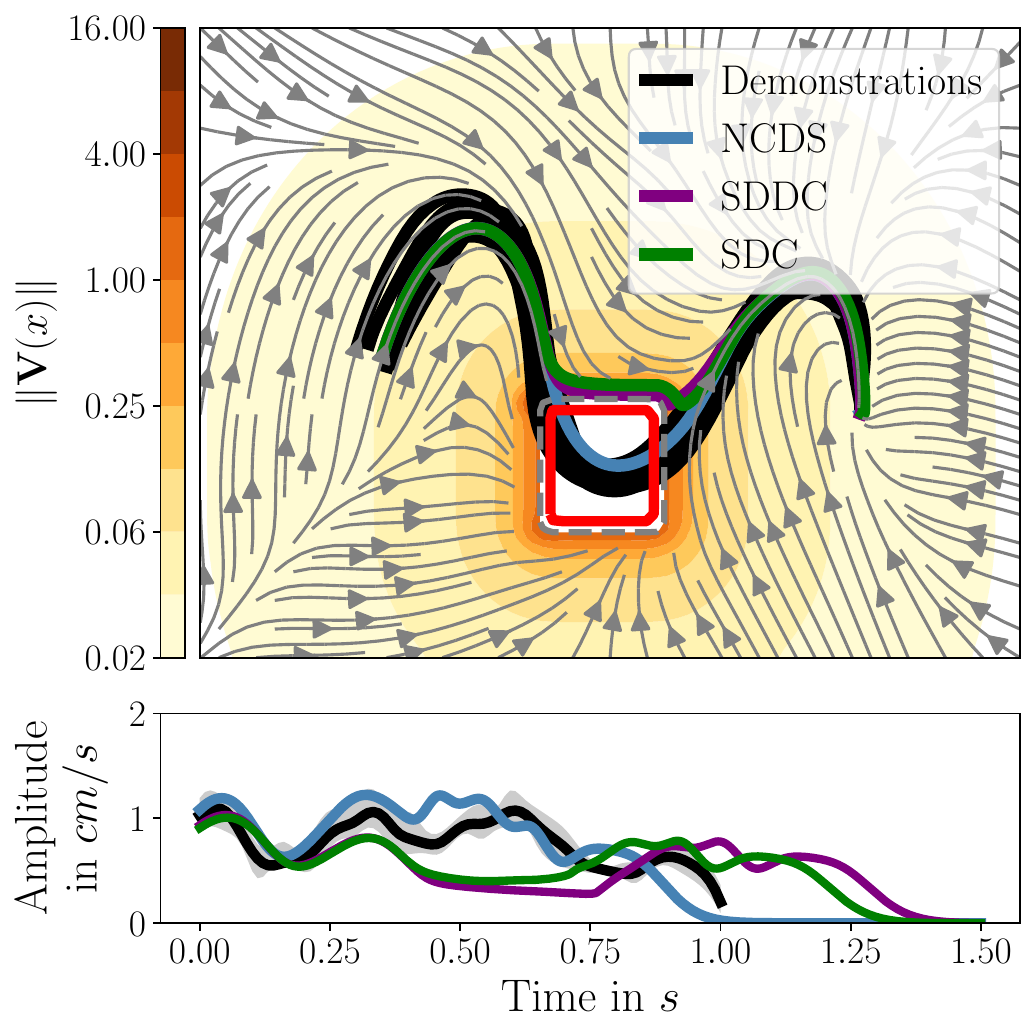}
         \caption{With friction and $\eta_f$ defined by Eq.~\eqref{eq:eta_friction}}
         \label{fig:SDT_friction}
     \end{subfigure}
        \caption{Comparison of contraction-preserving obstacle avoidance using \ac{SDDC} and \ac{SDC} on an example of the LASA dataset. The vector field modulated by the \ac{SDDC} is represented by a gray flow stream while the box-shaped obstacle is depicted by the red solid line. The bottom plots show the magnitude of the velocity profile along the trajectory.}
        \label{fig:three graphs}
\end{figure*}

In Equation~\eqref{eq:SDDC}, we consider only the differential coordinate change, which captures how velocity and direction transform under the diffeomorphism $\psi$. However, to fully account for geometric properties (e.g., lengths, distances, and angles), we must consider the Riemannian metric of the obstacle-avoiding manifold $\mathcal{Y}$ induced by $\psi$~\cite{Lee1997}. 
By applying the diffeomorphism $\psi$, the original dynamics $f_c$, defined on $\mathcal{C}$, can equivalently be described on the obstacle-avoiding manifold $\mathcal{Y}$ via,\looseness=-1
\begin{equation}
    \label{eq:sdt_transform}
    \dot{\mathbf{y}}=f_\text{SDC}(\mathbf{x},\mathbf{y})=\mathbf{J}_\psi(\mathbf{y})^{-1}f_c(\psi(\mathbf{x},\mathbf{y})).
\end{equation}
where $\mathbf{J}_\psi$ denotes the Jacobian of $\psi$. We refer to this transformed dynamical system as \textit{\ac{SDC}}. This coordinate change also induces the Riemannian metric $\mathbf{G}_\psi=\mathbf{J}_\psi^\top\mathbf{J}_\psi$, which characterizes the local geometry on $\mathcal{Y}$ and corresponds to the contraction metric. A proof of this equivalence is provided in Appendix~\ref{app:SDFDiffeomorphicTransform}.
To show that contraction is invariant under a change of coordinates we can use Theorem~\ref{theorem:Invariance_Under_Coordinate_Change}. 
Consequently, the proposed \ac{SDC} preserves the contraction property.
We illustrate our contraction-preserving obstacle avoidance methods \ac{SDDC} and \ac{SDC} using a toy example in Fig.~\ref{fig:SDT_no_friction}. This example shows that, given an obstacle, the original vector field $f_\text{c}$ from Fig.~\ref{fig:contractive_sys} is reshaped around the obstacle by the \ac{SDDC} and \ac{SDC}, thus avoiding it successfully.

\subsubsection{Enhancing Obstacle Avoidance with a Friction Term}
In the following, we introduce an extension to the proposed obstacle avoidance methods. 
To avoid notational clutter, we represent the vector field modulated by both the \ac{SDDC}~\eqref{eq:SDDC} and \ac{SDC}~\eqref{eq:sdt_transform} collectively as $f_\text{m}$.
As observed in Fig.~\ref{fig:SDT_no_friction}, the modulated velocity profile of $f_\text{m}$ may deviate significantly from the original velocity profile learned by the \ac{NCDS} skill. To address this, we introduce a friction term aimed at preserving the velocity profile of the underlying contractive vector field $f_\text{c}$. Interestingly, this friction term can also be employed to reduce the velocity near obstacles and to damp fluctuations caused by discontinuities in the learned \ac{SDF}.
Inspired by~\cite{Huber2022}, our friction term is defined as,
\begin{equation}
    \label{eq:friction_term}
    \dot{\mathbf{y}}_f = \eta_f \frac{\Vert f_\text{c}(\mathbf{y}) \Vert}{\Vert f_\text{m}(\mathbf{x},\mathbf{y}) \Vert} f_\text{m}(\mathbf{x},\mathbf{y}).
\end{equation}
When $\eta_f = 1$, the resulting dynamics adapts the velocity magnitude to the underlying learned vector field $f_\text{c}$, but keeps the modulated direction from $f_\text{m}$. 
Note that $\eta_f = 1$ must not be chosen for \ac{SDDC}, because it just guarantees strict collision avoidance at velocity level. If we enforce the \ac{SDDC} to keep the underlying velocity, it could potentially force the system to penetrate the obstacle.

To slow down near obstacles, we design $n_f$ as follows, 
\begin{equation}
\label{eq:eta_friction}
\eta_f = 1 - \frac{1}{(1+\beta_f \Gamma_\text{SDF}(\mathbf{x},\mathbf{y}))^{2}} ,
\end{equation}
with hyperparameter $\beta_f$. The resulting vector field under this friction term is depicted in Fig.~\ref{fig:SDT_friction}.
Since $\frac{\Vert f_\text{c}(\mathbf{y}) \Vert}{\Vert f_\text{SDC}(\mathbf{x},\mathbf{y}) \Vert}$ is strictly positive and $\eta_f>0$, they induce no directional changes. 
According to Theorem~\ref{theorem:Invariance_Under_Coordinate_Change}, contraction is preserved under affine transformations, consequently the friction~\eqref{eq:friction_term} does not compromise the contraction guarantees.

\subsubsection{Modulated Skill Execution}
Finally, trajectories can be integrated along $f_\text{m}$, defined by the Equations~\eqref{eq:SDDC} or~\eqref{eq:sdt_transform}.
First, $f_\text{c}$ is computed and then mapped into the obstacle-avoiding space via \ac{SDC}~\eqref{eq:sdt_transform} or \ac{SDDC}~\eqref{eq:SDDC}. Note that integrating the flow~\eqref{eq:sdt_flow_forward} requires solving an \ac{ODE}, which we address using off-the-shelf solvers (more information is provided in the Appendix in Table~\ref{tab:flow_solver}). After computing the flow, applying Equation~\eqref{eq:SDDC} or~\eqref{eq:sdt_transform} reshapes the vector fields to avoid obstacles while preserving global contraction guarantees. Algorithm~\ref{alg:sdt_pipeline} summarizes how a robot is controlled using our approach.

\begin{algorithm}[t]
\caption{Modulated Robot Skill Execution}\label{alg:sdt_pipeline}
\SetKwInOut{Input}{Input}
\SetKwInOut{Initialize}{Initialize}

\Input{Learned \ac{NCDS} $f_{\text{NCDS},\boldsymbol{\theta}}$, learned \ac{SDF} $\Gamma_{\mathcal{R},\hat{\theta}}$}
\Initialize{$N \gets$ maximum number of steps, $t \gets 0$}

\While{$t \leq N$}{
    Sample point cloud $\{\mathbf{s}_i \}_{i=1}^M \sim \mathcal{S}$ from the scene \\
    Read robot joint and task state $\mathbf{q}_t \in \mathcal{C}, \mathbf{x}_{t} \in \mathcal{X}$ \\
    Find nearest point $\mathbf{s}^*$ to robot surface such that $\Gamma_\mathcal{R}(\mathbf{s}^*,\mathbf{q})= \min_{\{\mathbf{s}_i \}_{i=1}^M}\Gamma_\mathcal{R}(\mathbf{s}_i,\mathbf{q})$ \\
    Solve the ODE from Equation~\eqref{eq:sdt_flow_forward} $\to \psi(\mathbf{x},\mathbf{q}_t)$ \\
    Solve \ac{ODE} from Equation~\eqref{eqn:f_NCDS} $ \to \mathbf{\dot{y}}_{\text{NCDS},t}$ \\
    Modulate $\dot{\mathbf{y}}_{\text{NCDS},t}$ via Equation~\eqref{eq:SDDC} or~\eqref{eq:sdt_transform} $\to \mathbf{\dot{y}}_{t}$ \\
    Integrate modulated joint velocity $\dot{\mathbf{y}}_{t}$ $\to \mathbf{y}_{t+1}$\\
    Send $\mathbf{y}_{t+1}$ to the robot's joint impedance controller \\
    $t \gets t + 1$\;
}
\end{algorithm}

\section{Obstacle Avoidance Metrics}
\label{sec:evaluation_metrics}
We here introduce two novel metrics: \ac{RFC} and \ac{VM}, to quantitatively assess the obstacle avoidance performance of our approach and state-of-the-art methods. In contrast to standard metrics such as the \ac{DTWD}, that measure reconstruction errors, the proposed metrics focus instead on changes of the learned vector field introduced by the obstacle avoidance modulation.

\subsection{Relative Flow Curvature}
To ensure that the obstacle avoidance maneuvers do not introduce jerky or sharp trajectories, we analyze the curvature of the modulated vector field. This metric, termed \ac{RFC}, should ideally remain similar to the original vector field's curvature, thus indicating smooth and gradual adjustments around obstacles.
Specifically, \ac{RFC} compares the maximal curvature $\kappa$ along the modulated trajectory $\tau_{m}$ against the maximal curvature of the base trajectory $\tau_\text{base}$. The curvature $\kappa$ measures how sharply a trajectory $\tau$ bends at a given point relative to the tangential direction~\cite{Pressley2010}, and it is defined as,
\begin{equation}
    \label{eq:curvature}
    \kappa(t)=\frac{\Vert \ddot{\tau}(t) \times \dot{\tau}(t) \Vert}{\Vert \dot{\tau}(t)\Vert ^{3}}.
\end{equation}
with $\times$ denoting the cross product, $\dot{\tau}(t)$ and $\ddot{\tau}(t)$ being the velocity and acceleration along the trajectory.
Given the curvature~\eqref{eq:curvature}, the \ac{RFC} metric is defined as follows,
\begin{equation}
    \label{eq:RFC}
    \text{RFC} = \left\vert \max_{\mathbf{x} \in \tau_\text{base}} \lVert \kappa(\mathbf{x}) \rVert - \max_{\mathbf{x} \in \tau_{m}} \lVert \kappa(\mathbf{x}) \rVert\right\vert.
\end{equation}

\subsection{Vector Field Misalignment}
When considering contractive stable vector fields, it is particularly important to preserve the learned vector field patterns as much as possible in order to maintain the essence of the skill. Therefore, we propose to compute the cosine similarity between the learned vector field $f_c$ and the modulated vector field $f_m$ along the rolled-out obstacle-avoiding trajectory. Since we aim to quantify the degree of misalignment between the two vector fields, we define this metric as,
\begin{equation}
    \label{eq:VM}
    \text{VM} = \frac{1}{|\tau_m|} \sum_{\mathbf{x} \in \tau_m}  \arccos{ \left( \frac{f_c(\mathbf{x}) \cdot f_m(\mathbf{x})}{\vert f_c(\mathbf{x}) \vert \ \vert f_m(\mathbf{x}) \vert} \right)} 
\end{equation}
A high \ac{VM} value indicates that the base vector field and the modulated vector field significantly deviate from each other. Therefore a low \ac{VM} is desirable as this may indicate that the learned vector field patterns are just slightly modified.

%-----------------------------------------
% Experiments
%-----------------------------------------
\section{Experiments} 
\label{sec:experiments}
\begin{SCfigure}[0.5][b]
\caption{Illustration of the obstacles \ac{SDF} used in the 2D LASA dataset experiments. The contour lines represent the distance $\Gamma_\text{SDF}$ to the obstacle surface, depicted as a solid red line. Four simple obstacles are considered: Circle, Box, Triangle, Arc.}
\includegraphics[width=0.30\textwidth]{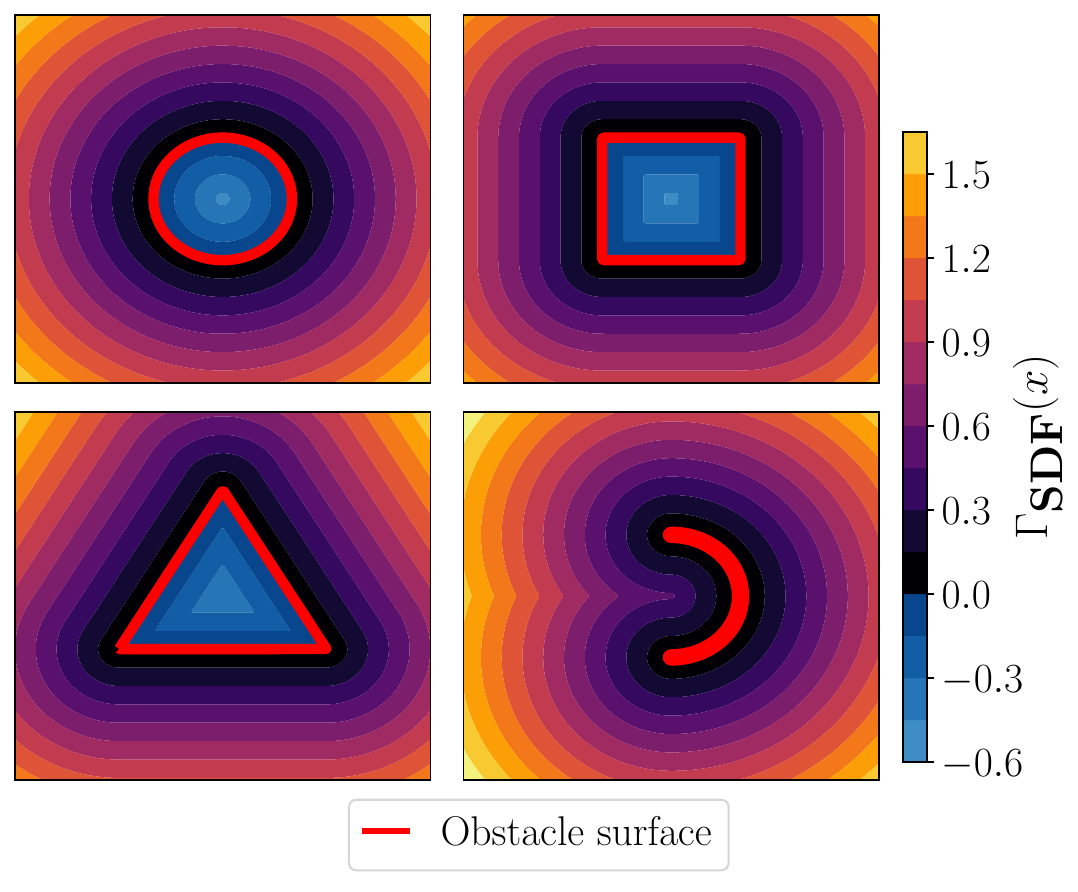}
\label{fig:obstacles}
\end{SCfigure}

In the following experiments, we systematically analyze the performance, real-time feasibility, and potential edge-cases of the proposed obstacle avoidance methods.
Thereby, we evaluate the proposed methods \ac{SDDC} and \ac{SDC} on the 2D LASA dataset~\cite{Lemme2015:LasaDataset} and on two real-world tasks: \textsc{Emptying a dishwasher} and \textsc{Opening a dishwasher}. These experiments are carried out in a kitchen environment with unknown obstacles. 
For the real-world experiments, the robot surface is learned as an \ac{SDF} function while the scene is represented by an incrementally-sampled point cloud.
For the 2D LASA dataset experiments, only implicit models of the obstacles are learned while the robot is assumed to be a virtual point. Here, simple geometries such as circles, rectangles, triangles or arcs are used as obstacles, as shown in Fig.~\ref{fig:obstacles}.

Our methods are compared against \ac{MM}~\cite{Huber2022} as this is the only contraction-preserving obstacle avoidance method known to us, which was also employed in combination with \ac{NCDS}~\citep{BeikMohammadi2024}. 
Moreover, we compare our methods against the \ac{DT}~\citep{Zhi2022}, which employs diffeomorphic transformations but relies on a natural gradient formulation and thus does not preserve contraction. 
Additionally, we consider classical \ac{ARPF}~\cite{Khatib1985}. 
We ablate our methods using different \ac{SDF} functions based on \ac{MLP}, \ac{BP}, and simple geometric primitives. These different versions are compared against Hilbert maps~\cite{Ramos2016} as this method was used in combination with \ac{DT} for obstacle avoidance~\cite{Zhi2022}.

\subsection{Evaluation Metrics}
\label{sec:evaluation_metrics}
The following metrics are used to quantitatively evaluate the obstacle avoidance performance.
Firstly, we evaluate our method's ability to preserve the underlying learned skill during obstacle avoidance. To do so, we compare the skill's execution with and without obstacles using the \ac{DTWD}~\cite{Berndt1994}.
A low \ac{DTWD} indicates that the reproduced trajectory was slightly modified by the obstacle avoidance term.
Secondly, we compute the minimum distance $\text{D}_\text{min}= \min_{\mathbf{x}\in \tau_m}\Gamma_\text{SDF}(\mathbf{x})$ between the robot surface and the obstacle over the course of the skill execution.
Our goal should be to keep this distance sufficiently large to guarantee a collision-free skill execution.
Thirdly, to prevent obstacle avoidance from generating abrupt or discontinuous trajectories, we use our proposed \ac{RFC} metric~\eqref{eq:RFC} to measure the curvature of the modulated vector field.
As fourth metric we analyze the trajectory jerkiness, which measures abrupt acceleration changes possibly introduced by the obstacle avoidance term. To assess this, we compare the maximal trajectory jerkiness in both obstacle-free $\dddot{\tau}_\text{base}(\mathbf{x})$ and obstacle-avoidance settings $ \dddot{\tau}_{m}(\mathbf{x})$, resulting in the \textit{\ac{MJ}} metric. The jerk difference should be as small as possible.
Finally, we evaluate the proposed \ac{VM} metric~\eqref{eq:VM}, which measures the deviation between the base vector field and the modulated vector field.
Details about the \ac{DTWD}, \ac{MJ} and $\text{D}_\text{min}$ metrics are given in Appendix~\ref{app:metrics}.

\subsection{Implementation Details}
All experiments are run on a NVIDIA RTX A2000 12GB GPU and a 12th Gen Intel(R) Core(TM) i5-12600K CPU.
Next, we provide the architectures and the hyperparameters used for learning the skill models and the obstacle implicit representations.\looseness=-1

\subsubsection{Neural Contractive Dynamical System}
Our \ac{NCDS} implementation fundamentally follows that of~\citet{BeikMohammadi2024}. The \ac{NCDS} Jacobian network is modeled using an \ac{MLP} with $2$ hidden layers, each consisting of $100$ units and employing \textsc{tanh} activation functions. We define the total loss function as a combination of Equations~\eqref{eq:NCDS_loss}, \eqref{eq:NCDS_loss_state_independant} and \eqref{eq:NCDS_loss_noise}, with weighting factors $\beta_\epsilon=\num{1.0e-3}$ and $\beta_\text{noise}=1.0$
All models are trained for $1000$ epochs with a learning rate of $\num{1e-3}$ using the Adam optimizer. A decay factor of $0.1$ is applied every $250$ epochs to improve training stability. Furthermore, the demonstration data of each skill are linearly interpolated such that each trajectory consists of exactly $1000$ points.
At inference time, a fourth-order Runge-Kutta solver with the $3/8$ rule is employed to solve the \ac{ODE} in Equation~\eqref{eqn:f_NCDS} using two solver steps.

\subsubsection{Signed Distance Fields}
For the LASA dataset experiments, we first model the \ac{SDF} based on the mathematical description of the simple geometric shapes, as displayed in Fig.~\ref{fig:obstacles}. Using this representation, we sample $N=160K$ points $\{\mathbf{x}_i\}_{i=0}^N$ from a grid around the obstacle surface and we then assign each point its corresponding \ac{SDF} $d_i=\Gamma_\text{SDF}(\mathbf{x}_i)$ and gradient values $\mathbf{g}_i=\nabla_\mathbf{x}\Gamma_\text{SDF}(\mathbf{x}_i)$. Given this training set $\mathcal{D}=\{(\mathbf{x}_i,d_i,\mathbf{g}_i)\}_{i=0}^N$, a two-dimensional \ac{SDF} is learned via an \ac{MLP} with an hourglass-shaped architecture of size ($128$, $64$, $32$, $32$) and \textsc{ReLU} activation functions, inspired by~\citep{Gropp2020,Yiming2024}.
The loss function is a combination of a reconstruction loss $\mathcal{L}_\text{SDF}$ ~\cite{Park2019}, Eikonal loss~\cite{Gropp2020}, gradient-based loss~\cite{Ortiz2022}, and a tension loss~\cite{Jttler2002}, leading to,
\begin{equation}
    \mathcal{L}(\boldsymbol{\theta}) = w_\text{SDF}\mathcal{L}_\text{SDF} + w_\text{grad} \mathcal{L}_\text{grad} + w_\text{eik} \mathcal{L}_\text{eik} + w_\text{ten} \mathcal{L}_\text{ten} ,
\end{equation}
where $w_\text{SDF}$, $w_\text{grad}$, $w_\text{eik}$ and $w_\text{tension}$ are weighting factors with values $w_\text{SDF} = 10.0$, $w_\text{eik} = 0.01$, $w_\text{grad} = 0.1$ and $w_\text{ten} = 0.01$. The models are trained for $5000$ epochs using an Adam optimizer with a learning rate of $\num{1e-3}$. In addition, we learn a \ac{BP}-based \ac{SDF} similarly to~\citet{YimlinLi2023}. We choose a model size of $8$ basis functions, which are trained for $200$ epochs using the training set $\mathcal{D}$.

For the robot experiments, we learn an implicit representation of the $7$-DoF Franka-Emika Panda robot. On the one hand, we use the \ac{BP} architecture proposed by~\citet{YimlinLi2023}. We choose $14$ basis functions and train the model for $200$ epochs. On the other hand, we learn a \ac{CDF} of the robot based on a \ac{MLP} model combined with positional encoding and adopt the architecture of~\citet{Yiming2024}. We choose the same loss as described for the 2D \ac{MLP} model but with an architecture of size ($1024$, $512$, $256$, $128$, $128$) and learn the model for $50000$ epochs.\looseness=-1

\subsection{Baselines}
\label{sec:baselines}
To compare our approach with state-of-the-art methods, we must adapt them to work with \ac{SDF}s. 
Firstly, \citet{Huber2022} used \ac{MM} with a custom distance function $\Gamma_\text{M}$ designed for simple geometric primitives, which is not directly comparable to our framework.
Thus, we define a distance function such that $\mathcal{H}^s = \{ \mathbf{x} \in \mathbb{R}^d \mid \Gamma_{\text{M}_\text{SDF}}(\mathbf{x}) = 1 \} $, resulting in,
\begin{equation}
    \Gamma_{\text{M}_\text{SDF}}(\mathbf{x}) = \left( \Gamma_\text{SDF}(\mathbf{x}) + \mathbf{1} \right)^{2p}.
\end{equation}
Given this distance function $\Gamma_{\text{M}_\text{SDF}}$, we compute the modulation matrix $\mathbf{M}$ according to \citet{Huber2022}. So, the modulated vector field is,
\begin{equation}
    \dot{\mathbf{x}} = \mathbf{M}(\mathbf{x}) f_\text{c}(\mathbf{x}).
\end{equation}

Secondly, to compare against the \ac{DT} approach, we replace the diffeomorphism based on Hilbert maps~\cite{Zhi2022} by our diffeomorphism $\psi$ from Eq.~\eqref{eq:sdt_flow_forward}. This results in the modified natural gradient system,
\begin{equation}
    \label{eq:dt_transform}
    \dot{\mathbf{y}}=f_\text{DT}(\mathbf{x},\mathbf{y})=\mathbf{G}_\psi(\mathbf{y})^{-1}f_c(\psi(\mathbf{x},\mathbf{y}))
\end{equation}
with the Riemannian Metric $\mathbf{G}_\psi=\mathbf{J}_\psi^\top\mathbf{J}_\psi$.
Finally, we adapt the \ac{ARPF} method~\cite{Khatib1985} by defining the repulsive potential $V_\text{ARPF}$ via the \ac{SDF} function, as follows, 
\begin{equation}
    V_\text{ARPF}(\mathbf{x}) = \frac{1}{2}\eta \left( \frac{1}{\Gamma_\text{SDF}(\mathbf{x})-t_\text{safe}} - 1 \right)^{2} ,
\end{equation}
with safety threshold $t_\text{safe}=0.1$ and a positive constant $\eta= 10^{-4}$. 
Given this repulsive potential, the resulting dynamics of the adapted contractive system $f_\text{c}$ is,
\begin{equation}
    \mathbf{\dot x}=f_\text{c}-\nabla_\mathbf{x} V_\text{ARPF}(\mathbf{x}).
\end{equation}

\subsection{Validation on the LASA Dataset}
The aim of these experiments is to determine the quality and efficiency of obstacle avoidance as well as the limitation of the proposed methods on the basis of simplified scenarios.

\subsubsection{Experimental Setup}
The LASA dataset~\cite{Lemme2015:LasaDataset} is used to validate the proposed pipeline using a 2D dataset, consisting of $30$ 2D handwritten movements. Each movement consists of seven demonstrations.
As an example, we use the \textit{Sine} movement from the LASA dataset for the following experiments (see Fig.~\ref{fig:contractive_sys}). In this case, we assume a virtual point-shaped robot following the learned contractive NCDS models trained on the \textit{Sine} dataset. The task for the robot is to avoid one of the obstacles shown in Fig.~\ref{fig:obstacles}. In addition, in the Appendix (see Fig.~\ref{fig:SDT_multi_modal}) multi-modal robot skills are also investigated.

The obstacle is represented by an implicit distance function.
Here, we use three different \ac{SDF} methods for this purpose.
First, we define distance functions w.r.t geometric primitives (see Fig.~\ref{fig:obstacles}). These are also used to generate the training data, which serve as the ground truth. Additionally, we learn an \ac{SDF} represented by a \ac{MLP} and another represented via \ac{BP}.
We analyze the performance of the methods for $40$ different trajectories per obstacle, each with different obstacle positions.

\subsubsection{Obstacle Avoidance Results}
As an exemplary scene, we place the box-shaped obstacle in the trajectory path and apply our proposed method and compute both the trajectory with and without obstacle avoidance.
As shown in Fig.~\ref{fig:SDT_friction} the learned contractive vector field is deflected around the obstacle and thus we can successfully avoid an convex obstacle using our proposed method. If the barrier function is additionally extended with the \textit{swept} feature, Fig.~\ref{fig:SDT_swept} shows that the vector field away from the obstacle is less influenced and thus the movement in this area follows more closely the dynamics learned by \ac{NCDS}.

\begin{SCfigure}[0.5][t]
\caption{Obstacle avoidance with \ac{SDDC} and \ac{SDC} using a inverse barrier on a \ac{NCDS} model trained on the LASA dataset. The arc-shaped obstacle is depicted by the red line. The velocity profile shows the absolute velocity value along the trajectory. The vector field modulated by the \ac{SDDC} is represented by gray arrows.}
\includegraphics[width=0.31\textwidth]{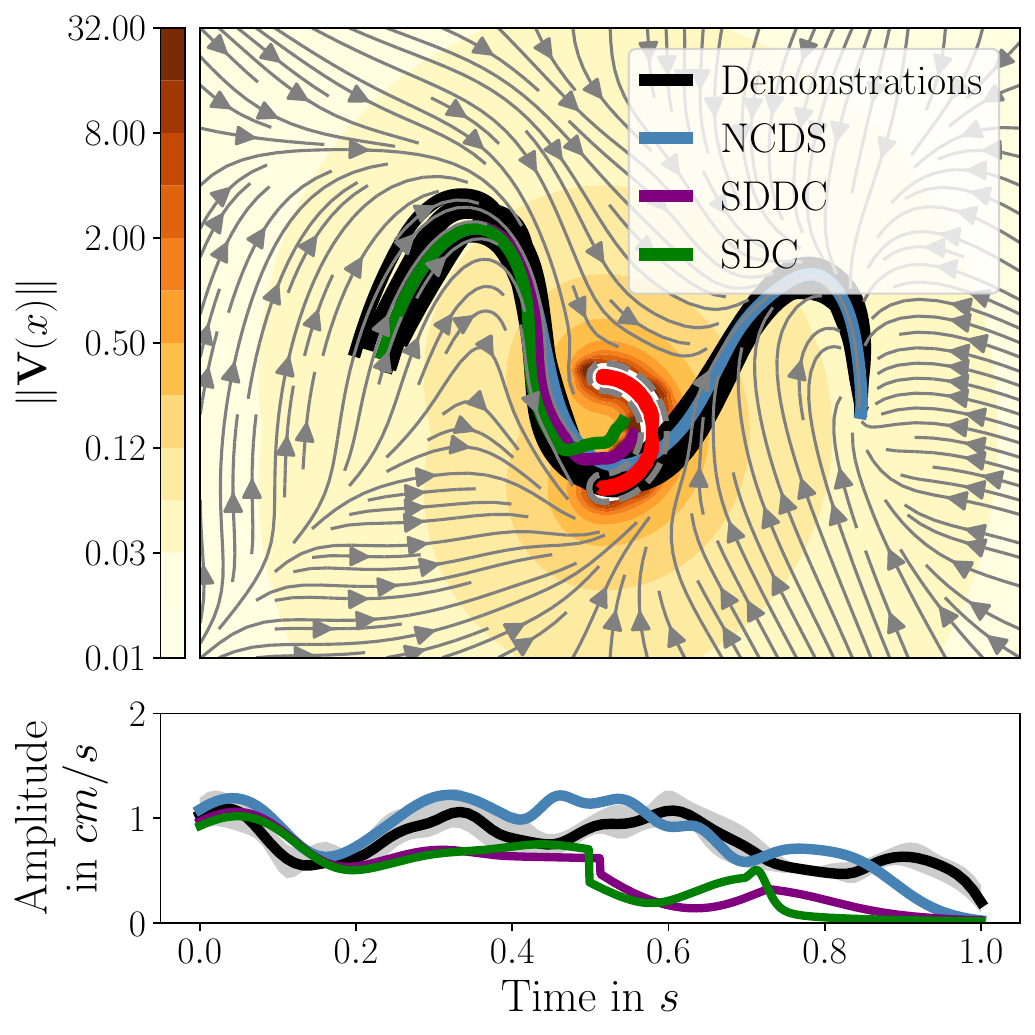}
\label{fig:SDT_concave}
\end{SCfigure}

However, the obstacles considered so far have convex shapes. In the following, we investigate how the methods behave with concave surfaces. As shown in Fig.~\ref{fig:SDT_concave}, when using the \ac{SDT} framework, the flow gets stuck in the concave region. Note that the learned \ac{SDF} does not provide any global information regarding the shape of the obstacle. The curvature can therefore only be calculated locally. Consequently, the vector field is deflected at the edges of the obstacle profile, regardless of the obstacle shape. Therefore, \citet{Huber2022} suggest defining a reference point $\mathbf{x}^{r}$ that specifies a clear direction for the flow. However, this does not serve any purpose for a \ac{SDF} function, as the global information is only implicitly represented. In addition, the reference point would have to be adjusted continuously for dynamic objects. 
This confirms our expectations that avoidance of concave obstacles may prevent a complete execution of the trajectory, thus we consider our method primarily relevant for convex obstacles.

\subsubsection{Barrier Functions Hyperparameter}
We here analyze the influence of the hyperparameter $s_\text{grad}$ in Eq.~\eqref{eq:barrier_advanced} on the obstacle avoidance behavior, using the metrics proposed in Section~\ref{sec:evaluation_metrics}. Table~\ref{tab:s_grad} reports the metrics for three different values of $s_\text{grad}$. 
The experiments show that higher values of $s_\text{grad}$ increase the minimum distance to the obstacle. This in turn has a negative influence on the \ac{DTWD} and the \ac{VM} metrics, as this also increases the area of influence of the obstacle avoidance and thus modulates a larger region of the learned vector field. In addition, the results show that the curvature decreases for larger values of $s_\text{grad}$, except for $\text{SDDC}_\text{inv,swept}$, and thus the obstacle can be avoided more smoothly.
We therefore see a potential trade-off between a smooth modulation with higher distance to the obstacle (i.e., high $s_\text{grad}$ values), and a more accurate alignment to the skill vector field, but with less smooth modulation (i.e., low $s_\text{grad}$ values).
Moreover, the swept features have a significant positive impact on the majority of the metrics, particularly noticeable in the \ac{RFC} and \ac{VM} metric. This can be explained by the fact that these swept features minimize the area of influence of the obstacle avoidance term.

\begin{table}[tbp]
  \centering
  \caption{Quantitative analysis of the influence of $s_\text{grad}$ on the obstacle avoidance behavior according to Eq.~\ref{eq:barrier_advanced}. Only obstacles based on geometric primitives (GP) are considered. Each metric is computed as the average over $40$ trajectories.}
  \label{tab:s_grad}
  \footnotesize
    \begin{tabular}{lrrrrr}
        \rowcolor{gray!15}
        & \textbf{MJ} & \textbf{RFC} & \textbf{VM} & \textbf{DTWD} & $\textbf{D}_\textbf{min}$ \\
        \rowcolor{gray!15}
        \multicolumn{6}{l}{$\mathbf{s}_\textbf{grad}=\mathbf{0.05}$} \\
        $\text{SDDC}_\text{inv}$& 
        $0.74 e\!-\!4$ & $10.19$ & $0.49$ & $123.8$ & $0.28$ \\
        \rowcolor{gray!5}
        $\text{SDDC}_\text{inv,swept}$& 
        $0.69 e\!-\!4$ & $3.80$ & $0.42$ & $\mathbf{84.9}$ & $0.29$ \\
        $\text{SDC}_\text{inv}$&
        $5.08 e\!-\!4$ & $23.58$ & $0.20$ & $104.1$ & $0.35$ \\
        \rowcolor{gray!5}
        $\text{SDC}_\text{inv,swept}$& 
        $3.93 e\!-\!4$ & $21.26$ & $\mathbf{0.15}$ & $99.9$ & $0.34$\\
        \rowcolor{gray!15}
        \multicolumn{6}{l}{$\mathbf{s}_\textbf{grad}=\mathbf{0.10}$} \\
        $\text{SDDC}_\text{inv}$& 
        $0.62 e\!-\!4$ & $8.80$ & $0.57$ & $184.2$ & $0.31$ \\
        \rowcolor{gray!5}
        $\text{SDDC}_\text{inv,swept}$& 
        $0.62 e\!-\!4$ & $\mathbf{3.75}$ & $0.54$ & $127.8$ & $0.32$ \\
        $\text{SDC}_\text{inv}$&
        $4.81 e\!-\!4$ & $15.50$ & $0.26$ & $163.4$ & $0.42$ \\
        \rowcolor{gray!5}
        $\text{SDC}_\text{inv,swept}$& 
        $3.25 e\!-\!4$ & $12.59$ & $0.20$ & $149.9$ & $0.41$\\
        \rowcolor{gray!15}
        \multicolumn{6}{l}{$\mathbf{s}_\textbf{grad}=\mathbf{0.20}$} \\
        $\text{SDDC}_\text{inv}$& 
        $\mathbf{0.56 e\!-\!4}$ & $8.14$ & $0.69$ & $288.5$ & $0.34$ \\
        \rowcolor{gray!5}
        $\text{SDDC}_\text{inv,swept}$& 
        $0.62 e\!-\!4$ & $3.99$ & $0.70$ & $177.2$ & $0.35$ \\
        $\text{SDC}_\text{inv}$&
        $3.81 e\!-\!4$ & $8.58$ & $0.36$ & $269.6$ & $\mathbf{0.53}$ \\
        \rowcolor{gray!5}
        $\text{SDC}_\text{inv,swept}$& 
        $3.33 e\!-\!4$ & $5.99$ & $0.29$ & $238.6$ & $0.52$\\
        
    \end{tabular}
\end{table}

\begin{SCfigure}[0.5][b]
\caption{Obstacle avoidance with \ac{SDC} and \ac{SDDC} using a inverse barrier on a \ac{NCDS} trained on the Sine trajectories of the LASA dataset. The circle-shaped obstacle is depicted by the red line. The velocity profile shows the absolute velocity value along the trajectory. The vector field modulated by the \ac{SDDC} is represented by gray arrows.}
\includegraphics[width=0.31\textwidth]{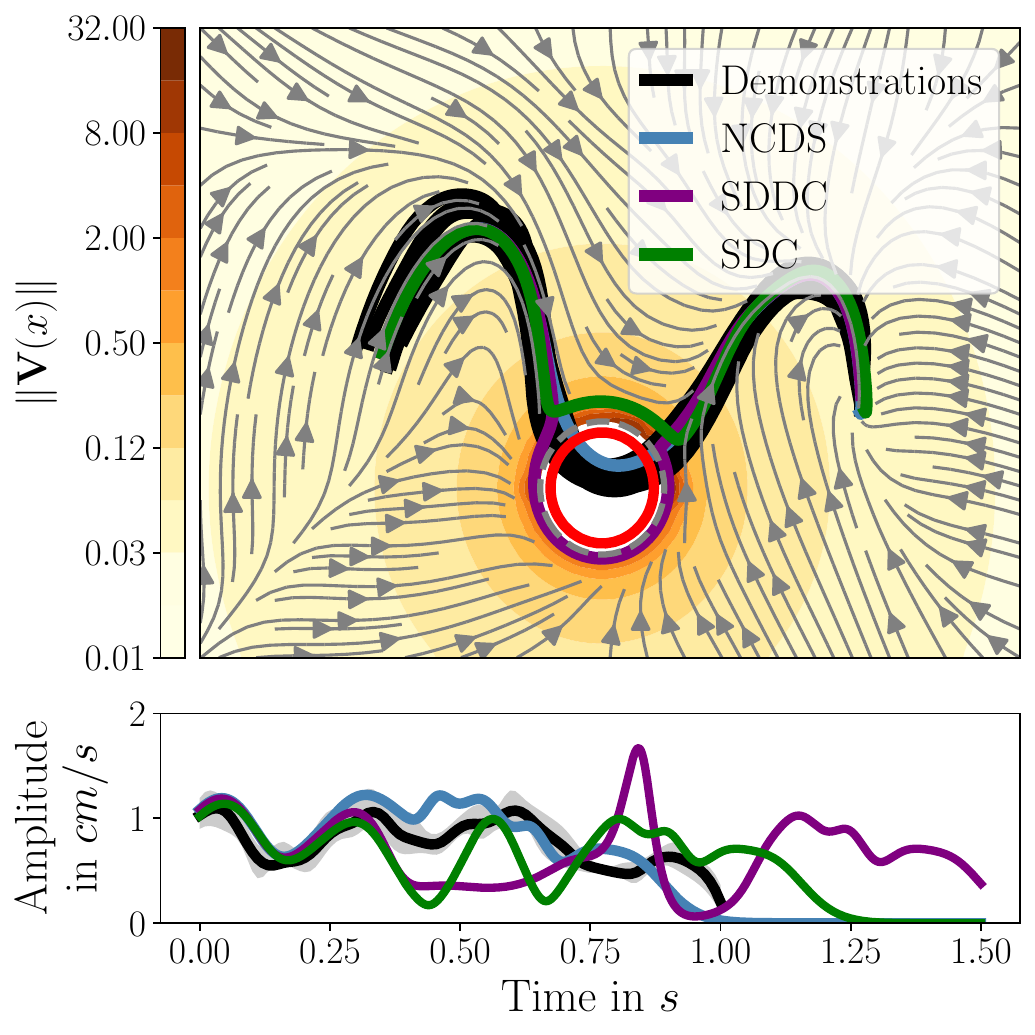}
\label{fig:SDT_vs_SDDC}
\end{SCfigure}

\subsubsection{Comparison to State-Of-The-Art}
To compare our proposed \ac{SDC} and \ac{SDDC} methods with state-of-the-art approaches, we first adapt the \ac{MM}~\cite{Huber2022}, \ac{ARPF}~\cite{Khatib1985} and \ac{DT}~\cite{Zhi2022} methods, as detailed in Section~\ref{sec:baselines}, to ensure a fair comparison.
We then evaluate all approaches using three distinct \acp{SDF} architectures. The results are summarized in Table~\ref{tab:obstacle_avoidance_comparison}.
Overall, both \ac{SDC} and \ac{SDDC} outperform the baselines in key metrics such as \ac{RFC}, \ac{VM} and \ac{DTWD}. It is noteworthy that \ac{ARPF} exhibits particularly strong performance with respect to the \ac{MJ} metric, indicating jerk-free obstacle avoidance. However, \ac{ARPF} does not guarantee contraction, as proved in Appendix~\ref{app:reactive_obs_avoidance}.

Another noticeable finding is that \ac{SDDC} produces relatively large \ac{VM} values, indicating weaker alignment with the original contractive vector field during obstacle avoidance. In addition, its low $\text{D}_\text{min}$ and high \ac{DTWD}$\,$ values imply that \ac{SDDC} follows a suboptimal path. This observation is qualitatively reinforced by Fig.~\ref{fig:SDT_vs_SDDC}, where \ac{SDDC} takes a longer path around the obstacle. Nevertheless, this path is characterized by a particularly smooth trajectory, as indicated by its low \ac{RFC} value.

The differences among the three \ac{SDF} functions are noticeably, but do not change the ranking of the methods regarding the metrics. However, the \ac{MLP}-based \ac{SDF} tends to exhibit higher \ac{MJ} and \ac{RFC} values, indicating increased irregularities. By contrast, the \ac{BP}-based \ac{SDF} appears more smooth, which aligns with the features provided by polynomial-based functions.
\textbf{In conclusion}, the foregoing results demonstrate that our \ac{SDC} and \ac{SDDC} methods provide smooth and stable obstacle avoidance behaviors without compromising contraction guarantees and reconstruction accuracy. If closer alignment to the underlying contractive vector field is preferred, \ac{SDC} is recommended, whereas \ac{SDDC} offers a more uniform obstacle avoidance strategy.

\begin{table*}[tbp]
  \centering
  \caption{Comparison of the proposed \ac{SDC} and \ac{SDDC} methods against state-of-the-art baselines. Three different implicit distance functions are considered for all methods: \ac{SDF} of geometric primitive (GP) as ground truth, \ac{MLP} as standard architecture for \ac{SDF}~\cite{Gropp2020,Park2019}, and \ac{BP} as polynomial architecture~\cite{YimlinLi2023}. The methods \ac{SDC} and \ac{DT} use a barrier with $s_\text{grad}=0.1$, while \ac{SDDC} uses a barrier with $s_\text{grad}=0.05$. An additional safety margin of $t_\text{save}=0.1$ applies to all methods. Each metric was averaged over $40$ trajectories.}
  \label{tab:obstacle_avoidance_comparison}
  \footnotesize
  \resizebox{\linewidth}{!}{ 
    \begin{tabular}{lrrrrr|rrrrr|rrrrr}
        \rowcolor{gray!15}
        & \multicolumn{5}{c}{\textbf{GP}} & \multicolumn{5}{c}{\textbf{MLP}} &
         \multicolumn{5}{c}{\textbf{BP}}\\
        \rowcolor{gray!15}
        \textbf{Methods} & \textbf{MJ} & \textbf{RFC} & \textbf{VM} & \textbf{DTWD} & $\textbf{D}_\textbf{min}$ & \textbf{MJ} & \textbf{RFC} & \textbf{VM} & \textbf{DTWD} & $\textbf{D}_\textbf{min}$ & \textbf{MJ} & \textbf{RFC} & \textbf{VM} & \textbf{DTWD} & $\textbf{D}_\textbf{min}$\\ 
        $\text{SDDC}_\text{inv}$& 
        $0.62 e\!-\!4$ & $8.80$ & $0.57$ & $184.2$ & $0.31$ &
        $4.02 e\!-\!4$ & $7.78$ & $0.61$ & $174.2$ & $0.34$ & 
        $1.51 e\!-\!4$ & $4.38$ & $0.74$ & $222.4$ & $0.31$\\
        \rowcolor{gray!5}
        $\text{SDDC}_\text{inv,swept}$& 
        $0.62 e\!-\!4$ & $\mathbf{3.75}$ & $0.54$ & $127.8$ & $0.32$ & 
        $3.37 e\!-\!4$ & $\mathbf{6.96}$ & $0.47$ & $129.9$ & $0.33$& 
        $1.16 e\!-\!4$ & $\mathbf{4.14}$ & $0.49$ & $134.8$ & $0.30$\\ 
        $\text{SDC}_\text{inv}$&
        $5.08 e\!-\!4$ & $23.58$ & $0.20$ & $104.1$ & $0.35$ & 
        $7.44 e\!-\!4$ & $28.47$ & $0.20$ & $98.5$ & $0.37$ & 
        $3.16 e\!-\!4$ & $8.42$ & $0.10$ & $98.8$ & $0.34$\\
        \rowcolor{gray!5}
        $\text{SDC}_\text{inv,swept}$& 
        $3.93 e\!-\!4$ & $21.26$ & $\mathbf{0.15}$ & $\mathbf{99.9}$ & $0.34$ & 
        $6.33 e\!-\!4$ & $27.79$ & $\mathbf{0.15}$ & $\mathbf{97.7}$ & $0.35$ & 
        $2.12 e\!-\!4$ & $7.71$ & $\mathbf{0.09}$ & $\mathbf{95.4}$ & $0.33$\\
        $\text{DT}_\text{inv}$&
        $0.42 e\!-\!4$ & $9.74$ & $0.42$ & $151.3$ & $\mathbf{0.40}$ & 
        $3.27 e\!-\!4$ & $12.31$ & $0.34$ & $132.9$ & $\mathbf{0.41}$ & 
        $1.58 e\!-\!4$ & $5.32$ & $0.42$ & $156.8$ & $\mathbf{0.37}$\\
        \rowcolor{gray!5}
        $\text{DT}_\text{inv,swept}$& 
        $0.52 e\!-\!4$ & $8.34$ & $0.40$ & $143.4$ & $0.38$ & 
        $2.67 e\!-\!4$ & $8.95$ & $0.34$ & $132.9$ & $0.40$ & 
        $1.10 e\!-\!4$ & $5.67$ & $0.28$ & $147.9$ & $0.34$\\
        MM & 
        $2.66 e\!-\!4$ & $9.66$ & $0.45$ & $199.4$ & $0.34$ & 
        $9.13 e\!-\!4$ & $18,63$ & $0.42$ & $177.1$ & $0.35$ & 
        $0.34 e\!-\!4$ & $5,13$ & $0.62$ & $208.5$ & $0.33$\\
        \rowcolor{gray!5}
        ARPF & 
        $\mathbf{0.15 e\!-\!4}$ & $8.08$ & $0.61$ & $146.6$ & $0.32$ & 
        $\mathbf{2.00 e\!-\!4}$ & $22,13$ & $0.57$ & $135.1$ & $0.33$ & 
        $\mathbf{0.02 e\!-\!4}$ & $6.72$ & $0.84$ & $198.3$ & $0.31$\\
    \end{tabular}
  }
\end{table*}

\begin{figure}[t]
    \centering
    \includegraphics[width=0.4\textwidth]{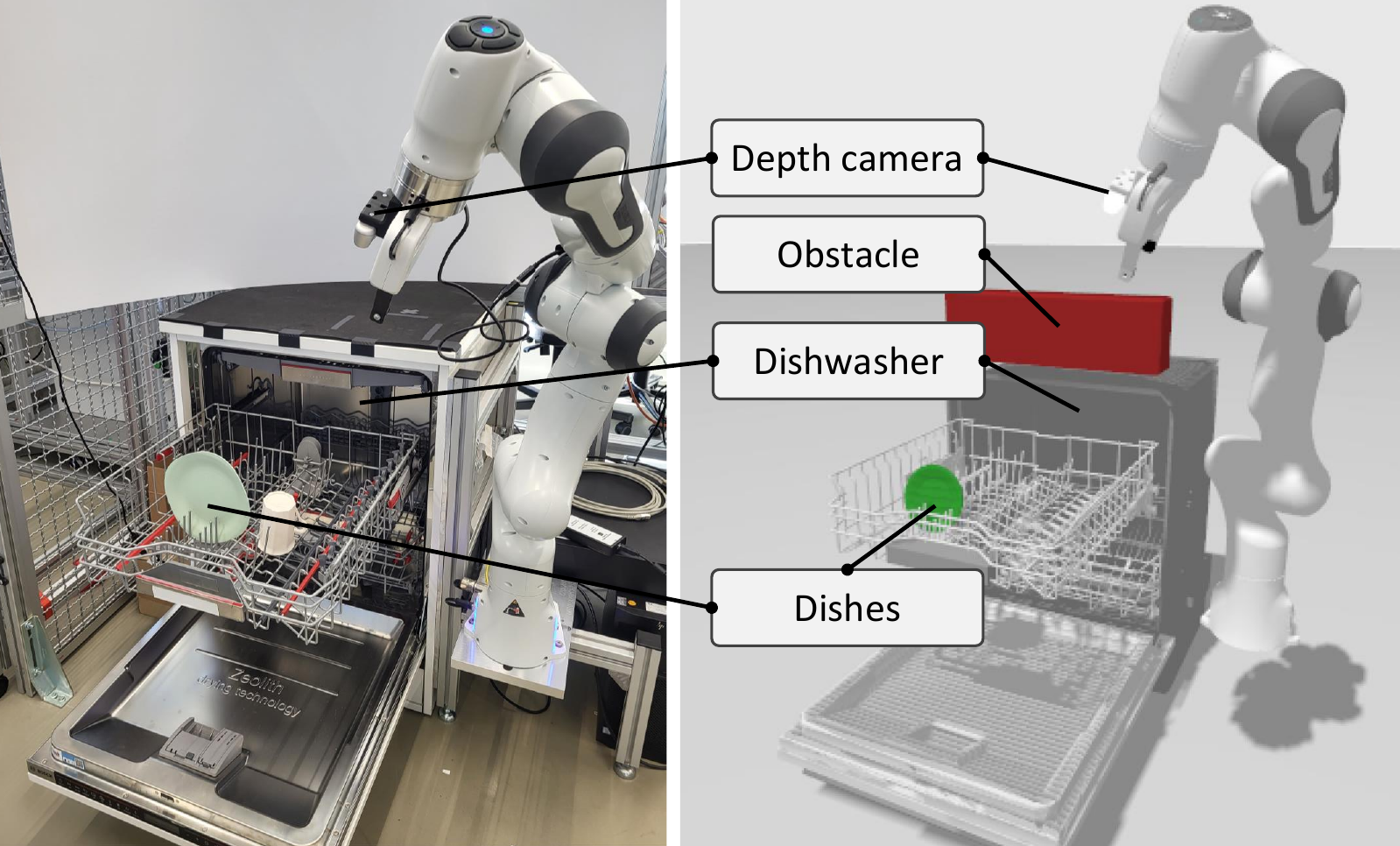}
    \caption{Real-world robot setup in a kitchen environment. It consists of a Franka-Emika Panda robot manipulator with a depth camera attached to its end-effector and a dishwasher filled with a few plastic dishes. The left image shows the real setup, while the right image displays the setup in a Gazebo simulation.}
    \label{fig:robot_setup}
\end{figure}

\subsubsection{Inference Time}
\begin{table}[tbp]
  \centering
  \caption{Comparison of the inference time $t_\text{step}$ in milliseconds for a single inference step for obstacle avoidance methods based on implicit obstacle representations. The obstacle representations are: \ac{SDF} of geometric primitive (GP) as ground truth, \ac{MLP} as standard architecture for \ac{SDF}~\cite{Gropp2020,Park2019}, \ac{BP} as polynomial \ac{SDF} architecture~\cite{YimlinLi2023}, and a Hilbert Map (HM)~\cite{Ramos2016}.}
  \label{tab:inference_time}
  \footnotesize
  % \resizebox{\linewidth}{!}{ % Consider \textwidth for better width distribution
    \begin{tabular}{l c c c c}
      %\toprule
      \rowcolor{gray!15} % Light gray for the header row
      \textbf{Methods} &  \textbf{GP} & \textbf{MLP} & \textbf{BP} & \textbf{HM} \\ 
      SDC/SDDC/DT & $2.2$ & $4.0$ & $9.0$ & $2.6$\\
      \rowcolor{gray!5} % Light gray for alternating rows
      MM & $1.7$ & $3.0$ & $5.7$ & -\\
      ARPF & $1.2$ & $2.5$ & $5.2$ & -\\
    \end{tabular}
  % }
\end{table}

Our goal here is to analyze the computational cost of the inference process, which is relevant when implementing these methods in real-world settings. As a reference, consider that \ac{NCDS} without obstacle avoidance has an inference time of $5.1 \text{ms}$.
As shown in Table~\ref{tab:inference_time}, \ac{MM} and \ac{ARPF} are the fastest obstacle avoidance methods, with inference times  $0.5\text{ms}-1.0\text{ms}$ faster than the proposed \ac{SDDC} and \ac{SDC}. However, the influence of the chosen \ac{SDF} method becomes evident. While a geometric primitive \ac{SDF} is computationally cheap, it is only applicable to objects that are known in advance and can be modeled mathematically. In contrast, both \ac{MLP} and \ac{BP} representations can handle more complex shapes, where the former provides a faster inference time, making it a preferred choice.
Compared to \ac{HM}, the \ac{MLP} architecture is slower. However, \ac{HM} consists of radial basis functions and learns an occupancy model~\cite{Ramos2016}. Consequently, \ac{HM} cannot be employed as a distance function and can only modulate the vector field locally, as used in the diffeomorphic transform approach~\citep{Zhi2022}. This limitation renders \ac{HM} unsuitable for complex scenes, and it is not possible to maintain a safe distance from obstacles. Additionally, the gradient of \ac{HM} does not provide a strict safety barrier, since any vector field stronger than the \ac{HM} gradient can potentially lead to collisions.

\begin{table}[tbp]
  \centering
  \caption{A comparison of the inference time $t_\text{flow}$ in milliseconds required for a single step in the diffeomorphism $\psi$ (see Equation \eqref{eq:sdt_flow_forward}), where $t_\text{Jacobian}$ represents the inference time needed to compute the Jacobian $\mathbf{J}_\psi$ for the \ac{SDC}, across different solver types (see Table~\ref{tab:flow_solver} in the Appendix). The following obstacle representations method are used: \ac{SDF} of geometric primitive (GP) as ground truth, \ac{MLP} as standard architecture for \ac{SDF}~\cite{Gropp2020}~\cite{Park2019}, \ac{BP} as polynomial \ac{SDF} architecture ~\cite{YimlinLi2023} and a Hilbert Map (HM) \cite{Ramos2016}.}
  \label{tab:inference_time_sdf}
  \footnotesize
  \resizebox{\linewidth}{!}{ % Consider \textwidth for better width distribution
    \begin{tabular}{lrrrr|rrrr}%\toprule
    \rowcolor{gray!15}
 & \multicolumn{4}{c}{$\mathbf{t}_\textbf{flow}$} & \multicolumn{4}{c}{$\mathbf{t}_\textbf{jacobian}$} \\
 \rowcolor{gray!15}
%\cmidrule{2-5} \cmidrule{6-9}
\rowcolor{gray!15}
\textbf{Solvers} & \textbf{GP} & \textbf{MLP} & \textbf{BP} & \textbf{HM} & \textbf{GP} & \textbf{MLP} & \textbf{BP} & \textbf{HM} \\
\rowcolor{gray!15}
\textbf{SDC/SDDC} &  &  &  &  &  &  &  & \\
Convex & $1.8$ & $3.4$ & $7.3$ & $1.9$ & $0.4$ & $0.5$ & $1.5$ & $0.6$\\
\rowcolor{gray!5}
Euler & $3.7$ & $6.7$ & $14.8$ & $3.2$ & $0.5$ & $0.8$ & $3.0$ & $0.9$\\
\makecell{Runge- \\ Kutta} & $10.2$ & $18.6$ & $40.5$ & $10.0$ & $1.0$ & $1.4$ & $5.7$ & $2.0$\\
%\bottomrule
\end{tabular}
  }
\end{table}

We also report the computational cost when evaluating the diffeomorphic transform for obstacle avoidance as in Eq.~\eqref{eq:sdt_transform}. We distinguish between the computational time of the flow $t_\text{flow}$ from Eq.~\eqref{eq:sdt_flow_forward} and the computational cost of the Jacobian of Eq.~\eqref{eq:SDDC}. Table~\ref{tab:inference_time_sdf} reports the  impact of the ODE solver on the overall inference time. While convex solvers are the fastest, they are restricted to convex surfaces. In contrast, the Runge--Kutta solver is the slowest, with inference times approximately five times higher than those of the convex solver. Furthermore, Table~\ref{tab:inference_time_sdf} shows that the largest portion of the total computation time stems from evaluating the flow in Equation~\eqref{eq:sdt_flow_forward}. This observation underlines the importance of selecting an efficient implicit distance function to reduce the computation time overhead. Notably, the calculation of the Jacobian has only a minor impact on the total inference time.

\subsection{Real Robot Experiments}
\label{sec:real_world_exp}
Here we aim at demonstrating the applicability of the proposed obstacle avoidance methods on real-world settings.

\subsubsection{Experimental Setup}
We use a $7$ DoF Franka-Emika Panda robot endowed with a depth camera. The robot workspace is a part of a kitchen environment featuring a dishwasher, as shown in Fig.~\ref{fig:robot_setup}.
\begin{figure*}[t]
    \centering
    \begin{subfigure}[b]{0.29\textwidth}
        \centering
        \includegraphics[width=\textwidth]{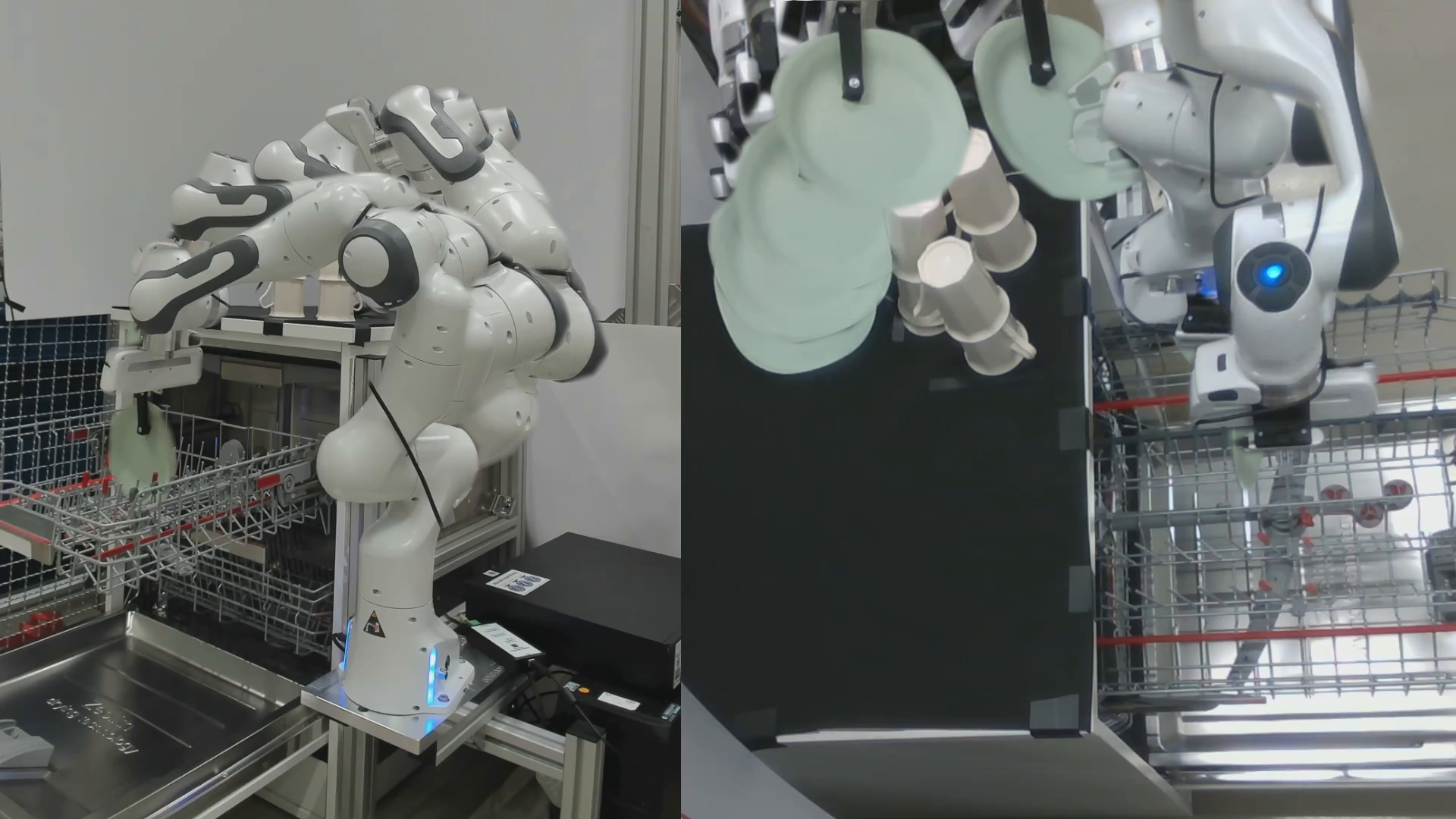}
        \caption{Motion sequence without extended \ac{SDF}}
        \label{fig:run_obs_rdf_ms}
    \end{subfigure}
    \vspace{0mm}
    \begin{subfigure}[b]{0.29\textwidth}
        \centering
        \includegraphics[width=\textwidth]{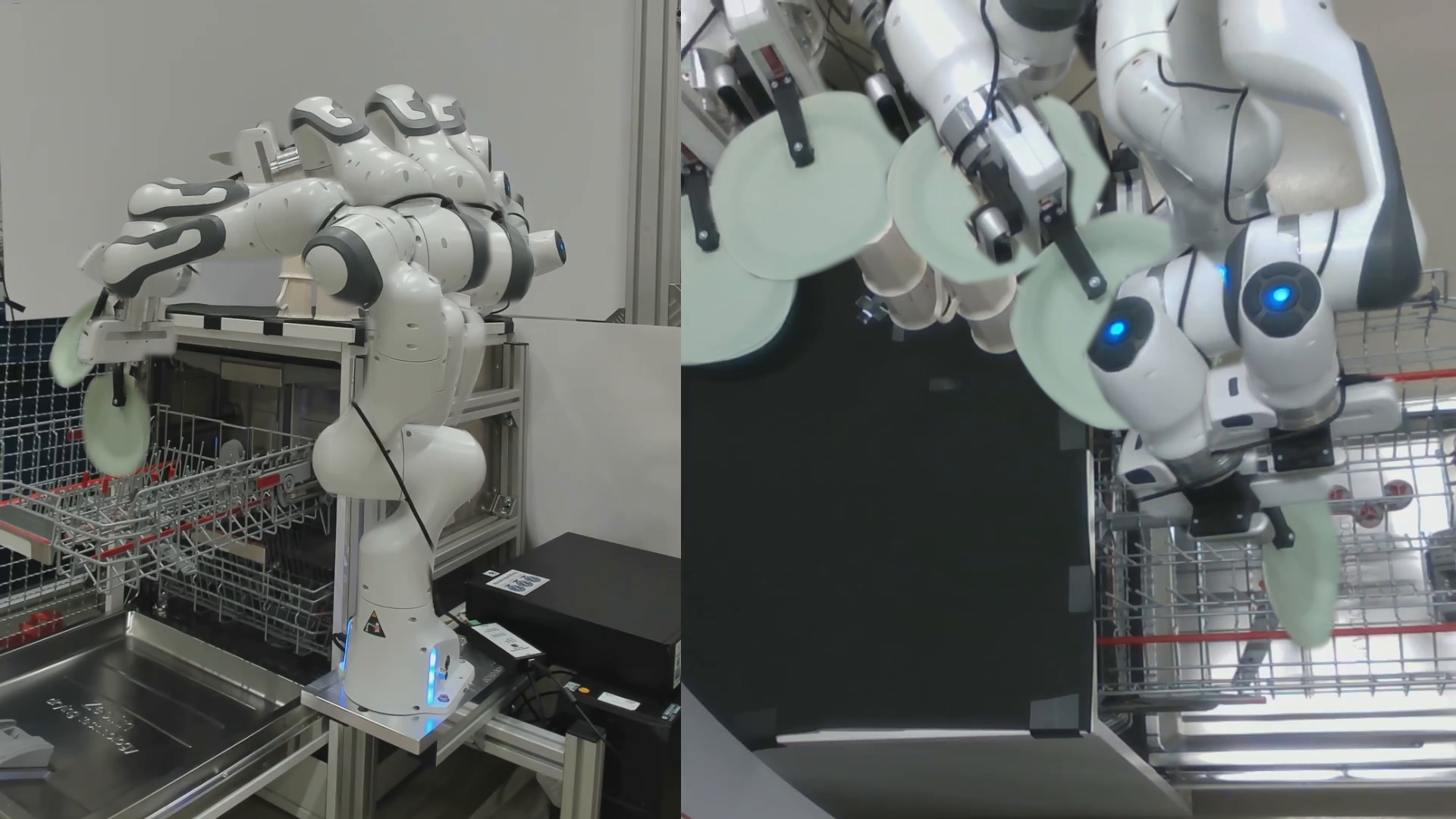}
        \caption{Motion sequence with extended \ac{SDF}}
        \label{fig:run_obs_rdf_ms_extend}
    \end{subfigure}
    \vspace{0mm}
    \begin{subfigure}[b]{0.39\textwidth}
        \centering
        \includegraphics[width=\textwidth]{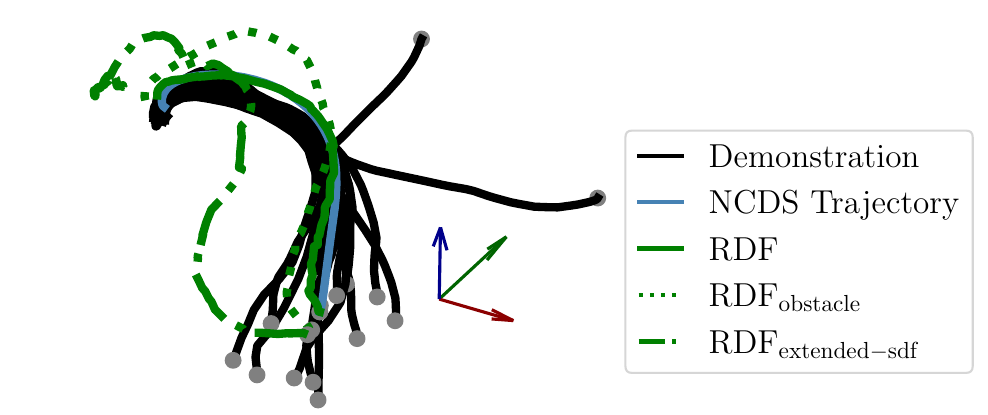}
        \caption{Comparison of task space trajectories}
        \label{fig:run_obs_rdf_traj_extend}
    \end{subfigure}
    \caption{\textsc{Emptying the dishwasher} task: A series of overlapped snapshots shows the \textsc{Place Plate} skill performed by the Franka-Emika Panda robot using an \ac{RDF} with grasped plate (left and middle panel). The right plot shows a comparison of the resulting trajectories using different methods w.r.t the provided demonstrations starting from the gray points.}
    \label{fig:run_obs_rdf_extend}
\end{figure*}
The robot is tasked to learn and execute two real-world tasks: \textsc{Emptying a dishwasher} and \textsc{Opening a dishwasher}.
In all two tasks, the robot should safely and autonomously interact with a dishwasher, maintaining contraction stability against external disturbances and ensuring collision-free operation even with grasped objects. In this context, the robot must not only reach specific target positions but also follow a defined motion trajectory to, for example,  remove dishes from the dishwasher and place them while executing a designated motion. Moreover, multiple obstacles, both unknown and cluttered, may be present along the robot’s path.

To model the obstacle, we sample a point cloud $\mathbf{P}=\{ \mathbf{p}_i \}_{i=0}^N$ of the robot workspace on the fly using the depth camera mounted at the robot end-effector. To prevent the robot surface points from being represented in the point cloud, the point cloud is filtered, defining a new point cloud $\mathbf{P}^{+}=\{\mathbf{p_i}\in \mathbf{P} \vert \Gamma_\mathcal{R} > 0\}$.
The sampled and filtered point clouds are merged incrementally and subsequently downsampled to a voxel size of $2\text{cm}$.
Using these processed point cloud data, an \ac{SDF} function of the robot's surface is used for obstacle avoidance by means of the proposed methods \ac{SDC} and \ac{SDDC}.

We collect the training data for each skill using kinesthetic teaching. The following skills were recorded: \textsc{Place Cup}, \textsc{Place Plate}, \textsc{Place Plate in Rack}, \textsc{Open Door}, \textsc{Grasp Doorhandle}, \textsc{Move Sine}. The recorded trajectories for these skills are shown in Fig.~\ref{fig:Dishwasher_Dataset} in the Appendix.
To execute a robot skill, the integrated output of the corresponding NCDS model is sent out as a reference to the robot joint impedance controller~\cite{Ott2008,Spong1987}, using a control frequency $f_\text{jimp}=5 \text{Hz}$.

\subsubsection{Real-world Obstacle Avoidance Behavior}
Given the set of contractive skills learned via NCDS, we evaluate our obstacle avoidance method in the tasks: \textsc{Open Dishwasher with Stacked Cups} and  \textsc{Place Plate}. Specifically, we test the \ac{SDT} pipeline with an \ac{RDF} as distance function and the \ac{SDC} method.
The robot's implicit distance function is extended to include the plate's \ac{SDF} that was previously learned via Equation~\eqref{eq:RDF}. The experiment results are shown in Fig.~\ref{fig:run_obs_rdf_extend}.
Specifically, Fig.~\ref{fig:run_obs_rdf_ms_extend} shows a successful obstacle avoidance behavior, without collisions with the stacked cups. Note that,
by closely analyzing Fig.~\ref{fig:run_obs_rdf_ms}, it can be observed that if the robot \ac{SDF} is not extended with implicit representation of the grasped object a collision can occur. Figure~\ref{fig:run_obs_rdf_traj_extend} shows the difference among the reproduced trajectories when the plate is attached (where the demonstrations are also plotted as a reference). Notably, the robot avoids both the cups and dishwasher base, through an initial path deviation. 

\subsubsection{Skill Modulation Robustness}
\begin{figure}[t]
    \centering
    \begin{subfigure}[b]{0.4\textwidth}
        \centering
        \includegraphics[width=\textwidth]{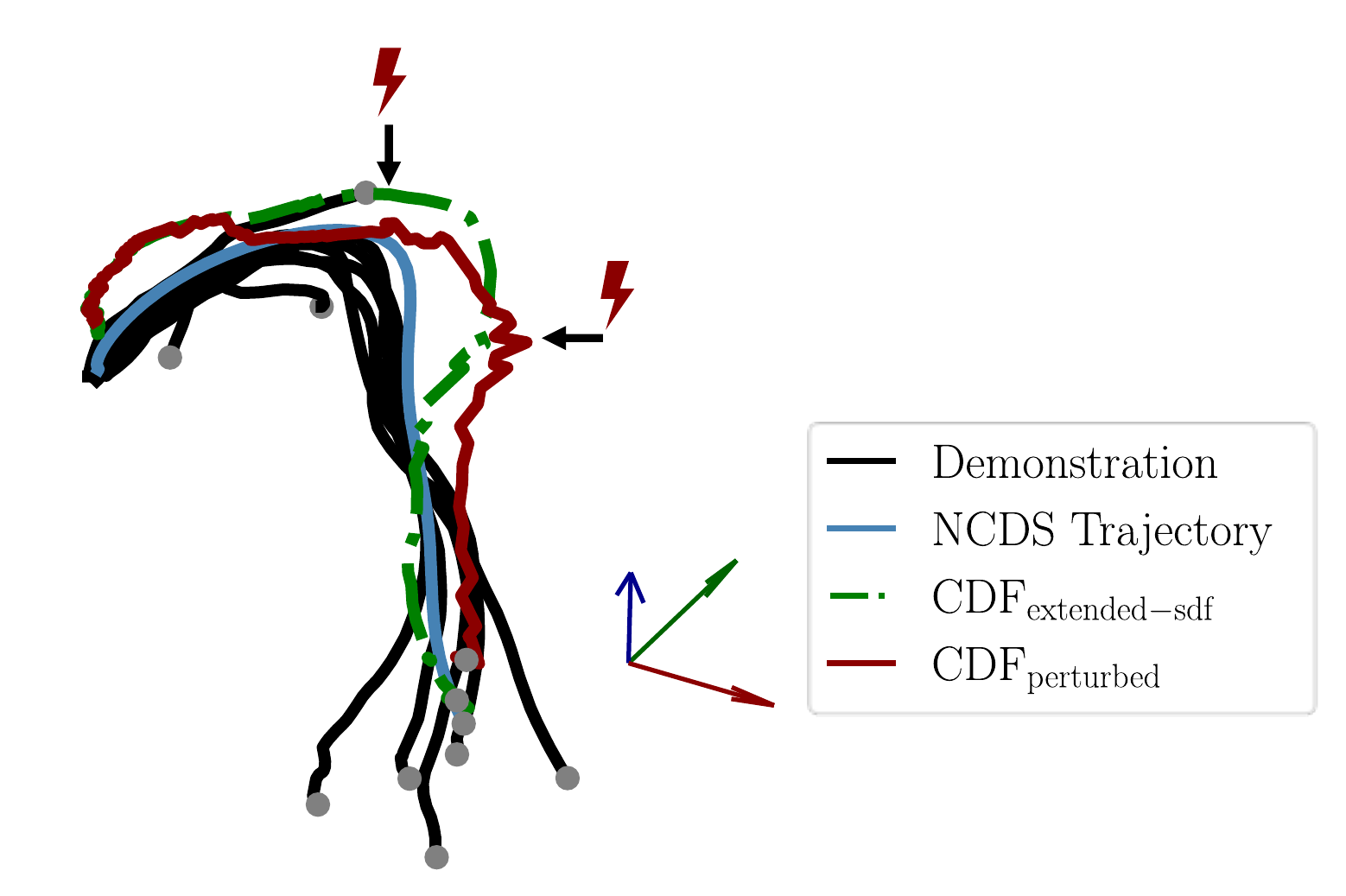}
        \caption{Position trajectories}
        \label{fig:pert_cdf_traj}
    \end{subfigure}
    \vspace{0mm}
    \begin{subfigure}[b]{0.48\textwidth}
        \centering
        \includegraphics[width=\textwidth]{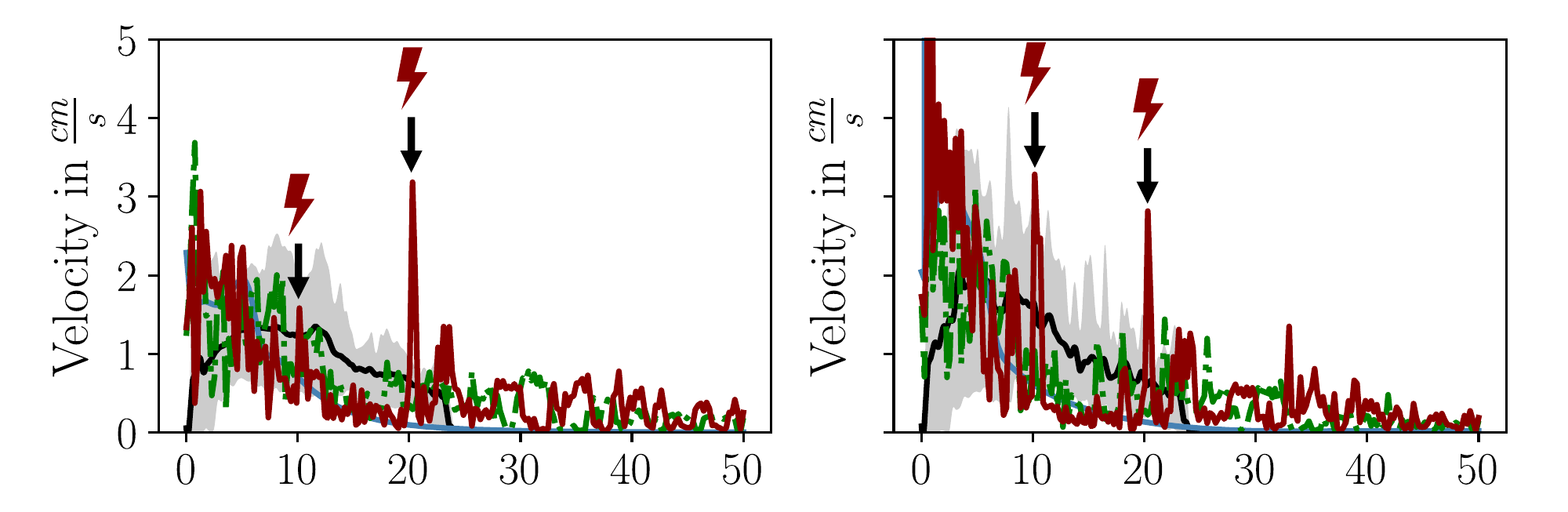}
        \caption{Velocity time series: Position (left), Orientation (right)}
        \label{fig:pert_cdf_vel}
    \end{subfigure}
    \caption{Adaptation robustness test for the \textsc{Place Plate in Rack} skill performed on the Frank-Emika Panda robot using \ac{CDF} and under perturbations at $10s$ and $21s$.}
    \label{fig:pert_cdf}
\end{figure}

We assess the skill robustness using the \textsc{Place Plate in Rack} skill in the \textsc{Open Dishwasher with Stacked Cups} task. The disturbances that we introduce are: First, moving one joint of the robot at $10s$; and second, directly moving the end-effector at $21s$. The results are shown in Fig.~\ref{fig:pert_cdf}, where it is possible to observe sudden velocity changes at $10s$ and $21s$ (see Fig.~\ref{fig:pert_cdf_vel}), matching the aforementioned perturbations. The peak at $23s$ suggests oscillatory behavior caused by the disturbance at $21s$, while the smaller peak at $10s$ reflects the disturbance in joint space, which only affects the orientation of the end-effector.
Figure~\ref{fig:pert_cdf_traj} shows that disturbances vanish over time. Initial differences may stem from variations in configuration, but the perturbed path converges to the non-disturbed trajectory, suggesting effective contraction.

\subsubsection{Inference Time}
\begin{figure}[t]
    \centering
    \includegraphics[width=0.38\textwidth]{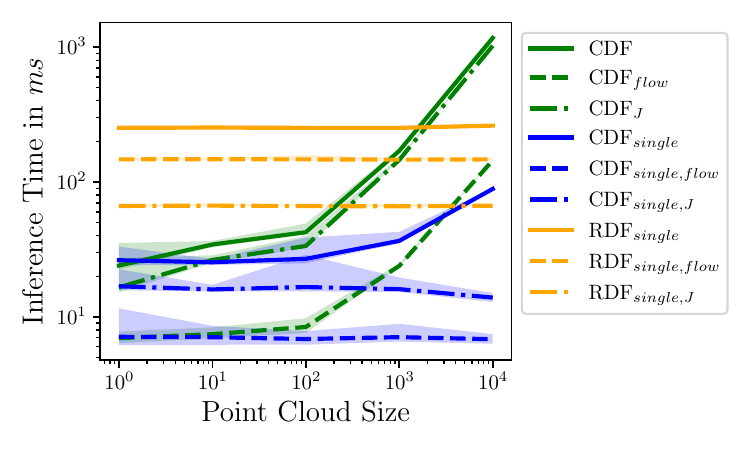}
    \caption{Comparison of the inference times for the \ac{SDC} method under three different approaches for computing the robot \ac{SDF}:
    (1)~\ac{CDF} evaluated at \emph{all} points,
    (2)~\ac{CDF} evaluated at a \emph{single closest} point,
    (3)~\ac{RDF} evaluated at a \emph{single closest} point.
    We decompose the total inference time into two components: The flow inference time and the inference time for computing the Jacobian of the diffeomorphism.}
    \label{fig:obs_inference_time}
\end{figure}

To assess the real-time capability of the proposed \ac{SDC} method, the computation time with respect to the point cloud size among different robot \ac{SDF} functions is reported in Fig.~\ref{fig:obs_inference_time}).

Note that for large point clouds, it is particularly beneficial to filter out only the nearest point from the input point cloud and then use this point to query the \ac{SDT}. Using this strategy, the flow and Jacobian computation times remain constant, and therefore the overall inference time increases only marginally, mainly because the robot \ac{SDF} must handle more query points. However, this additional overhead can be efficiently parallelized.

In contrast, when all points in the point cloud are used for inference, the computation time becomes unaffordable for real-time applications, particularly due to the increased computational load of the Jacobian calculation in the diffeomorphism. Furthermore, the \ac{CDF} exhibits significantly faster inference times overall and is therefore preferable to the \ac{RDF}.
In addition to the foregoing results, the inference time of the robot SDF as a function of the ODE solver is provided in Table~\ref{tab:inference_time_sdf_real_ex} in the Appendix.

\section{Discussion} 
\label{sec:Discussion}
Next we point out the limitations of the proposed approaches and elaborate on possible further research directions.

\subsection{Limitations}
\subsubsection{Concave Obstacles}
The experiments have shown that concave obstacle surfaces lead to local stability that limits the execution of robotic skills, indicating that only convex obstacles can be reliably avoided with the proposed methods. Although the minimal concave properties of the learned models, for example, local inaccuracies in the learned implicit distance function, did not show any local instability in the experiments, it could theoretically still occur.

\subsubsection{Inference Time}
To maintain the real-time capability of the framework, one is restricted by the size of the point cloud and the ODE solver used. This could result in less precise obstacle avoidance, especially when resources are limited.

\subsubsection{Discretization} The discretization of the system states could lead to a corner case. Specifically, if the sampling rate does not match the system dynamics, a collision might occur because the integrated state at the last time step could already lie within an obstacle. Although this issue is inherent to dynamical systems, it can be exacerbated if $s_\text{grad}$ in Eq.~\eqref{eq:barrier_advanced} makes the barrier too steep. Furthermore, an overly steep barrier might force the system to operate at high velocities, potentially exceeding system safe limits.

\subsection{Future Work}
\subsubsection{Concave Obstacles} In order to handle concave obstacles, \citet{Fourie2024} propose representing obstacles with convex properties using isosurface tracking and manifold modulation to handle non-convex obstacles. However, contraction preservation remains unproven and warrants future investigation.
In addition, the barrier function could be extended in such a way that the surface curvature is taken into account and has a stronger effect in non-convex regions in order to prevent local stability.

\subsubsection{Dynamic Obstacles} In addition to the static obstacles examined so far, dynamic obstacles can also be considered. In general, these have no influence on the stability of the presented approach as long as the sampling rate is significantly higher than the dynamic of the obstacle.
Learning the dynamics of the obstacle using an SDF is a substantial challenge and would have to be fundamentally investigated beforehand.
If the dynamics $\mathbf{q}_o$ of the obstacles in configuration space are known or estimated, we could potentially incorporate this into a barrier function $b_{\text{swept},o}$, analogous to the swept features in Eqs.~\eqref{eq:barrier_swept_r} and \eqref{eq:barrier_swept_s}, to adapt obstacle avoidance as a function of the relative motion of the obstacle. This function amplifies when the contractive system moves opposite to the direction of the obstacle, but relaxes when moving in the same direction,
\begin{align}
\label{eq:barrier_swept_o}
   b_{\text{swept},o}\left(\mathbf{q}_\text{SDT} , \mathbf{q}_o\right)&= \max\left(0, 1 - \frac{\dot{\mathbf{q}}_\text{SDT} \cdot \mathbf{q}_o}{\vert \dot{\mathbf{q}}_\text{SDT} \vert \ \vert \mathbf{q}_\text{SDT}  \vert} \right).
\end{align}
However, the exact effects or alternative approaches would have to be investigated in future research.

\subsubsection{Inference Time} To further optimize the computation time of the method for real-time use, it is recommended to consider only obstacles that are within the robot's operating space. 
Furthermore, incrementally learned \ac{SDF} functions \cite{Ortiz2022} could be leveraged to learn complex scenes as \ac{SDF} and use them for the proposed framework, which could be particularly relevant for mobile robot use cases. In addition, it would also be conceivable to use a model that directly predicts the distance between two surfaces in order to achieve even more precise obstacle avoidance and to become independent of any point cloud; the paper by \citet{goel2024} could be used as a inspiration.

\subsubsection{Control Barrier Functions} It may also be possible to extend our barrier function approach to include a \ac{CBF}~\citep{Ames2019} that dynamically accounts for system inputs.
The assumption is that the safe set defined by a \ac{CBF} $h(\mathbf{x})$~\cite[Eq. (5)]{Rodriguez2022}) coincides with the region outside obstacles exactly as given by the \ac{SDF}. In other words, each obstacle boundary is defined by the zero level set $\Gamma_\text{SDF}(\mathbf{x})=0$ and similarly the safe-set boundary is defined by $h(\mathbf{x})=0$.
This correspondence suggests that a \ac{CBF} constructed from an \ac{SDF} inherently embeds information regarding both the safe region and the direction of the boundary. We can therefore design a time dependent vector field guided by the gradient of the \ac{CBF}, whose flow induces a diffeomorphism that repels the system from unsafe areas.

\section{Conclusion} 
\label{sec:conclusion}

This paper introduces two novel contraction-preserving whole-body obstacle avoidance methods, namely the \acf{SDDC} and the \acf{SDC}. Both methods leverage learned implicit distance functions of the robot’s surface, combining their gradients with barrier functions to define a flow that maps learned contractive skills onto an obstacle-avoiding manifold. Specifically, the \ac{SDDC} relies on a differential coordinate change, while the \ac{SDC} employs a standard coordinate transformation enabled by the constructed flow.
Both approaches are particularly relevant for contractive robot skills or contractive dynamical systems in general, since most existing obstacle avoidance strategies fail to preserve the required stability properties. Furthermore, we propose new obstacle avoidance metrics that facilitate benchmarking and comparative evaluations across different methods. Experimental results show that the proposed approaches effectively preserve the underlying vector field and yield trajectories with minimal curvature changes. Real-world robot experiments confirm the practical applicability of the methods, underscoring the potential of the \ac{SDDC} and \ac{SDC} approaches for robust, contraction-preserving obstacle avoidance.

\bibliographystyle{plainnat}
\bibliography{references}

\begin{thebibliography}{71}
\providecommand{\natexlab}[1]{#1}
\providecommand{\url}[1]{\texttt{#1}}
\expandafter\ifx\csname urlstyle\endcsname\relax
  \providecommand{\doi}[1]{doi: #1}\else
  \providecommand{\doi}{doi: \begingroup \urlstyle{rm}\Url}\fi

\bibitem[Abyaneh et~al.(2025)Abyaneh, Boroujeni, Lin, and Ferrari-Trecate]{Abyaneh2025}
Amin Abyaneh, Mahrokh~Ghoddousi Boroujeni, Hsiu-Chin Lin, and Giancarlo Ferrari-Trecate.
\newblock Contractive dynamical imitation policies for efficient out-of-sample recovery.
\newblock In \emph{International Conference on Learning Representations (ICLR)}, 2025.
\newblock URL \url{https://openreview.net/forum?id=lILEtkWOXD}.

\bibitem[Ames et~al.(2019)Ames, Coogan, Egerstedt, Notomista, Sreenath, and Tabuada]{Ames2019}
Aaron~D. Ames, Samuel Coogan, Magnus Egerstedt, Gennaro Notomista, Koushil Sreenath, and Paulo Tabuada.
\newblock Control barrier functions: Theory and applications.
\newblock In \emph{European Control Conference (ECC)}, pages 3420--3431, 2019.
\newblock URL \url{https://doi.org/10.23919/ECC.2019.8796030}.

\bibitem[Argall et~al.(2009)Argall, Chernova, Veloso, and Browning]{Argall2009}
Brenna~D. Argall, Sonia Chernova, Manuela Veloso, and Brett Browning.
\newblock A survey of robot learning from demonstration.
\newblock \emph{Robotics and Autonomous Systems}, 57\penalty0 (5):\penalty0 469--483, 2009.
\newblock ISSN 0921-8890.
\newblock URL \url{https://doi.org/10.1016/j.robot.2008.10.024}.

\bibitem[Beik-Mohammadi et~al.(2024{\natexlab{a}})Beik-Mohammadi, Hauberg, Arvanitidis, Figueroa, Neumann, and Rozo]{BeikMohammadi2024}
Hadi Beik-Mohammadi, S{\o}ren Hauberg, Georgios Arvanitidis, Nadia Figueroa, Gerhard Neumann, and Leonel Rozo.
\newblock Neural contractive dynamical systems.
\newblock In \emph{International Conference on Learning Representations (ICLR)}, 2024{\natexlab{a}}.
\newblock URL \url{https://openreview.net/forum?id=iAYIRHOYy8}.

\bibitem[Beik-Mohammadi et~al.(2024{\natexlab{b}})Beik-Mohammadi, Hauberg, Arvanitidis, Neumann, and Rozo]{Beik-Mohammadi2024b}
Hadi Beik-Mohammadi, Søren Hauberg, Georgios Arvanitidis, Gerhard Neumann, and Leonel Rozo.
\newblock Extended neural contractive dynamical systems: On multiple tasks and {R}iemannian safety regions, 2024{\natexlab{b}}.
\newblock URL \url{https://arxiv.org/abs/2411.11405}.

\bibitem[Berndt and Clifford(1994)]{Berndt1994}
Donald~J. Berndt and James Clifford.
\newblock Using dynamic time warping to find patterns in time series.
\newblock In \emph{International Conference on Knowledge Discovery and Data Mining (KDD)}, page 359–370. AAAI Press, 1994.
\newblock URL \url{https://cdn.aaai.org/Workshops/1994/WS-94-03/WS94-03-031.pdf}.

\bibitem[Billard et~al.(2008)Billard, Calinon, Dillmann, and Schaal]{Billard2008}
Aude Billard, Sylvain Calinon, R{\"u}diger Dillmann, and Stefan Schaal.
\newblock \emph{Robot Programming by Demonstration}, pages 1371--1394.
\newblock Springer Berlin Heidelberg, Berlin, Heidelberg, 2008.
\newblock ISBN 978-3-540-30301-5.
\newblock URL \url{https://doi.org/10.1007/978-3-540-30301-5_60}.

\bibitem[Billard et~al.(2016)Billard, Calinon, and Dillmann]{Billard2016}
Aude~G. Billard, Sylvain Calinon, and R{\"u}diger Dillmann.
\newblock \emph{Learning from Humans}, pages 1995--2014.
\newblock Springer International Publishing, Cham, 2016.
\newblock ISBN 978-3-319-32552-1.
\newblock URL \url{https://doi.org/10.1007/978-3-319-32552-1_74}.

\bibitem[Blocher et~al.(2017)Blocher, Saveriano, and Lee]{Blocher2017}
Caroline Blocher, Matteo Saveriano, and Dongheui Lee.
\newblock Learning stable dynamical systems using contraction theory.
\newblock In \emph{International Conference on Ubiquitous Robots and Ambient Intelligence (URAI)}. IEEE, 2017.
\newblock \doi{10.1109/urai.2017.7992901}.
\newblock URL \url{http://dx.doi.org/10.1109/URAI.2017.7992901}.

\bibitem[Chen et~al.(2018)Chen, Rubanova, Bettencourt, and Duvenaud]{Chen2018}
Ricky T.~Q. Chen, Yulia Rubanova, Jesse Bettencourt, and David~K Duvenaud.
\newblock Neural ordinary differential equations.
\newblock In \emph{Advances in Neural Information Processing Systems (NeurIPS)}, 2018.
\newblock URL \url{https://proceedings.neurips.cc/paper_files/paper/2018/file/69386f6bb1dfed68692a24c8686939b9-Paper.pdf}.

\bibitem[Chen et~al.(2023)Chen, Gao, Yao, Niederhauser, Bekiroglu, and Billard]{Yiting2023}
Yiting Chen, Xiao Gao, Kunpeng Yao, Loïc Niederhauser, Yasemin Bekiroglu, and Aude Billard.
\newblock Differentiable robot neural distance function for adaptive grasp synthesis on a unified robotic arm-hand system, 2023.
\newblock URL \url{https://arxiv.org/abs/2309.16085}.

\bibitem[Dawson et~al.(2022)Dawson, Gao, and Fan]{Dawson2022SafeCW}
Charles Dawson, Sicun Gao, and Chuchu Fan.
\newblock Safe control with learned certificates: A survey of neural {L}yapunov, barrier, and contraction methods for robotics and control.
\newblock \emph{IEEE Transactions on Robotics (T-RO)}, 39:\penalty0 1749--1767, 2022.
\newblock URL \url{http://dx.doi.org/10.1109/TRO.2022.3232542}.

\bibitem[Fourie et~al.(2024)Fourie, Figueroa, and Shah]{Fourie2024}
Christopher~K. Fourie, Nadia Figueroa, and Julie~A. Shah.
\newblock On-manifold strategies for reactive dynamical system modulation with nonconvex obstacles.
\newblock \emph{IEEE Transactions on Robotics (T-RO)}, 40:\penalty0 2390--2409, 2024.
\newblock URL \url{https://doi.org/10.1109/TRO.2024.3378179}.

\bibitem[Goel and Tabib(2024)]{goel2024}
Kshitij Goel and Wennie Tabib.
\newblock Distance and collision probability estimation from gaussian surface models, 2024.
\newblock URL \url{https://arxiv.org/abs/2402.00186}.

\bibitem[Gropp et~al.(2020)Gropp, Yariv, Haim, Atzmon, and Lipman]{Gropp2020}
Amos Gropp, Lior Yariv, Niv Haim, Matan Atzmon, and Yaron Lipman.
\newblock Implicit geometric regularization for learning shapes.
\newblock In \emph{International Conference on Machine Learning (ICML)}, pages 3789--3799. PMLR, 2020.
\newblock URL \url{https://proceedings.mlr.press/v119/gropp20a.html}.

\bibitem[Hernandez~Moreno et~al.(2024)Hernandez~Moreno, Jansing, Polikarpov, Carmichael, and Deuse]{HernandezMoreno2024}
Victor Hernandez~Moreno, Steffen Jansing, Mikhail Polikarpov, Marc~G. Carmichael, and Jochen Deuse.
\newblock Obstacles and opportunities for learning from demonstration in practical industrial assembly: A systematic literature review.
\newblock \emph{Robotics and Computer-Integrated Manufacturing}, 86:\penalty0 102658, April 2024.
\newblock ISSN 0736-5845.
\newblock URL \url{http://dx.doi.org/10.1016/j.rcim.2023.102658}.

\bibitem[Huang et~al.(2019)Huang, Rozo, Silv{\'{e}}rio, and Caldwell]{Huang19:KMP}
Yanlong Huang, Leonel Rozo, Jo{\~{a}}o Silv{\'{e}}rio, and Darwin~G Caldwell.
\newblock {Kernelized movement primitives}.
\newblock \emph{International Journal of Robotics Research ({IJRR})}, 38\penalty0 (7):\penalty0 833--852, 2019.
\newblock URL \url{https://doi.org/10.1177/0278364919846363}.

\bibitem[Huber et~al.(2019)Huber, Billard, and Slotine]{Huber2019}
Lukas Huber, Aude Billard, and Jean-Jacques Slotine.
\newblock Avoidance of convex and concave obstacles with convergence ensured through contraction.
\newblock \emph{IEEE Robotics and Automation Letters (RA-L)}, 4\penalty0 (2):\penalty0 1462–1469, April 2019.
\newblock ISSN 2377-3774.
\newblock URL \url{http://dx.doi.org/10.1109/LRA.2019.2893676}.

\bibitem[Huber et~al.(2022)Huber, Slotine, and Billard]{Huber2022}
Lukas Huber, Jean-Jacques Slotine, and Aude Billard.
\newblock Avoiding dense and dynamic obstacles in enclosed spaces: Application to moving in crowds.
\newblock \emph{IEEE Transactions on Robotics (T-RO)}, 38\penalty0 (5):\penalty0 3113--3132, 2022.
\newblock \doi{10.1109/TRO.2022.3164789}.

\bibitem[Humphreys(1972)]{Humphreys1972}
James~E. Humphreys.
\newblock \emph{Introduction to Lie Algebras and Representation Theory}.
\newblock Springer New York, 1972.
\newblock ISBN 9781461263982.
\newblock \doi{10.1007/978-1-4612-6398-2}.
\newblock URL \url{http://dx.doi.org/10.1007/978-1-4612-6398-2}.

\bibitem[Ijspeert et~al.(2013)Ijspeert, Nakanishi, Hoffmann, Pastor, and Schaal]{Ijspeert13:dmp}
Auke~Jan Ijspeert, Jun Nakanishi, Heiko Hoffmann, Peter Pastor, and Stefan Schaal.
\newblock Dynamical movement primitives: Learning attractor models for motor behaviors.
\newblock \emph{Neural Computation}, 25:\penalty0 328--373, 2013.
\newblock URL \url{https://doi.org/10.1162/NECO_a_00393}.

\bibitem[Irshad et~al.(2024)Irshad, Comi, Lin, Heppert, Valada, Ambrus, Kira, and Tremblay]{Irshad2024}
Muhammad~Zubair Irshad, Mauro Comi, Yen-Chen Lin, Nick Heppert, Abhinav Valada, Rares Ambrus, Zsolt Kira, and Jonathan Tremblay.
\newblock Neural fields in robotics: A survey, 2024.
\newblock URL \url{https://arxiv.org/abs/2410.20220}.

\bibitem[Jaffe et~al.(2024)Jaffe, Davydov, Lapsekili, Singh, and Bullo]{Jaffe2024}
Sean Jaffe, Alexander Davydov, Deniz Lapsekili, Ambuj Singh, and Francesco Bullo.
\newblock Learning neural contracting dynamics: Extended linearization and global guarantees.
\newblock In \emph{Conference on Neural Information Processing Systems (NeurIPS)}, 2024.
\newblock URL \url{https://openreview.net/forum?id=YYnP3Xpv3y}.

\bibitem[J\"{u}ttler and Felis(2002)]{Jttler2002}
Bert J\"{u}ttler and Alf Felis.
\newblock Least-squares fitting of algebraic spline surfaces.
\newblock \emph{Advances in Computational Mathematics}, 17\penalty0 (1/2):\penalty0 135–152, 2002.
\newblock ISSN 1019-7168.
\newblock \doi{10.1023/a:1015200504295}.
\newblock URL \url{http://dx.doi.org/10.1023/A:1015200504295}.

\bibitem[Khansari-Zadeh and Billard(2011)]{Khansari-Zadeh2011}
Seyed~Mohammad Khansari-Zadeh and Aude Billard.
\newblock Learning stable nonlinear dynamical systems with {G}aussian mixture models.
\newblock \emph{IEEE Transactions on Robotics (T-RO)}, 27\penalty0 (5):\penalty0 943--957, 2011.
\newblock URL \url{http://dx.doi.org/10.1109/TRO.2011.2159412}.

\bibitem[Khansari-Zadeh and Billard(2012)]{KhansariZadeh2012}
Seyed~Mohammad Khansari-Zadeh and Aude Billard.
\newblock A dynamical system approach to realtime obstacle avoidance.
\newblock \emph{Autonomous Robots}, 32\penalty0 (4):\penalty0 433–454, March 2012.
\newblock ISSN 1573-7527.
\newblock URL \url{http://dx.doi.org/10.1007/s10514-012-9287-y}.

\bibitem[Khatib(1985)]{Khatib1985}
O.~Khatib.
\newblock Real-time obstacle avoidance for manipulators and mobile robots.
\newblock In \emph{IEEE International Conference on Robotics and Automation (ICRA)}, volume~2, pages 500--505, 1985.
\newblock URL \url{http://dx.doi.org/10.1109/ROBOT.1985.1087247}.

\bibitem[Kim and Khosla(1992)]{Kim1992}
Jin-Oh Kim and Pradeep~K. Khosla.
\newblock Real-time obstacle avoidance using harmonic potential functions.
\newblock \emph{IEEE Transactions on Robotics and Automation}, 8\penalty0 (3):\penalty0 338--349, 1992.
\newblock \doi{10.1109/70.143352}.

\bibitem[Klein et~al.(2023)Klein, Jaquier, Meixner, and Asfour]{Klein2023}
Holger Klein, Noémie Jaquier, Andre Meixner, and Tamim Asfour.
\newblock On the design of region-avoiding metrics for collision-safe motion generation on {R}iemannian manifolds.
\newblock In \emph{IEEE/RSJ International Conference on Intelligent Robots and Systems (IROS)}, pages 2346--2353, 2023.
\newblock URL \url{http://dx.doi.org/10.1109/IROS55552.2023.10341266}.

\bibitem[Koptev et~al.(2023)Koptev, Figueroa, and Billard]{Koptev2023}
Mikhail Koptev, Nadia Figueroa, and Aude Billard.
\newblock Neural joint space implicit signed distance functions for reactive robot manipulator control.
\newblock \emph{IEEE Robotics and Automation Letters (RA-L)}, 8\penalty0 (2):\penalty0 480–487, February 2023.
\newblock ISSN 2377-3774.
\newblock \doi{10.1109/lra.2022.3227860}.
\newblock URL \url{http://dx.doi.org/10.1109/LRA.2022.3227860}.

\bibitem[Koskinopoulou et~al.(2016)Koskinopoulou, Piperakis, and Trahanias]{Koskinopoulou16:LfD4HRC}
Maria Koskinopoulou, Stylianos Piperakis, and Panos Trahanias.
\newblock Learning from demonstration facilitates human-robot collaborative task execution.
\newblock In \emph{ACM/IEEE International Conference on Human-Robot Interaction (HRI)}, pages 59--66, 2016.
\newblock URL \url{https://doi.org/10.1109/HRI.2016.7451734}.

\bibitem[Lee(1997)]{Lee1997}
John~M. Lee.
\newblock \emph{Riemannian Manifolds}.
\newblock Springer New York, 1997.
\newblock ISBN 9780387227269.
\newblock \doi{10.1007/b98852}.
\newblock URL \url{http://dx.doi.org/10.1007/b98852}.

\bibitem[Lee(2012)]{Lee2012}
John~M. Lee.
\newblock \emph{Introduction to Smooth Manifolds}.
\newblock Springer New York, 2012.
\newblock ISBN 9781441999825.
\newblock \doi{10.1007/978-1-4419-9982-5}.
\newblock URL \url{http://dx.doi.org/10.1007/978-1-4419-9982-5}.

\bibitem[Lemme et~al.(2015)Lemme, Meirovitch, Khansari-Zadeh, Flash, Billard, and Steil]{Lemme2015:LasaDataset}
A.~Lemme, Y.~Meirovitch, M.~Khansari-Zadeh, T.~Flash, A.~Billard, and J.~J. Steil.
\newblock Open-source benchmarking for learned reaching motion generation in robotics.
\newblock \emph{Paladyn, Journal of Behavioral Robotics}, 6\penalty0 (1), 2015.
\newblock URL \url{http://doi.org/10.1515/pjbr-2015-0002}.

\bibitem[Li et~al.(2024{\natexlab{a}})Li, Chi, Razmjoo, and Calinon]{Yiming2024}
Yiming Li, Xuemin Chi, Amirreza Razmjoo, and Sylvain Calinon.
\newblock Configuration space distance fields for manipulation planning.
\newblock In \emph{Proc.\ Robotics: Science and Systems ({R:SS})}, 2024{\natexlab{a}}.
\newblock URL \url{https://arxiv.org/abs/2406.01137}.

\bibitem[Li et~al.(2024{\natexlab{b}})Li, Zhang, Razmjoo, and Calinon]{YimlinLi2023}
Yiming Li, Yan Zhang, Amirreza Razmjoo, and Sylvain Calinon.
\newblock Representing robot geometry as distance fields: Applications to whole-body manipulation.
\newblock In \emph{IEEE International Conference on Robotics and Automation (ICRA)}, pages 15351--15357, 2024{\natexlab{b}}.
\newblock URL \url{http://dx.doi.org/10.1109/ICRA57147.2024.10611674}.

\bibitem[Liu et~al.(2022)Liu, Zhang, Tateo, Jauhri, Peters, and Chalvatzaki]{Liu2022}
Puze Liu, Kuo Zhang, Davide Tateo, Snehal Jauhri, Jan Peters, and Georgia Chalvatzaki.
\newblock Regularized deep signed distance fields for reactive motion generation.
\newblock In \emph{IEEE/RSJ International Conference on Intelligent Robots and Systems (IROS)}, pages 6673--6680, 2022.
\newblock URL \url{https://doi.org/10.1109/IROS47612.2022.9981456}.

\bibitem[Lohmiller and Slotine(1998)]{LOHMILLER1998}
Winfried Lohmiller and Jean-Jacques~E. Slotine.
\newblock On contraction analysis for non-linear systems.
\newblock \emph{Automatica}, 34\penalty0 (6):\penalty0 683--696, 1998.
\newblock ISSN 0005-1098.
\newblock URL \url{https://doi.org/10.1016/S0005-1098(98)00019-3}.

\bibitem[Lorraine and Hossain(2019)]{Lorraine2019}
Jonathan Lorraine and Safwan Hossain.
\newblock {JacNet}: Learning functions with structured {J}acobians.
\newblock \emph{INNF Workshop at the International Conference on Machine Learning (ICML)}, 2019.
\newblock URL \url{https://invertibleworkshop.github.io/INNF_2019/accepted_papers/pdfs/INNF_2019_paper_10.pdf}.

\bibitem[Lozano-Perez(1983)]{ALozano-Perez1983}
Tomas Lozano-Perez.
\newblock Robot programming.
\newblock \emph{Proceedings of the IEEE}, 71\penalty0 (7):\penalty0 821--841, 1983.
\newblock \doi{10.1109/PROC.1983.12681}.

\bibitem[Manchester and Slotine(2017)]{Manchester2017}
Ian~R. Manchester and Jean-Jacques~E. Slotine.
\newblock Control contraction metrics: Convex and intrinsic criteria for nonlinear feedback design.
\newblock \emph{IEEE Transactions on Automatic Control}, 62\penalty0 (6):\penalty0 3046--3053, 2017.
\newblock URL \url{http://dx.doi.org/10.1109/TAC.2017.2668380}.

\bibitem[Marić et~al.(2024)Marić, Li, and Calinon]{Maric2024}
Ante Marić, Yiming Li, and Sylvain Calinon.
\newblock Online learning of piecewise polynomial signed distance fields for manipulation tasks, 2024.
\newblock URL \url{https://arxiv.org/abs/2401.07698}.

\bibitem[Marticorena et~al.(2024)Marticorena, Fischer, Haviland, and Suenderhauf]{Marticorena2024}
Nicolas Marticorena, Tobias Fischer, Jesse Haviland, and Niko Suenderhauf.
\newblock Rmmi: Enhanced obstacle avoidance for reactive mobile manipulation using an implicit neural map, 2024.
\newblock URL \url{https://arxiv.org/abs/2408.16206}.

\bibitem[Neumann and Steil(2015)]{NEUMANN2015}
Klaus Neumann and Jochen~J. Steil.
\newblock Learning robot motions with stable dynamical systems under diffeomorphic transformations.
\newblock \emph{Robotics and Autonomous Systems (RAS)}, 70:\penalty0 1--15, 2015.
\newblock ISSN 0921-8890.
\newblock URL \url{https://doi.org/10.1016/j.robot.2015.04.006}.

\bibitem[Ortiz et~al.(2022)Ortiz, Clegg, Dong, Sucar, Novotny, Zollhoefer, and Mukadam]{Ortiz2022}
Joseph Ortiz, Alexander Clegg, Jing Dong, Edgar Sucar, David Novotny, Michael Zollhoefer, and Mustafa Mukadam.
\newblock {iSDF}: Real-time neural signed distance fields for robot perception.
\newblock In \emph{Robotics: Science and Systems (R:SS)}, 2022.
\newblock URL \url{https://www.roboticsproceedings.org/rss18/p012.pdf}.

\bibitem[Ott et~al.(2008)Ott, Albu-Schaffer, Kugi, and Hirzinger]{Ott2008}
Christian Ott, Alin Albu-Schaffer, Andreas Kugi, and Gerd Hirzinger.
\newblock On the passivity-based impedance control of flexible joint robots.
\newblock \emph{IEEE Transactions on Robotics (T-RO)}, 24\penalty0 (2):\penalty0 416--429, 2008.
\newblock URL \url{https://doi.org/10.1109/TRO.2008.915438}.

\bibitem[Park et~al.(2019)Park, Florence, Straub, Newcombe, and Lovegrove]{Park2019}
Jeong~Joon Park, Peter Florence, Julian Straub, Richard Newcombe, and Steven Lovegrove.
\newblock {DeepSDF}: Learning continuous signed distance functions for shape representation.
\newblock In \emph{IEEE/CVF Conference on Computer Vision and Pattern Recognition (CVPR)}, pages 165--174, June 2019.
\newblock URL \url{http://dx.doi.org/10.1109/CVPR.2019.00025}.

\bibitem[Pressley(2010)]{Pressley2010}
Andrew Pressley.
\newblock \emph{Elementary Differential Geometry}.
\newblock Springer London, 2010.
\newblock ISBN 9781848828919.
\newblock \doi{10.1007/978-1-84882-891-9}.
\newblock URL \url{http://dx.doi.org/10.1007/978-1-84882-891-9}.

\bibitem[Pujol and Chica(2023)]{Pujol2023}
Eduard Pujol and Antonio Chica.
\newblock Adaptive approximation of signed distance fields through piecewise continuous interpolation.
\newblock \emph{Computers and Graphics}, 114:\penalty0 337--346, 2023.
\newblock ISSN 0097-8493.
\newblock \doi{https://doi.org/10.1016/j.cag.2023.06.020}.
\newblock URL \url{https://www.sciencedirect.com/science/article/pii/S0097849323001139}.

\bibitem[Ramos and Ott(2016)]{Ramos2016}
Fabio Ramos and Lionel Ott.
\newblock Hilbert maps: Scalable continuous occupancy mapping with stochastic gradient descent.
\newblock \emph{The International Journal of Robotics Research (IJRR)}, 35\penalty0 (14):\penalty0 1717–1730, December 2016.
\newblock ISSN 1741-3176.
\newblock URL \url{http://dx.doi.org/10.1177/0278364916684382}.

\bibitem[Rana et~al.(2020)Rana, Li, Fox, Boots, Ramos, and Ratliff]{Rana2020}
Muhammad~Asif Rana, Anqi Li, Dieter Fox, Byron Boots, Fabio Ramos, and Nathan Ratliff.
\newblock Euclideanizing flows: Diffeomorphic reduction for learning stable dynamical systems.
\newblock In \emph{Conference on Learning for Dynamics and Control (L4DC)}, pages 630--639, 2020.
\newblock URL \url{https://proceedings.mlr.press/v120/rana20a.html}.

\bibitem[Ravichandar and Dani(2015)]{Ravichandar2015}
Harish Ravichandar and Ashwin Dani.
\newblock Learning contracting nonlinear dynamics from human demonstration for robot motion planning.
\newblock In \emph{ASME Dynamic Systems, and Control Conference (DSCC)}. American Society of Mechanical Engineers, October 2015.
\newblock \doi{10.1115/dscc2015-9870}.
\newblock URL \url{http://dx.doi.org/10.1115/DSCC2015-9870}.

\bibitem[Ravichandar et~al.(2020)Ravichandar, Polydoros, Chernova, and Billard]{Ravichandar2020}
Harish Ravichandar, Athanasios~S. Polydoros, Sonia Chernova, and Aude Billard.
\newblock Recent advances in robot learning from demonstration.
\newblock \emph{Annual Review of Control, Robotics, and Autonomous Systems}, 3\penalty0 (1):\penalty0 297–330, May 2020.
\newblock ISSN 2573-5144.
\newblock \doi{10.1146/annurev-control-100819-063206}.
\newblock URL \url{http://dx.doi.org/10.1146/annurev-control-100819-063206}.

\bibitem[Ravichandar and Dani(2018)]{Ravichandar2018}
Harish~chaandar Ravichandar and Ashwin Dani.
\newblock Learning position and orientation dynamics from demonstrations via contraction analysis.
\newblock \emph{Autonomous Robots}, 43\penalty0 (4):\penalty0 897–912, May 2018.
\newblock ISSN 1573-7527.
\newblock \doi{10.1007/s10514-018-9758-x}.
\newblock URL \url{http://dx.doi.org/10.1007/s10514-018-9758-x}.

\bibitem[Rezazadeh et~al.(2022)Rezazadeh, Kolarich, Kia, and Mehr]{Rezazadeh2022}
Navid Rezazadeh, Maxwell Kolarich, Solmaz~S. Kia, and Negar Mehr.
\newblock Learning contraction policies from offline data.
\newblock \emph{IEEE Robotics and Automation Letters (RA-L)}, 7\penalty0 (2):\penalty0 2905--2912, 2022.
\newblock URL \url{https://doi.org/10.1109/LRA.2022.3145100}.

\bibitem[Rodriguez et~al.(2022)Rodriguez, Csomay-Shanklin, Yue, and Ames]{Rodriguez2022}
Ivan Dario~Jimenez Rodriguez, Noel Csomay-Shanklin, Yisong Yue, and Aaron~D. Ames.
\newblock Neural gaits: Learning bipedal locomotion via control barrier functions and zero dynamics policies.
\newblock In \emph{Learning for Dynamics and Control Conference (L4DC)}, volume 168, pages 1060--1072, 2022.
\newblock URL \url{https://proceedings.mlr.press/v168/rodriguez22a.html}.

\bibitem[Rozo et~al.(2016)Rozo, Calinon, Caldwell, Jiménez, and Torras]{Rozo16:pHRC_LfD}
Leonel Rozo, Sylvain Calinon, Darwin~G. Caldwell, Pablo Jiménez, and Carme Torras.
\newblock Learning physical collaborative robot behaviors from human demonstrations.
\newblock \emph{IEEE Transactions on Robotics (T-RO)}, 32\penalty0 (3):\penalty0 513--527, 2016.
\newblock URL \url{https://doi.org/10.1109/TRO.2016.2540623}.

\bibitem[Rozo et~al.(2018)Rozo, Amor, Calinon, Dragan, and Lee]{Rozo2018:EditorialHRC}
Leonel Rozo, Heni~Ben Amor, Sylvain Calinon, Anca Dragan, and Dongheui Lee.
\newblock Special issue on learning for human–robot collaboration.
\newblock \emph{Autonomous Robots}, 42\penalty0 (5):\penalty0 953--956, 2018.
\newblock URL \url{https://doi.org/10.1007/s10514-018-9756-z}.

\bibitem[Rozo et~al.(2024)Rozo, Kupcsik, Schillinger, Guo, Krug, {van Duijkeren}, Spies, Kesper, Hoppe, Ziesche, Bürger, and Arras]{Rozo2024}
Leonel Rozo, Andras~G. Kupcsik, Philipp Schillinger, Meng Guo, Robert Krug, Niels {van Duijkeren}, Markus Spies, Patrick Kesper, Sabrina Hoppe, Hanna Ziesche, Mathias Bürger, and Kai~O. Arras.
\newblock The e-bike motor assembly: Towards advanced robotic manipulation for flexible manufacturing.
\newblock \emph{Robotics and Computer-Integrated Manufacturing}, 85:\penalty0 102637, 2024.
\newblock ISSN 0736-5845.
\newblock \doi{https://doi.org/10.1016/j.rcim.2023.102637}.
\newblock URL \url{https://www.sciencedirect.com/science/article/pii/S0736584523001126}.

\bibitem[Schaal(1996)]{Schaal1996}
Stefan Schaal.
\newblock Learning from demonstration.
\newblock In \emph{Neural Information Processing Systems (NeurIPS)}, 1996.
\newblock URL \url{https://proceedings.neurips.cc/paper_files/paper/1996/file/68d13cf26c4b4f4f932e3eff990093ba-Paper.pdf}.

\bibitem[Simpson-Porco and Bullo(2014)]{SimpsonPorco2014}
John~W. Simpson-Porco and Francesco Bullo.
\newblock Contraction theory on riemannian manifolds.
\newblock \emph{Systems \& Control Letters}, 65:\penalty0 74--80, 2014.
\newblock ISSN 0167-6911.
\newblock URL \url{https://doi.org/10.1016/j.sysconle.2013.12.016}.

\bibitem[Singh et~al.(2021)Singh, Richards, Sindhwani, Slotine, and Pavone]{Singh2021}
Sumeet Singh, Spencer~M Richards, Vikas Sindhwani, Jean-Jacques~E Slotine, and Marco Pavone.
\newblock Learning stabilizable nonlinear dynamics with contraction-based regularization.
\newblock \emph{The International Journal of Robotics Research (IJRR)}, 40\penalty0 (10-11):\penalty0 1123--1150, 2021.
\newblock URL \url{https://doi.org/10.1177/0278364920949931}.

\bibitem[Spong(1987)]{Spong1987}
M.~W. Spong.
\newblock Modeling and control of elastic joint robots.
\newblock \emph{Journal of Dynamic Systems, Measurement, and Control}, 109\penalty0 (4):\penalty0 310–318, December 1987.
\newblock ISSN 1528-9028.
\newblock URL \url{http://dx.doi.org/10.1115/1.3143860}.

\bibitem[Su et~al.(2021)Su, Mariani, Ovur, Menciassi, Ferrigno, and De~Momi]{Su2021}
Hang Su, Andrea Mariani, Salih~Ertug Ovur, Arianna Menciassi, Giancarlo Ferrigno, and Elena De~Momi.
\newblock Toward teaching by demonstration for robot-assisted minimally invasive surgery.
\newblock \emph{IEEE Transactions on Automation Science and Engineering}, 18\penalty0 (2):\penalty0 484--494, 2021.
\newblock URL \url{https://doi.org/10.1109/TASE.2020.3045655}.

\bibitem[Sun et~al.(2020)Sun, Jha, and Fan]{sun2020}
Dawei Sun, Susmit Jha, and Chuchu Fan.
\newblock Learning certified control using contraction metric.
\newblock In \emph{Conference on Robot Learning (CoRL)}, pages 1519--1539, 2020.
\newblock URL \url{https://proceedings.mlr.press/v155/sun21b.html}.

\bibitem[Tsukamoto and Chung(2021)]{Tsukamoto2021b}
Hiroyasu Tsukamoto and Soon-Jo Chung.
\newblock Neural contraction metrics for robust estimation and control: A convex optimization approach.
\newblock \emph{IEEE Control Systems Letters}, 5\penalty0 (1):\penalty0 211--216, 2021.
\newblock URL \url{https://doi.org/10.1109/LCSYS.2020.3001646}.

\bibitem[Tsukamoto et~al.(2021)Tsukamoto, Chung, and Slotine]{Tsukamoto2021}
Hiroyasu Tsukamoto, Soon-Jo Chung, and Jean-Jacques~E. Slotine.
\newblock Contraction theory for nonlinear stability analysis and learning-based control: A tutorial overview.
\newblock \emph{Annual Reviews in Control}, 52:\penalty0 135–169, 2021.
\newblock ISSN 1367-5788.

\bibitem[Urain et~al.(2020)Urain, Ginesi, Tateo, and Peters]{Urain2020}
J.~Urain, M.~Ginesi, D.~Tateo, and J.~Peters.
\newblock Imitationflow: Learning deep stable stochastic dynamic systems by normalizing flows.
\newblock In \emph{International Conference on Intelligent Robots and Systems (IROS)}, pages 5231--5237. IEEE, 2020.
\newblock URL \url{https://doi.org/10.1109/IROS45743.2020.9341035}.

\bibitem[Welschehold et~al.(2016)Welschehold, Dornhege, and Burgard]{Welschehold16:LfDhousehold}
Tim Welschehold, Christian Dornhege, and Wolfram Burgard.
\newblock Learning manipulation actions from human demonstrations.
\newblock In \emph{International Conference on Intelligent Robots and Systems (IROS)}, pages 3772--3777, 2016.
\newblock URL \url{https://doi.org/10.1109/IROS.2016.7759555}.

\bibitem[Zhang et~al.(2022)Zhang, Mohammadi, and Rozo]{Zhang2022}
Jiechao Zhang, Hadi~Beik Mohammadi, and Leonel Rozo.
\newblock Learning {R}iemannian stable dynamical systems via diffeomorphisms.
\newblock In \emph{Conference on Robot Learning (CoRL)}, 2022.
\newblock URL \url{https://openreview.net/forum?id=o8dLx8OVcNk}.

\bibitem[Zhi et~al.(2022)Zhi, Lai, Ott, and Ramos]{Zhi2022}
Weiming Zhi, Tin Lai, Lionel Ott, and Fabio Ramos.
\newblock Diffeomorphic transforms for generalised imitation learning.
\newblock In \emph{Learning for Dynamics and Control Conference (L4DC)}, pages 508--519, 2022.
\newblock URL \url{https://proceedings.mlr.press/v168/zhi22a.html}.

\end{thebibliography}

\clearpage
\appendix
\subsection{Signed Distance Fields}
\subsubsection{Loss Functions} 
According to \citet{Park2019}, learned \ac{SDF} methods with parameters $\boldsymbol{\theta}$ can be interpreted as a decision boundary that assigns points to either the inside $\mathcal{H}^i$ or outside $\mathcal{H}^f$ of an object, while additionally providing the distance to the surface.
To achieve these properties, an appropriate loss function must be chosen. \citet{Park2019} propose using an $\mathcal{L}_1$ reconstruction loss,
\begin{equation}
    \mathcal{L}_\text{SDF}(\Gamma_\text{SDF}(\mathbf{x};\boldsymbol{\theta})) = \left\vert \text{clamp}\left(\Gamma_\text{SDF}(\mathbf{x};\boldsymbol{\theta}), \delta\right) - \text{clamp}(s, \delta) \right\vert ,
\end{equation}
where $\text{clamp}(x,\delta):= \min(\delta,\max(-\delta,x))$ limits the values within the range $[-\delta, \delta]$, and $s$ defines a boundary for the region over which the \ac{SDF} function should be defined.

Furthermore, \citet{Gropp2020} suggested using the Eikonal loss,
\begin{equation}
    \mathcal{L}_\text{eik}(\Gamma_\text{SDF}(\mathbf{x};\boldsymbol{\theta})) = \left\vert \Vert \nabla_\mathbf{x}\Gamma_\text{SDF}(\mathbf{x};\boldsymbol{\theta}) \Vert - 1 \right\vert ,
\end{equation}
which encourages the \ac{SDF} to maintain equidistance properties across the domain, meaning that for any point in space, the value of the \ac{SDF} corresponds to the shortest distance to the surface $\mathcal{H}^s$~\cite{Gropp2020}.
In addition, \citet{Ortiz2022} introduced a gradient-based loss,
\begin{equation}
    \mathcal{L}_\text{grad}(\Gamma_\text{SDF}(\mathbf{x};\boldsymbol{\theta}), \mathbf{g}) = 1 - \frac{\nabla_\mathbf{x}\Gamma_\text{SDF}(\mathbf{x};\boldsymbol{\theta}) \cdot \mathbf{g}}{\Vert \nabla_\mathbf{x}\Gamma_\text{SDF}(\mathbf{x};\boldsymbol{\theta}) \Vert \Vert \mathbf{g} \Vert} ,
\end{equation}
which penalizes deviations of the gradient of the \ac{SDF} from the surface normal by computing the cosine distance between the predicted gradient and the true normal vector $\mathbf{g}$.
\citet{Jttler2002} additionally introduced the concept of tension loss, which enhances the smoothness of the function.
\begin{equation}
    \mathcal{L}_\text{tension}(\Gamma_\text{SDF}(\mathbf{x};\boldsymbol{\theta})) = \Vert \nabla^{2}_\mathbf{x}\Gamma_\text{SDF}(\mathbf{x};\boldsymbol{\theta}) \Vert ^{2}
\end{equation}

\subsection{Contraction-preserving Obstacle Avoidance}
\subsubsection{Multi-modal demonstrations}
\begin{SCfigure}[0.5][b]
\caption{Obstacle avoidance with SDC using an inverse barrier with swept features on an NCDS model trained on a multi-modal LASA dataset skill. Two circular obstacles are depicted by red lines.}
\includegraphics[width=0.3\textwidth]{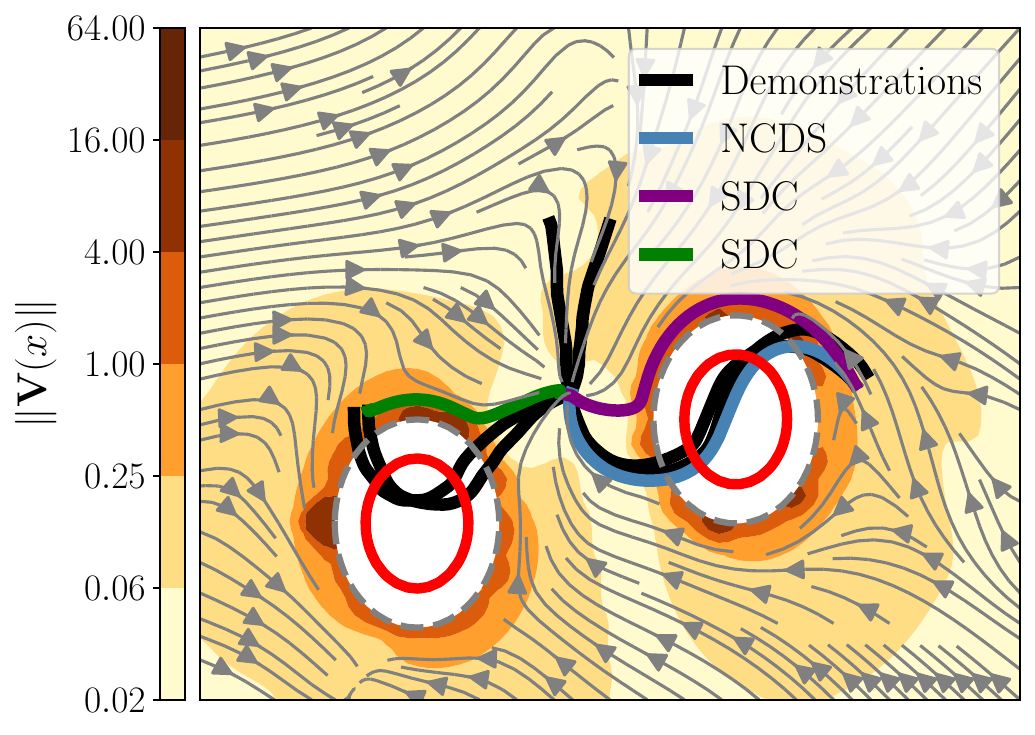}
\label{fig:SDT_multi_modal}

\vspace{-2em}

\end{SCfigure}

Figure~\ref{fig:SDT_multi_modal} displays the evaluation of the SDC on a multi-modal LASA dataset skill. Under a high contraction rate, the green trajectory remains in its initial modal state. Conversely, a reduced contraction rate, as evidenced by the purple trajectory, can induce a mode switch.

\subsubsection{Flow Solver}
Table~\ref{tab:flow_solver} shows a overview of different types of solvers that can be used to solve the flow in Eq.~\eqref{eq:sdt_flow_forward}.

\begin{table}[tbp]
    \centering
    \caption{Overview of various ODE solvers}
    \label{tab:flow_solver}
    \renewcommand{\arraystretch}{1.2}
    \begin{tabular}{>{\raggedright\arraybackslash} p{0.9cm}>{\raggedright\arraybackslash}p{1.5cm}>{\raggedright\arraybackslash}p{1.4cm}>{\raggedright\arraybackslash}P{1.4cm}>{\raggedright\arraybackslash}p{1.4cm}}
        % \toprule
        \rowcolor{gray!15}
        \textbf{ODE Solver} &  \textbf{Description} & \textbf{Pros} & \textbf{Cons} & \textbf{Complexity} \\
        Convex & 
        One-step solution of flow
        & 
        \begin{minipage}[t]{\linewidth}
        \begin{itemize}[nosep,after=\strut, leftmargin=*]
            \item[+] Fastest solver
        \end{itemize}
        \end{minipage}
        & 
        \begin{minipage}[t]{\linewidth}
        \begin{itemize}[nosep,after=\strut, leftmargin=*]
           \item[-] Only for convex surfaces
        \end{itemize}
        \end{minipage}
        & $\mathcal{O}(1)$\\
        \rowcolor{gray!5}
        Euler method & 
        Numerical first-order ODE solver
        &
        \begin{minipage}[t]{\linewidth}
        \begin{itemize}[nosep,after=\strut, leftmargin=*]
           \item[+] Any curvature
            \item[+] Fast
        \end{itemize}
        \end{minipage}
        & 
        \begin{minipage}[t]{\linewidth}
        \begin{itemize}[nosep,after=\strut, leftmargin=*]
           \item[-] Very imprecise
        \end{itemize}
        \end{minipage}
        & $\mathcal{O}(N)$ \\
        
        Runge-Kutta (RK4) & 
        Numerical fourth-order ODE solver
        & 
        \begin{minipage}[t]{\linewidth}
        \begin{itemize}[nosep,after=\strut, leftmargin=*]
            \item[+] Any curvature
            \item[+] Precise
        \end{itemize}
        \end{minipage}
        & 
        \begin{minipage}[t]{\linewidth}
        \begin{itemize}[nosep,after=\strut, leftmargin=*]
            \item[-] Slowest method
        \end{itemize}
        \end{minipage}
        & $\mathcal{O}(4N)$ \\
        % \bottomrule
    \end{tabular}
\end{table}

\subsubsection{Reactive Obstacle Avoidance}
\label{app:reactive_obs_avoidance}
The following is a proof which shows that reactive obstacle avoidance as in Eq.~\eqref{eq:naive_obs_avoidance} can violate contraction stability.

\begin{mdframed}[hidealllines=true,backgroundcolor=gray!8,innerleftmargin=.2cm,
  innerrightmargin=.2cm]
\begin{proof}
\label{proof:violation_contraction_property}
Consider the obstacle-avoidance term as described in Equation~\eqref{eq:naive_obs_avoidance}, we need to show the contraction conditions as described in Theorem~\ref{theorem:Contraction_Conditions}. Therefore, taking the derivative of $\hat{f}_\text{c}$ leads to,
\begin{equation}
    \frac{\partial \hat{f}_\text{c}}{\partial \mathbf{q}} = \frac{\partial}{\partial \mathbf{q}} \left(f_\text{c}(\mathbf{q}) - b(\mathbf{x}, \mathbf{q})\nabla_\mathbf{q}\Gamma_\text{SDF} \right) = \mathbf{F} - \mathbf{H}_\text{SDF}(\mathbf{x}, \mathbf{q}),
\end{equation}
where, 
\begin{equation}
    \mathbf{H}_\text{SDF}(\mathbf{x}, \mathbf{q}) = \frac{\partial^2\bigl(b(\mathbf{x}, \mathbf{q})\Gamma_\text{SDF}(\mathbf{x}, \mathbf{q})\bigr)}{\partial \mathbf{q}^2} ,
\end{equation}
is the Hessian of the \ac{SDF} with the proposed barrier function.

Substituting $\frac{\partial \hat{f}_\text{c}}{\partial \mathbf{q}}$ into Theorem~\ref{theorem:Contraction_Conditions}, we must satisfy,
\begin{equation}
    \dot{\mathbf{G}} + \mathbf{G} \frac{\partial \hat{f}_\text{c}}{\partial \mathbf{q}} + \frac{\partial \hat{f}_\text{c}^\top}{\partial \mathbf{q}} \mathbf{G} \preceq -2 \alpha \mathbf{G} .
\end{equation}
Multiplying out all terms using Equation~\eqref{eq:naive_obs_avoidance} leads to,
\begin{equation}
    \mathbf{\dot{G}} + \mathbf{G} \mathbf{F} + \mathbf{F}^\top \mathbf{G} + \mathbf{G} \mathbf{H}_\text{SDF} + \mathbf{H}_\text{SDF}^\top \mathbf{G} \preceq -2 \alpha \mathbf{G} .
\end{equation}
Since $f_\text{c}$ is contractive by design, we known from Theorem~\ref{theorem:Contraction_Conditions} that,
\begin{equation}
    \mathbf{\dot{G}} + \mathbf{G} \frac{\partial f_\text{c}}{\partial \mathbf{q}} + \frac{\partial f_\text{c}^\top}{\partial \mathbf{q}} \mathbf{G} \preceq -2 \alpha \mathbf{G},
\end{equation}
so it remains to show that,
\begin{equation}
    \mathbf{G} \mathbf{H}_\text{SDF} + \mathbf{H}_\text{SDF}^\top \mathbf{G} \preceq \mathbf{0} .
\end{equation}
If $\mathbf{H}_\text{SDF}$ has positive eigenvalues, then $\mathbf{G} \mathbf{H}_\text{SDF} + \mathbf{H}_\text{SDF}^\top \mathbf{G}$ can be positive definite, thus violating contraction. As $\Gamma_\text{SDF}$ may have any shape, we cannot avoid having positive eigenvalues in $\mathbf{H}_\text{SDF}$ and thus violate the contraction condition.
\end{proof}
\end{mdframed}

\subsubsection{Signed Distance Field Diffeomorphic Transform}
\label{app:SDFDiffeomorphicTransform}
In the following, we prove that the \ac{SDC} preserves the contraction property and that the contraction metric is equivalent to the Riemannian metric of the obstacle-avoiding manifold $\mathcal{Y}$. The proof is inspired by \citet{Manchester2017}.
\vspace{1cm}
\begin{mdframed}[hidealllines=true,backgroundcolor=gray!8,innerleftmargin=.2cm,
  innerrightmargin=.2cm]
\begin{proof}
\label{proof:sdt_contraction}

Consider a contractive system,

\begin{equation}
\label{eq:proof_c_sys}
\mathbf{\dot q} = f_{\text{c}}(\mathbf{q}),
\end{equation}
which can be described via its differential dynamics,

\begin{equation}
\label{eq:proof_diff_c_sys}
\dot{\delta}_{\mathbf{q}} = \mathbf{A}(\mathbf{q}, t) \delta_{\mathbf{q}} \quad \text{s.t.} \quad \mathbf{A} := \frac{\partial f_{\text{c}}}{\partial \mathbf{q}} .
\end{equation}

Given the differential dynamics, a differential coordinate change $\delta_\mathbf{q} = \mathbf{J}_\psi\delta_\mathbf{y}$ can be introduced based on the smooth diffeomorphism $\psi$, defined by the flow in Equation \ref{eq:sdt_flow_forward}. By applying the coordinate change $\delta_\mathbf{q} = \mathbf{J}_\psi\delta_\mathbf{y}$ to the differential contractive system, this results in,

\begin{equation}
\label{eq:proof_diff_sys}
\dot \delta_{\mathbf{y}}= \mathbf{A}_{\mathbf{y}}(\mathbf{y}, t)\delta_{\mathbf{y}}.
\end{equation}

Given $\delta_\mathbf{q} = \mathbf{J}_\psi\delta_\mathbf{y}$, $\dot \delta_{\mathbf{y}}$ can be calculated via,

\begin{equation}
\label{eq:proof_phi_coord_change}
\dot{\delta}_{\mathbf{q}} = \overset{\cdot}{\mathbf{J}_\psi \delta_{\mathbf{y}}} = \dot{\mathbf{J}_\psi} \delta_{\mathbf{y}} + \mathbf{J}_\psi \dot{\delta}_{\mathbf{y}} .
\end{equation}
Given Equation \ref{eq:proof_diff_c_sys} and the coordinate changes $\delta_\mathbf{q} = \mathbf{J}_\psi\delta_\mathbf{y}$, Equation \eqref{eq:proof_phi_coord_change} can be transformed as follows,

\begin{equation}
\label{eq:proof_phi_coord_change2}
\dot{\delta}_{\mathbf{y}} =\underbrace{ \left(- \mathbf{J}_\psi^{-1} \dot{\mathbf{J}}_\psi + \mathbf{J}_\psi^{-1} \mathbf{A} \mathbf{J}_\psi\right)}_{\mathbf{A}_y} \delta_{\mathbf{y}} .
\end{equation}

Note that the contraction metric $\mathbf{G}_\mathbf{q}$ of $f_\text{c}$ is given by $\mathbf{G}_\mathbf{q} = \mathbf{I}$, since the contractive system $f_\text{c}$ is defined in Euclidean space. For the contraction metric $\mathbf{G}_\mathbf{y}$ of the dynamical system mapped into the obstacle-avoiding manifold we choose,
\begin{equation}
    \mathbf{G}_\mathbf{y}=\mathbf{J}_\psi^\top \mathbf{G}_\mathbf{q}\mathbf{J}_\psi=\mathbf{J}_\psi^\top \mathbf{J}_\psi,
\end{equation}
where $\mathbf{J}_\psi$ is the the Jacobian of the flow $\psi$.

The derivation of the metric results in,
\begin{equation}
   \mathbf{\dot G}_\mathbf{y}= \mathbf{\dot J}^\top\mathbf{J} + \mathbf{J}^\top  \mathbf{\dot J} .
\end{equation}
Given $\mathbf{G}_\mathbf{y}, \mathbf{\dot G}_\mathbf{y}, \mathbf{A}_\mathbf{y}$, the contraction condition from Theorem~\ref{theorem:Contraction_Conditions} can be shown in the following,
\begin{align}
&\mathbf{\dot G}_\mathbf{y} + \mathbf{A}_\mathbf{y}^\top \mathbf{G}_\mathbf{y} + \mathbf{G}_\mathbf{y} \mathbf{A}_\mathbf{y} + 2\alpha \mathbf{G}_\mathbf{y} \prec \mathbf{0} ,\nonumber \\
\Leftrightarrow 
  &\dot{\mathbf{J}}\psi^\top \mathbf{J}_\psi + \mathbf{J}\psi^\top \dot{\mathbf{J}}_\psi + \mathbf{J}_\psi^\top \mathbf{J}_\psi(\mathbf{J}_{\psi}^{-1} A_c \mathbf{J}_\psi - \mathbf{J}_\psi^{-1}\dot {\mathbf{J}}_{\psi}) + \nonumber \\
  &(\mathbf{J}_{\psi}^{-1} A_c \mathbf{J}_\psi - \mathbf{J}_\psi^{-1}\dot {\mathbf{J}}_{\psi})^\top \mathbf{J}_\psi^\top \mathbf{J}_\psi + 2\alpha \mathbf{J}_\psi^\top \mathbf{J}_\psi \preceq \mathbf{0} ,
 \nonumber \\
\Leftrightarrow &\dot{\mathbf{J}}\psi^\top \mathbf{J}_\psi + \mathbf{J}\psi^\top \dot{\mathbf{J}}_\psi 
+ \mathbf{J}_\psi^\top A_c \mathbf{J}_\psi - \mathbf{J}_\psi^\top \dot {\mathbf{J}}_{\psi}+ \mathbf{J}_\psi^\top A_c^\top  \mathbf{J}_\psi - \nonumber \\
&\dot {\mathbf{J}}_{\psi}^\top \mathbf{J}_\psi +2\alpha \mathbf{J}_\psi^\top \mathbf{J}_\psi \preceq \mathbf{0} ,\nonumber \\
\Leftrightarrow &\mathbf{J}_\psi^\top 
(\mathbf{A}  + \mathbf{A}^\top  +2\alpha \mathbf{I}) \mathbf{J}_\psi \preceq \mathbf{0} .
\end{align}

It is known from matrix calculations that the definiteness of a matrix $\mathbf{A}$ under $\mathbf{J}_\psi^\top\mathbf{A}\mathbf{J}_\psi$ is preserved. Therefore, only the term $(\mathbf{A}  + \mathbf{A}^\top + 2\alpha \mathbf{I})$ must be negative definite. This property holds by the construction of the contractive system, implying that the entire \ac{SDC} system is also contractive. \end{proof}
\end{mdframed}

\subsection{Experiments}
\begin{figure}[t]
    \centering
    \begin{subfigure}[b]{0.4\textwidth}
        \centering
        \includegraphics[width=\textwidth]{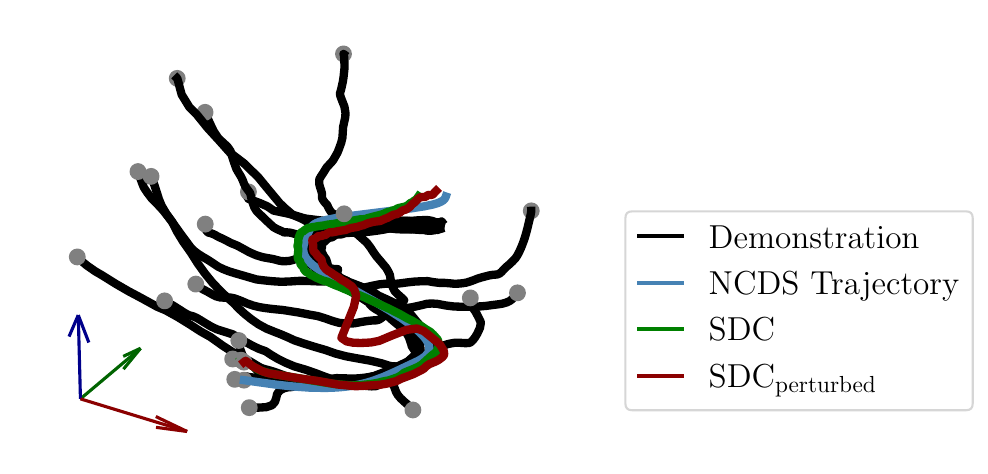}
        \caption{ Trajectories}
        \label{fig:run_pert_pos}
    \end{subfigure}
    \vspace{0mm}
    \begin{subfigure}[b]{0.45\textwidth}
        \centering
        \includegraphics[width=\textwidth]{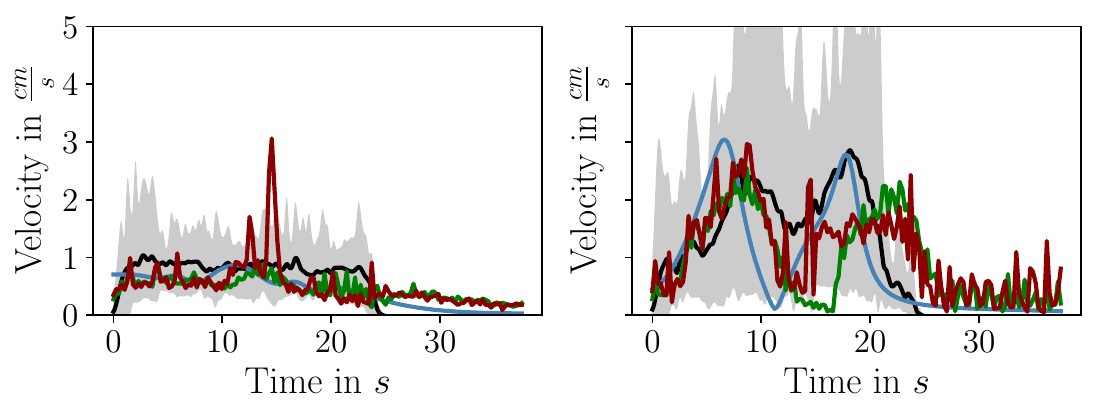}
        \caption{Velocities Series - Position (left), Orientation (right)}
        \label{fig:run_pert_vel}
    \end{subfigure}
    \caption{Move Sine skill performed on the robot with external perturbation}
    \label{fig:run_pert_plots}
\end{figure}
\subsubsection{Evaluation Metrics}
\label{app:metrics}

We use the \ac{DTWD}~\cite{Berndt1994} defined as,
\begin{equation}
    \label{eq:dtwd}
    \begin{split}
    &\text{DTWD}\big(\tau_\text{base}(\mathbf{x}),\tau_{m}(\mathbf{x})\big) = \\ 
    &\sum_{j \in l(\tau_{\text{base}}(\mathbf{x}))}\min_{i \in  l(\tau_{m}(\mathbf{x}))} d\big(\tau_{i,\text{base}}(\mathbf{x}),\tau_{j,m}(\mathbf{x})\big) + \\
    &\sum_{i \in l(\tau_{m}(\mathbf{x}))}\min_{j \in  l(\tau_\text{base}(\mathbf{x}))} d\big(\tau_{i,\text{base}}(\mathbf{x}),\tau_{j,m}(\mathbf{x})\big) ,
    \end{split}
\end{equation}
measuring the similarity between two trajectories, which may vary in length and/or velocity~\cite{Berndt1994}. In our context, $\tau_\text{base}$ denotes the base trajectory of length $l(\tau_\text{base})$, and it is compared to a collision-free modulated trajectory $\tau_{m}$ of length $l(\tau_{m})$. A low \ac{DTWD} indicates that the original vector field was slightly modulated by the obstacle avoidance term.

The minimum distance between the robot surface and the obstacle over the course of the skill execution is defined as follows,
\begin{equation}
    \label{eq:min_distance}
    D_\text{min} = \min_{\mathbf{x}\in \tau_m}\Gamma_\text{SDF}(\mathbf{x}) .
\end{equation}

The \textit{\ac{MJ}} metric is defined as,
\begin{equation}
    \label{eq:mj}
    \text{MJ} = \left\vert \max_{\mathbf{x} \in \tau_\text{base}} \Vert \dddot{\tau}_\text{base}(\mathbf{x}) \Vert - \max_{\mathbf{x} \in \tau_{m}} \Vert \dddot{\tau}_{m}(\mathbf{x}) \Vert \right\vert .
\end{equation}
In our context, $\tau_\text{base}$ denotes the base trajectory, and it is compared to a collision-free modulated trajectory $\tau_{m}$. A high \ac{MJ} indicates that the modulated vector field has a stronger jerk than in the original skill motion profile.

\subsubsection{Dishwasher Dataset}
\begin{figure*}[h]
     \centering
     \begin{subfigure}[b]{0.16\textwidth}
         \centering
         \includegraphics[width=\textwidth]{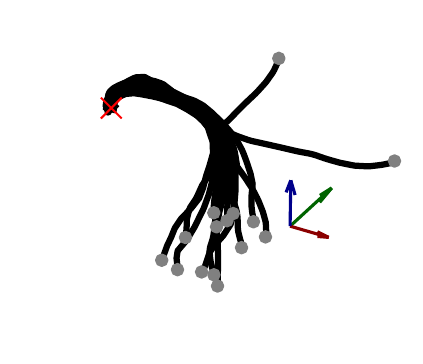}
         \caption{Place Plate}
         \label{fig:dataset_place_plate}
     \end{subfigure}
     \vspace{0mm}
     \begin{subfigure}[b]{0.16\textwidth}
         \centering
         \includegraphics[width=\textwidth]{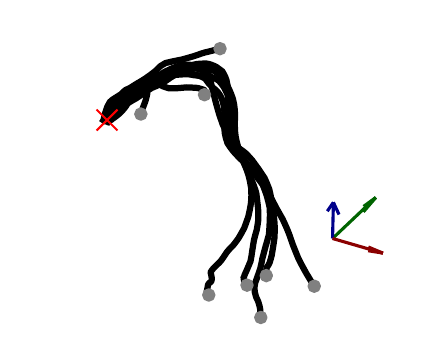}
         \caption{Place Plate in Rack}
         \label{fig:dataset_place_plate_rack}
     \end{subfigure}
     \vspace{0mm}
     \begin{subfigure}[b]{0.16\textwidth}
         \centering
         \includegraphics[width=\textwidth]{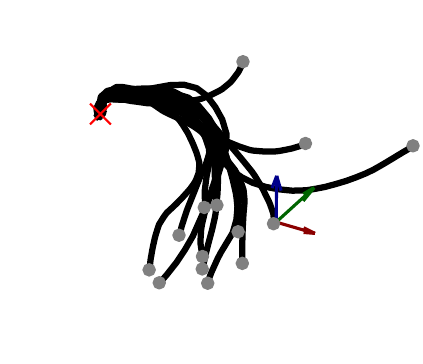}
         \caption{Place Cup}
         \label{fig:dataset_place_cup}
     \end{subfigure}
     \vspace{0mm}
     \begin{subfigure}[b]{0.16\textwidth}
         \centering
         \includegraphics[width=\textwidth]{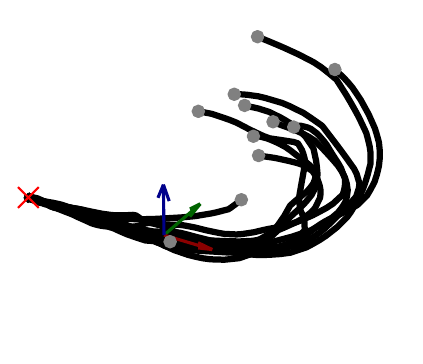}
         \caption{Grasp Doorhandle}
         \label{fig:dataset_grasp_doorhandle}
     \end{subfigure}
     \vspace{0mm}
     \begin{subfigure}[b]{0.16\textwidth}
         \centering
         \includegraphics[width=\textwidth]{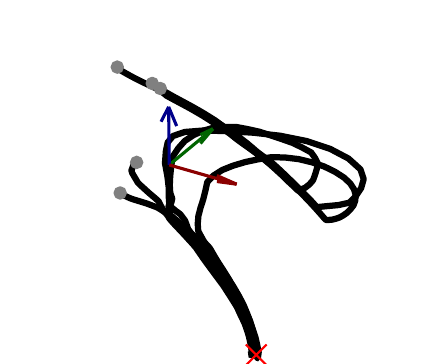}
         \caption{Open Door}
         \label{fig:dataset_open_door}
     \end{subfigure}
     \vspace{0mm}
     \begin{subfigure}[b]{0.16\textwidth}
         \centering
         \includegraphics[width=\textwidth]{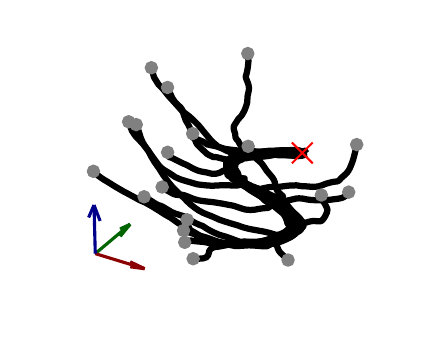}
         \caption{Move Sine}
         \label{fig:dataset_move_sine_3}
     \end{subfigure}
        \caption{Overview of all $6$ motion sequences in the dishwasher dataset. Demonstrations are displayed as black solid trajectories, with start and end points represented as red crosses and gray points, respectively.}
        \label{fig:Dishwasher_Dataset}
\end{figure*}

Figure~\ref{fig:Dishwasher_Dataset} shows all the demonstrated skills used for the interaction with the dishwasher. Note that the last state in the demonstration is repeated $N=20$ times to ensure that the contractive model learns a dynamic that reaches a target state at the end of the trajectory. If the goal is to learn an oscillating motion or a motion without a stable final state, this is not necessary.

\begin{figure*}[t]
    \centering
    \begin{subfigure}[b]{0.32\textwidth}
        \centering
        \includegraphics[width=\textwidth]{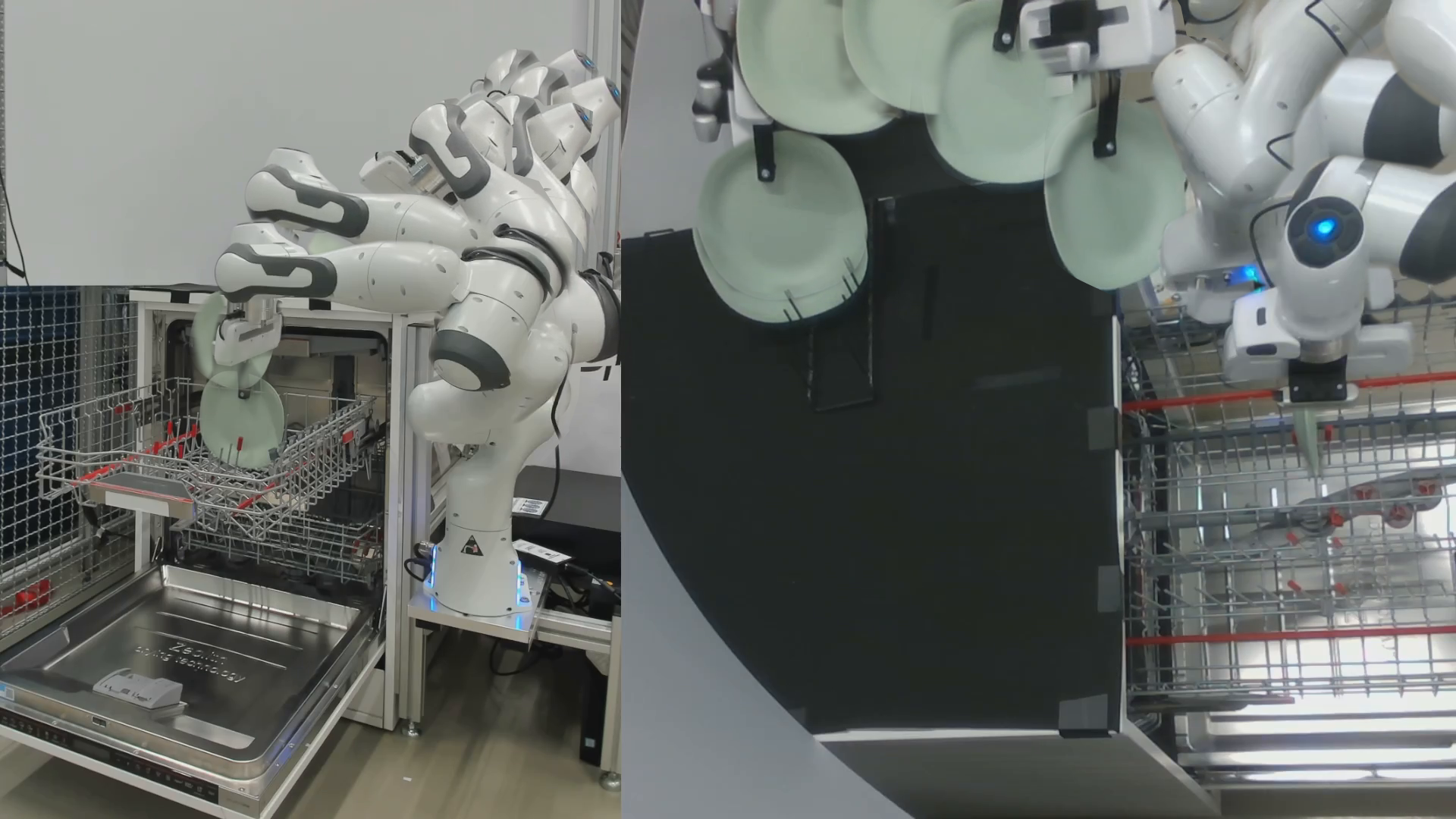}
        \caption{Motion sequence without obstacle avoidance}
        \label{fig:run_base_p}
    \end{subfigure}
    \vspace{0mm}
    \begin{subfigure}[b]{0.32\textwidth}
        \centering
        \includegraphics[width=\textwidth]{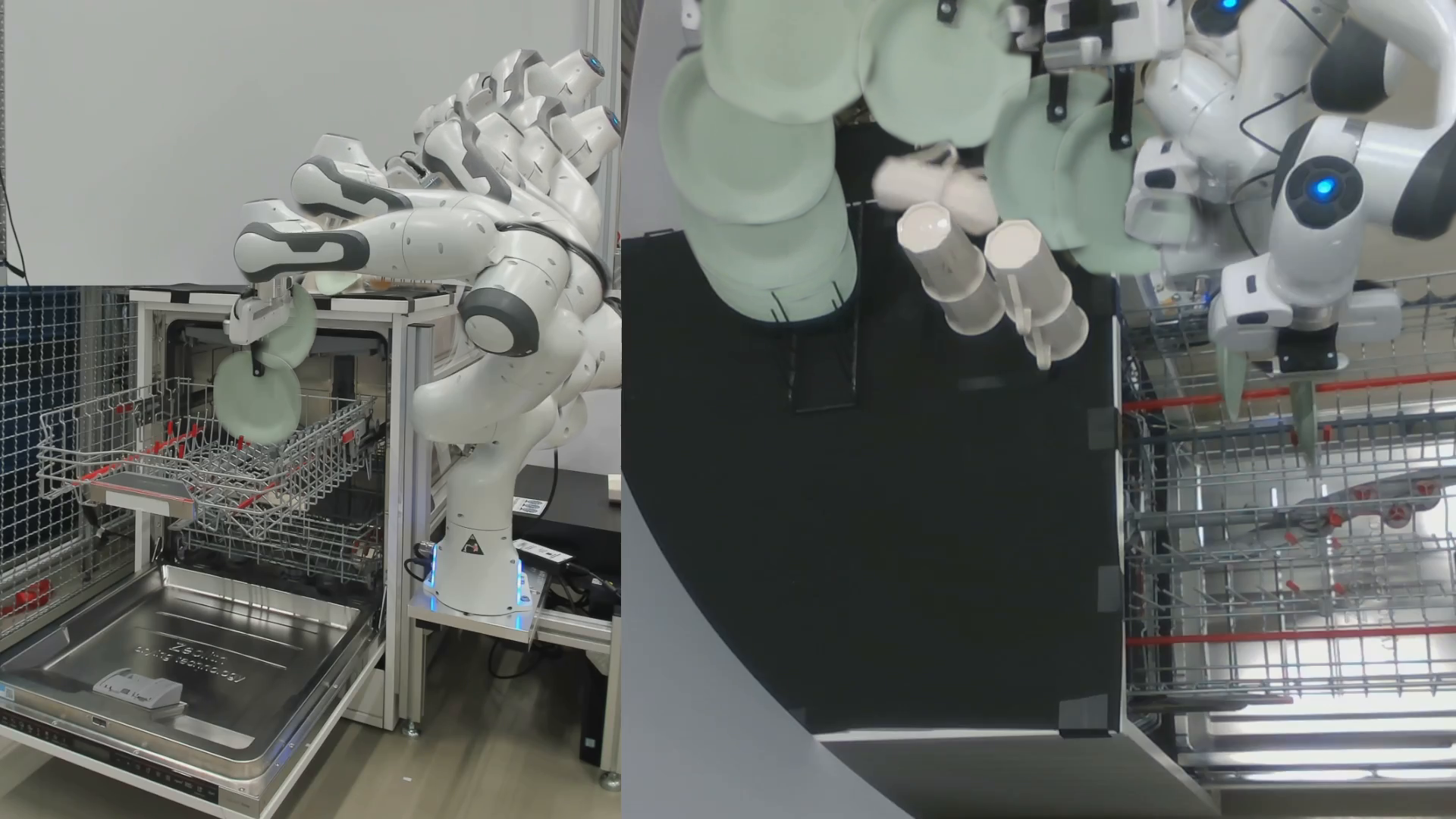}
        \caption{Obstacle avoidance without plate SDF}
        \label{fig:run_obs_cdf_ms_extend}
    \end{subfigure}
    \vspace{0mm}
    \begin{subfigure}[b]{0.32\textwidth}
        \centering
        \includegraphics[width=\textwidth]{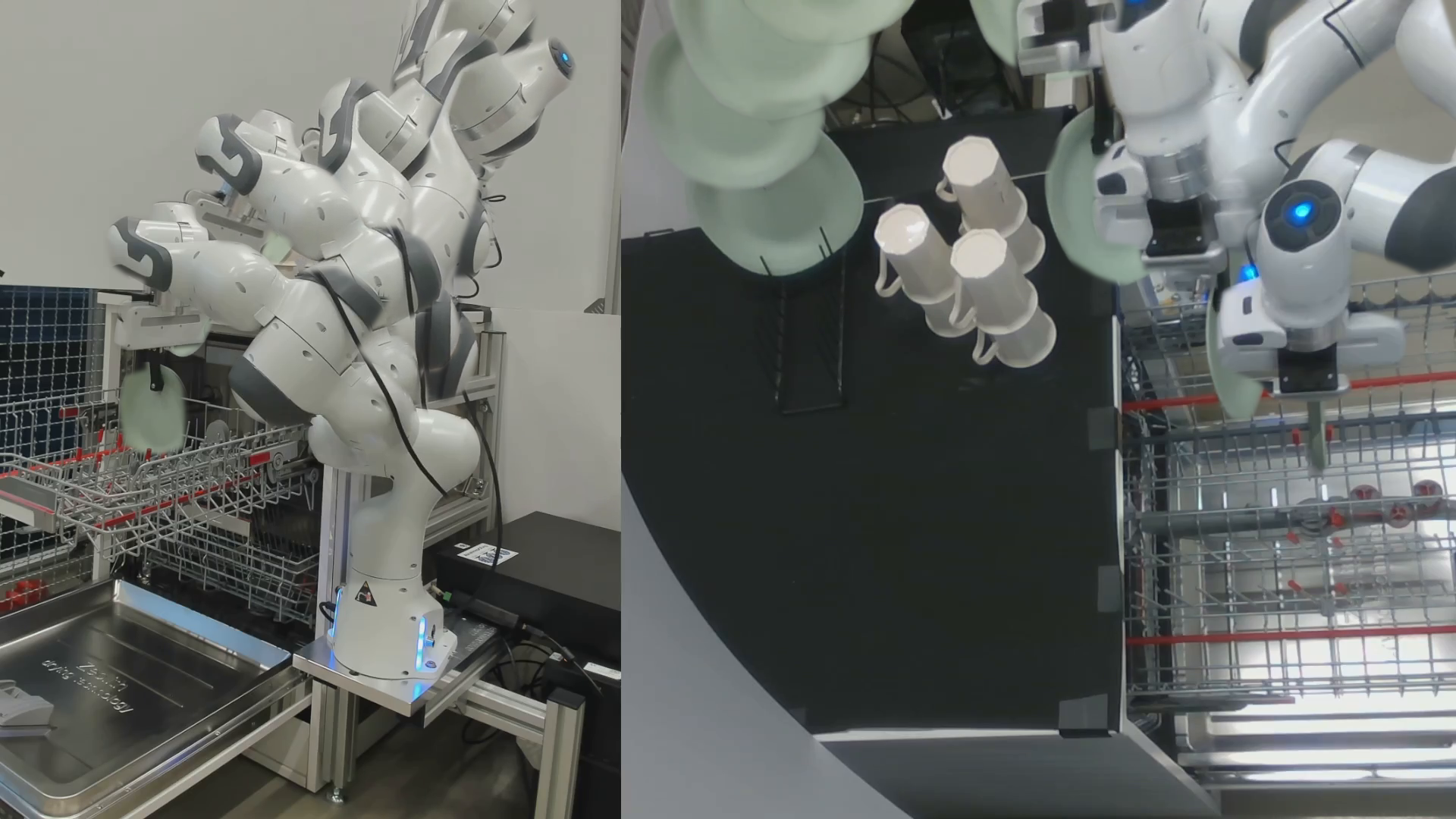}
        \caption{Obstacle avoidance with plate SDF}
        \label{fig:run_obs_cdf_traj_extend}
    \end{subfigure}
    \caption{
    \textsc{Emptying the dishwasher} task: A series of overlapped snapshots shows the \textsc{Place Plate in Rack} skill performed by the Franka-Emika Panda robot using an \ac{CDF} with grasped plate in the right panel and without the grasped plate in the middle panel. The left plot shows the execution of the skill without obstacles.}
    \label{fig:run_obs_cdf_extend}
\end{figure*}

\subsubsection{Robustness Test}
In the following, the contraction behavior and robustness of the \textit{Move Sine} skill is tested by introducing a manual perturbation during skill execution. The result is shown in Fig.~\ref{fig:run_pert_plots}.
The disturbances are observed with a velocity spike (Fig.~\ref{fig:run_pert_vel}), a position deflection (Fig.~\ref{fig:run_pert_pos}) and a slight orientation offset (Fig.~\ref{fig:run_pert_vel}). The successful stabilized disturbance in Fig.~\ref{fig:run_pert_pos} shows the robustness and contraction stability of the learned model.

\subsubsection{Real World Skill Execution}
Fig.~\ref{fig:run_base_p} shows the robot successfully executing the \textit{Place Plate in Rack} skill. 

\subsubsection{Real World Obstacle Avoidance}

Further experiments in Fig.~\ref{fig:run_obs_cdf_ms_extend} and Fig.~\ref{fig:run_obs_cdf_traj_extend} demonstrate the need to extend the robot's \ac{SDF} with the \ac{SDF} of the grasped object.

\subsubsection{Real World Inference time} Table~\ref{tab:inference_time_sdf_real_ex} shows the inference times of ODE solvers during real-world experiments, which approximately match the complexities $\mathcal{O}$ listed in Table~\ref{tab:flow_solver} of the paper's appendix. Clearly, the choice of the \ac{SDF} is most critical. 

\begin{table}[h]
\centering
\begin{minipage}{0.23\textwidth}
    \centering
    \footnotesize
    \begin{tabular}{lrr}%\toprule
        \rowcolor{gray!15}
        \textbf{Solvers} & \textbf{RDF} & \textbf{CDF} \\
        \rowcolor{gray!15}
        \textbf{SDC} &  &  \\
        Convex & $20.9$ & $2.4$ \\
        \rowcolor{gray!5}
        Euler & $41.1$ & $4.5$ \\
        \makecell{Runge-Kutta} & $91.2$ & $9.3$\\
    \end{tabular}
\end{minipage}%
\begin{minipage}{0.25\textwidth}
    \caption{Comparison of the inference time $t_\text{step}$ (in $ms$) required for a diffeomorphism step $\psi$ for the \ac{SDC} with $1000$ points, across different solvers.}
    \label{tab:inference_time_sdf_real_ex}
\end{minipage}
\vspace{-2em}
\end{table}

\end{document}